%% file: acl_latex.tex
\pdfoutput=1

\documentclass[11pt]{article}

\usepackage[preprint]{acl}

\input{math_commands.tex}

\usepackage{times}
\usepackage{latexsym}
\usepackage{subcaption}
\usepackage{booktabs}
\usepackage{multirow}

\usepackage{tikz}
\usetikzlibrary{shapes.geometric, arrows.meta, positioning, calc}

\usepackage[T1]{fontenc}

\usepackage[utf8]{inputenc}

\usepackage{microtype}

\usepackage{inconsolata}

\usepackage{graphicx}

\title{Circuit Compositions: Exploring Modular\\Structures in Transformer-Based Language Models}

\author{
Philipp Mondorf$^{1,2}$\thanks{Equal Contribution. Author order decided by coin flip.}\hspace{6mm}Sondre Wold$^{3}$\footnotemark[1]\hspace{6mm}Barbara Plank$^{1,2}$ \\
$^1$MaiNLP, Center for Information and Language Processing, LMU Munich, Germany\\
$^2$Munich Center for Machine Learning (MCML), Munich, Germany\\
$^3$Language Technology Group, University of Oslo\\
{\tt\footnotesize \{p.mondorf, b.plank\}@lmu.de, sondrewo@uio.no} \\
}

\begin{document}
\maketitle
\begin{abstract}
A fundamental question in interpretability research is to what extent neural networks, particularly language models, implement reusable functions through subnetworks that can be composed to perform more complex tasks. Recent advances in mechanistic interpretability have made progress in identifying \emph{circuits}, which represent the minimal computational subgraphs responsible for a model’s behavior on specific tasks. However, most studies focus on identifying circuits for individual tasks without investigating how functionally similar circuits \emph{relate} to each other. To address this gap, we study the modularity of neural networks by analyzing circuits for highly compositional subtasks within a transformer-based language model. Specifically, given a probabilistic context-free grammar, we identify and compare circuits responsible for ten modular string-edit operations. Our results indicate that functionally similar circuits exhibit both notable node overlap and cross-task faithfulness. Moreover, we demonstrate that the circuits identified can be reused and combined through set operations to represent more complex functional model capabilities.
\end{abstract}

\section{Introduction}\label{sec:introduction}
Neural networks can be effectively modeled as causal graphs that illustrate how inputs are mapped to the output space~\citep{mueller2024questrightmediatorhistory}. For instance, the feed-forward and attention modules within the Transformer architecture~\citep{Vaswani@2017attention} can be viewed as a series of causal nodes that guide the transformation from input to output via the residual stream~\citep{ferrando2024primer}. This abstraction is commonly used in mechanistic interpretability to identify computational subgraphs, or \emph{circuits}, responsible for the network's behavior on specific tasks~\citep{wang2023interpretability}. Circuits are typically identified through causal mediation analysis~\citep{pearl2001direct}, which quantifies the causal influence of model components on the network's predictions~\citep{mueller2024questrightmediatorhistory}. Techniques such as activation patching~\citep{meng2022locating}, attribution patching~\citep{syed-etal-2024-attribution}, and their variants~\citep{hanna2024have} have been applied to identify circuits in language models for tasks such as indirect object recognition~\citep{wang2023interpretability, merullo2024circuit}, entity tracking~\citep{Prakash2024FineTuningEE}, and factual recall \citep{meng2022locating}.

\begin{figure}[t!]
  \centering 
  \input{figures/tikz/intro_figure}
  \caption{
  Schematic overview of our approach: we identify and compare circuits for functionally related string edit operations, such as reversing a string or swapping its first and last characters.
  }
  \label{fig:schematic_intro}
\end{figure}

However, a notable limitation of existing studies is their focus on identifying circuits for isolated, individual tasks. Few studies have \emph{compared} circuits responsible for different functional behaviors of the model, and those that do primarily focus on tasks with limited cross-functional similarity \citep{hanna2024have}. In this study, we explore the modularity of neural networks by comparing circuits responsible for highly compositional subtasks within a transformer-based sequence-to-sequence model. Specifically, we identify circuits associated with ten modular string-edit operations (Table~\ref{tab:set_operations}) on a probabilistic context-free grammar, introduced by PCFG SET~\citep{hupkes2020compositionality}. We analyze these circuits in terms of both node overlap and cross-task faithfulness, assessing their performance on functionally related tasks. To facilitate the study of circuits related to sequence-prediction tasks beyond single-token predictions, we propose an automatic circuit identification method called \emph{activation pruning through continuous sparsification}, which jointly optimizes for \emph{faithfulness} and \emph{minimality}\textemdash two key objectives in circuit discovery (for further details, see Section~\ref{subsec:circuit_properties}). Finally, we demonstrate that the circuits identified can be combined through subnetwork set operations to explain more complex functional capabilities of the model.

\begin{table}[tbp]
    \centering
    \small
    \begin{tabular}{@{}c@{\hskip 10pt}lll@{}}
        \toprule
        & Operation & Input & Output \\
        \midrule
        \multirow{6}{*}{\rotatebox[origin=c]{90}{Unary}} 
        & \texttt{copy} & $x_1 \dots x_n$ & $x_1 \dots x_n$ \\
        & \texttt{echo} & $x_1 \dots x_n$ & $x_1 \dots x_n x_n$ \\
        & \texttt{repeat} & $x_1 \dots x_n$ & $x_1 \dots x_n x_1 \dots x_n$ \\
        & \texttt{reverse} & $x_1 \dots x_n$ & $x_n \dots x_1$ \\
        & \texttt{swap} & $x_1 \dots x_n$ & $x_n x_2 \dots x_{n-1} x_1$ \\
        & \texttt{shift} & $x_1 \dots x_n$ & $x_2 \dots x_n x_1$ \\
        \midrule
        \multirow{4}{*}{\rotatebox[origin=c]{90}{Binary}} 
        & \texttt{append} & $x, y$ & $x \, y$ \\
        & \texttt{prepend} & $x, y$ & $y \, x$ \\
        & \texttt{remove\_first} & $x, y$ & $y$ \\
        & \texttt{remove\_second} & $x, y$ & $x$ \\
        \bottomrule
    \end{tabular}
    \caption{The different string-edit operations from PCFG SET~\citep{hupkes2020compositionality}. Unary operations modify a single string, while binary operations involve two different strings separated by ``,''.}
    \label{tab:set_operations}
\end{table}

In summary, our contributions are as follows: \emph{i}) we demonstrate the application of continuous sparsification to automatically discover both faithful and minimal circuits for sequence-to-sequence tasks from PCFG SET; \emph{ii}) we analyze the relationships between functionally related circuits by examining node overlap and cross-task faithfulness, providing insights into the model's modular structure; and \emph{iii}) we show that computational subgraphs from functionally related circuits can be combined using set operations, resulting in novel subnetworks that explain model behavior on tasks beyond the scope of the initial circuits.

\section{Background}\label{sec:background}
This section offers an overview of key concepts in circuit discovery and continuous sparsification used in this work. For a broader perspective on interpretability research based on causal mediation analysis, we refer to~\citet{mueller2024questrightmediatorhistory}.

\subsection{Activation Patching}\label{subsec:activation_patching}
A widely used approach for identifying circuits within language models is activation patching~\citep{NEURIPS2020_92650b2e, meng2022locating, wang2023interpretability}. Activation patching quantifies the causal influence of model components on the model's task output. More formally, it measures the \emph{indirect effect} (\text{IE}) \citep{pearl2001direct} of a node $\vx$ in the model's computation graph on a downstream node $\vy$, usually the final output \citep{mueller2024questrightmediatorhistory}. Given a counterfactual intervention $\tilde{\vz}$ on a mediator $\vz$, the indirect effect of $\vx$ on $\vy$ through $\vz$ is the difference in a metric $\mathbb{P}$ that captures $\vy$'s state before and after the intervention:

\begin{align}
    \text{IE}(\mathbb{P}, \vx, \vz, \vz_x, \tilde{\vz}) 
    &= \mathbb{P}\left(\vy(\vx) \mid \vz = \vz_{x}\right) \notag \\
    &\quad - \mathbb{P}\left( \vy(\vx) \mid \text{do}(\vz = \tilde{\vz})\right)
\end{align}

where $\vx$ typically is the model's task input, while $\vy$ denotes the corresponding output. The variable $\vz_x$ represents the mediator's natural value for $\vx$ without intervention, and $\tilde{\vz}$ its counterfactual value. In its original form, activation patching iteratively assesses the indirect effect of mediators, assigning causal significance when the indirect effect exceeds a predefined threshold \citep{NEURIPS2020_92650b2e, meng2022locating, wang2023interpretability}. The choice of mediator varies between studies, ranging from the full output activation $\va \in \R^{d}$ of a module (e.g., the multi-head attention module) or a submodule (e.g., a linear layer), to individual neurons $\eva_i \in \va$ \citep{mueller2024questrightmediatorhistory}. High-granularity interventions can lead to a combinatorial explosion of the search space when exhaustively exploring all mediators~\citep{mueller2024questrightmediatorhistory}. To mitigate this, alternatives like attribution patching balance accuracy and causal guarantees with improved search efficiency~\citep{syed-etal-2024-attribution, nanda2023attribution, hanna2024have}. In this study, we overcome this combinatorial problem by optimizing over a continuous approximation of the discrete search space (see Section~\ref{sec:activation_pruning} for details).

\subsection{Circuit Properties}\label{subsec:circuit_properties}
Let $\vz_i \in \gZ$ represent a single node\footnote{Depending on the granularity of the identification method, this can range from entire modules to individual neurons within the model.} in the model's causal graph, denoted as $\gG = \left(\gZ, \gE\right)$, where $\gE \subseteq \gZ \times \gZ$ and $|\gZ| = N$. Once a circuit has been successfully identified for a given model $\mathcal{M}$ and task $T$, it can be represented as a binary mask $\vm \in \{0, 1\}^N$ over the model's node space. This mask signifies whether a specific model component is causally relevant to the model's task behavior ($\evm_i = 1$) or not ($\evm_i = 0$)~\citep{bhaskar2024finding}. Circuits are generally evaluated based on three key criteria \citep{wang2023interpretability}:

\begin{enumerate}
    \item \emph{Faithfulness}: A circuit is considered faithful to the task $T$ if it accurately captures the full model’s task output while ablating all nodes not identified as causally relevant ($\evm_i = 0$) by replacing them with some ablation value $\tilde{\vz}$~\citep{hanna2024have, miller2024transformer}. Given a metric $P$ that compares the outputs of two models for task $T$, task faithfulness $F_T$ is typically quantified as:
    \begin{align}
    F_T &= P\Big( \mathcal{M}(\cdot), \, 
    \mathcal{M}\big(\cdot \mid \text{do} \big( \vz = \vm \odot \vz_x \notag \\
    &\quad + (\mathbf{1} - \vm) \odot \tilde{\vz} \big) \big) \Big)
    \label{eq:task_faithfulness}
    \end{align}
    \item \emph{Minimality}: A circuit is deemed minimal if it excludes nodes that are not causally relevant~\citep{mueller2024questrightmediatorhistory}. Formally, given a candidate set of circuits $C$, minimality is encouraged by selecting the circuit with the smallest norm: $\min_{\vm \in C} \| \vm \|_1$.
    \item \emph{Completeness}: A circuit is said to be complete if it captures all nodes necessary to explain the model’s behavior for task $T$.
\end{enumerate}

\subsection{Continuous Sparsification}\label{subsec:continuous_sparsification}
Continuous sparsification originates from model pruning and has been introduced to sparsify networks, specifically their weight space~\citep{savarese2020winning}. Unlike other pruning approaches \citep{8014795, louizos2018learning}, continuous sparsification approximates $l_0$ regularization by learning a \emph{deterministic} mask $\vm \in \{0, 1\}^N$ over the network's parameters $\vw \in \R^N$ that indicates which weights to prune. The search for such a sparse subnetwork can be represented by the following minimization problem:

\begin{align}
    \min_{\vw \in \R^N, \, \vm \in \{0, 1\}^N} 
    &\mathcal{L} \left( \mathcal{M} \left( \cdot \, ; \vm \odot \vw \right) \right) \notag \\
    &\quad + \lambda \, \| \vm \|_1
\end{align}

which uses the fact that $\| \vm \|_0 = \| \vm \|_1$ for binary masks, and where $\mathcal{L}$ denotes the loss of the network $\mathcal{M}$, while $\lambda$ controls the trade-off between loss and number of parameters $\| \vw \|_0$. To circumvent the combinatorial constraint imposed by the discrete space of $\vm \in \{0, 1\}^N$, the mask is deterministically re-parameterized as a sigmoid function $\sigma\left(\cdot\right)$ of the new variable $\vs \in \R^N$:

\begin{align}
    \min_{\vw \in \R^N, \, \vs \in \R^N} 
    &\mathcal{L} \left( \mathcal{M} \left( \cdot \, ; \sigma (\beta \cdot \vs) \odot \vw \right) \right) \notag \\
    &\quad + \lambda \, \| \sigma (\beta \cdot \vs) \|_1
    \label{eq:continuous_sparsification}
\end{align}

where $\beta \in [1, \infty]$ represents a temperature parameter for which the sigmoid function converges to the Heaviside function with $\lim_{\beta \rightarrow \infty} \sigma\left(\beta \cdot \vs\right) = H(\vs)$. By minimizing the above loss while annealing $\beta$, and recovering the binary mask from re-parameterization via $\vm = H(\vs)$, a sparse representation of the network's parameters can be learned.

\section{Activation Pruning Through Continuous Sparsification}\label{sec:activation_pruning}
To automatically identify circuits for sequence-to-sequence prediction tasks, we adopt an approach similar to~\citet{bhaskar2024finding} and~\citet{conmy2023towards}, and formulate the identification process as a minimization problem. For this, we leverage techniques from model pruning, specifically \emph{continuous sparsification} (see Section~\ref{subsec:continuous_sparsification}). Unlike traditional pruning, which aims to reduce model complexity by creating a sparse representation of the model's weight space, we focus on intervening on the model's activations, thereby linking our method to causal mediation analysis.

As outlined in Section~\ref{subsec:circuit_properties}, we represent a circuit as a binary mask $\vm \in \{0, 1\}^N$ over the model's mediator space\textemdash here the activation space of the model components is considered\textemdash{}indicating which activations $\vz$ of a frozen model $\mathcal{M}$ are responsible for its behavior on a task $T$. To find a circuit $\vm$ that is both \emph{faithful} and \emph{minimal} (we exclude \emph{completeness} for now; see Section~\ref{sec:Limitations} for a respective discussion), we aim to minimize the following loss:

\begin{align}
    \min_{\vm \in \{0, 1\}^N} 
    &\mathcal{L}_{T} \big( \mathcal{M}(\cdot),  
    \mathcal{M}(\cdot \mid \text{do} (\vz = \vm \odot \vz_x  \notag \\
    &\quad + (1 - \vm) \odot \tilde{\vz})) \big)
    + \lambda \cdot \mathcal{L}_{reg}(\vm)
    \label{eq:final_binary_loss}
\end{align}

where $\mathcal{L}_{T}$ captures the circuit's task faithfulness, with lower values indicating greater faithfulness, while $\mathcal{L}_{reg}$ assesses the size of the circuit. The hyperparameter $\lambda$ controls the influence of $\mathcal{L}_{reg}$. Given the combinatorial complexity of optimizing a binary mask over a potentially large activation space, we follow the approach of continuous sparsification (Section~\ref{subsec:continuous_sparsification}) and deterministically re-parameterize $\vm$ using a sigmoid function $\sigma(\cdot)$ over the new variable $\vs \in \R^N$. Additionally, we assess faithfulness by measuring the Kullback-Leibler divergence $D_{KL}\left(\vy^{\vm} \parallel \vy^{\mathcal{M}}\right)$ between the circuit's predicted output distribution and that of the full model. Since the full model's predicted output distribution $\vy^{\mathcal{M}}$ is independent of the variable $\vs$, minimizing the KL divergence between $\vy^{\vm}$ and $\vy^{\mathcal{M}}$ simplifies to minimizing their cross-entropy loss $\mathcal{L}_{CE}$. For regularization, we apply $l_1$ regularization in line with Equation~\ref{eq:continuous_sparsification}. This yields:

\begin{align}
    \min_{\vs \in \R^N} 
    &\mathcal{L}_{CE} \big( \mathcal{M}(\cdot),  
    \mathcal{M}(\cdot \mid \text{do} (\vz = \sigma(\beta \cdot \vs) \odot \vz_x  \notag \\
    &\quad + (1 - \sigma(\beta \cdot \vs)) \odot \tilde{\vz})) \big)  \notag \\
    &\quad + \lambda \cdot \|\sigma(\beta \cdot \vs)\|_1
    \label{eq:final_loss}
\end{align}

where the sigmoid function $\sigma$ is applied element-wise, and $\beta$ serves as a temperature parameter that increases progressively after each training epoch, following an exponential schedule until it reaches a maximum value, $\beta_{max}$, as proposed by \citet{lepori2023break}. By minimizing the expression in Equation~\ref{eq:final_loss}, we obtain an approximation of $\vm$ that strikes a balance between \emph{faithfulness} and \emph{minimality}, with $\lambda$ governing the emphasis on the latter. Once training converges, the binary mask is derived through re-parameterization via $\vm = H(\vs)$. In comparison to edge pruning~\citep{bhaskar2024finding} and the subnetwork probing method proposed by \citet{conmy2023towards}, our approach identifies circuits through \emph{deterministic} re-parameterization. While our method supports different levels of node granularity, we focus on neuron-level interventions.

\section{Experiments}\label{sec:experiments}
This work studies the modularity of a transformer-based sequence-to-sequence model by analyzing circuits responsible for its behavior on highly compositional subtasks. We identify and compare circuits for ten compositional string-edit operations introduced by PCFG SET \citep{hupkes2020compositionality}.\footnote{Our code is available at~\href{https://github.com/mainlp/circuit-compositions}{https://github.com/mainlp/circuit-compositions}.} Section~\ref{subsec:setup} outlines the experimental setup, including dataset, training, and evaluation details, while Section~\ref{subsec:results} presents the results.

\subsection{Setup}\label{subsec:setup}

\subsubsection{PCFG SET}\label{subsubsec:dataset}
As shown in Table~\ref{tab:set_operations}, PCFG SET~\citep{hupkes2020compositionality}  comprises ten string-edit operations applied to sequences generated by a probabilistic context-free grammar. All tasks resemble translation problems, where an input sequence is transformed into a corresponding output sequence through the recursive application of the operations specified in the input sequence. The dataset includes \emph{unary} operations (applied to a single string) and \emph{binary} operations (requiring two arguments). For instance, the binary function $\texttt{prepend}$ places the second argument before the first (e.g., $\texttt{prepend}\, A1\, , B1 \rightarrow B1\, A1$). The input alphabet in PCFG SET consists of three components: \emph{i}) words representing string-edit operations (e.g.,  $\texttt{copy}$ or  $\texttt{echo}$), \emph{ii}) symbols forming the input sequence (e.g.,  $\texttt{A1}$,  $\texttt{B1}$, etc.), and \emph{iii}) a separator ``, '' that distinguishes for binary arguments. For additional information and examples, please refer to Appendix~\ref{appendix:section:dataset:examples}.

\citet{hupkes2020compositionality} construct PCFG SET such that compositionality is a salient feature. Notably, all operators in PCFG SET are functionally related. For example, the $\texttt{repeat}$ operator can be replicated by applying the $\texttt{copy}$ operation two times in succession (see Table~\ref{tab:set_operations}). To identify circuits for each operation, we generate ten distinct data subsets, each containing 20,000 examples from a specific string-edit operation (16,000 for training, 4,000 for testing). Further details on data generation and individual sub-datasets are provided in Appendix~\ref{appendix:section:dataset:generating-isolated-function-data}.

\subsubsection{Training}\label{subsubsec:training}
\paragraph{Base Model Training.} As a first step, we train a base model $\mathcal{M}$ to perform all operations in PCFG SET. Similar to~\citet{hupkes2020compositionality}, we use an encoder-decoder model, comprising six encoder and decoder layers with a hidden state dimension of 512, resulting in approximately 58 million parameters. The model is trained on the official data splits of PCFG SET, which include around 83,000 training samples covering all string-edit operations and their compositions. Additional details on the training procedure can be found in Appendix~\ref{appendix:section:base-model-training}.

\begin{figure*}[tbp]
    \centering
    \begin{subfigure}[b]{0.3\textwidth}
    \centering
        \includegraphics[width=\textwidth]{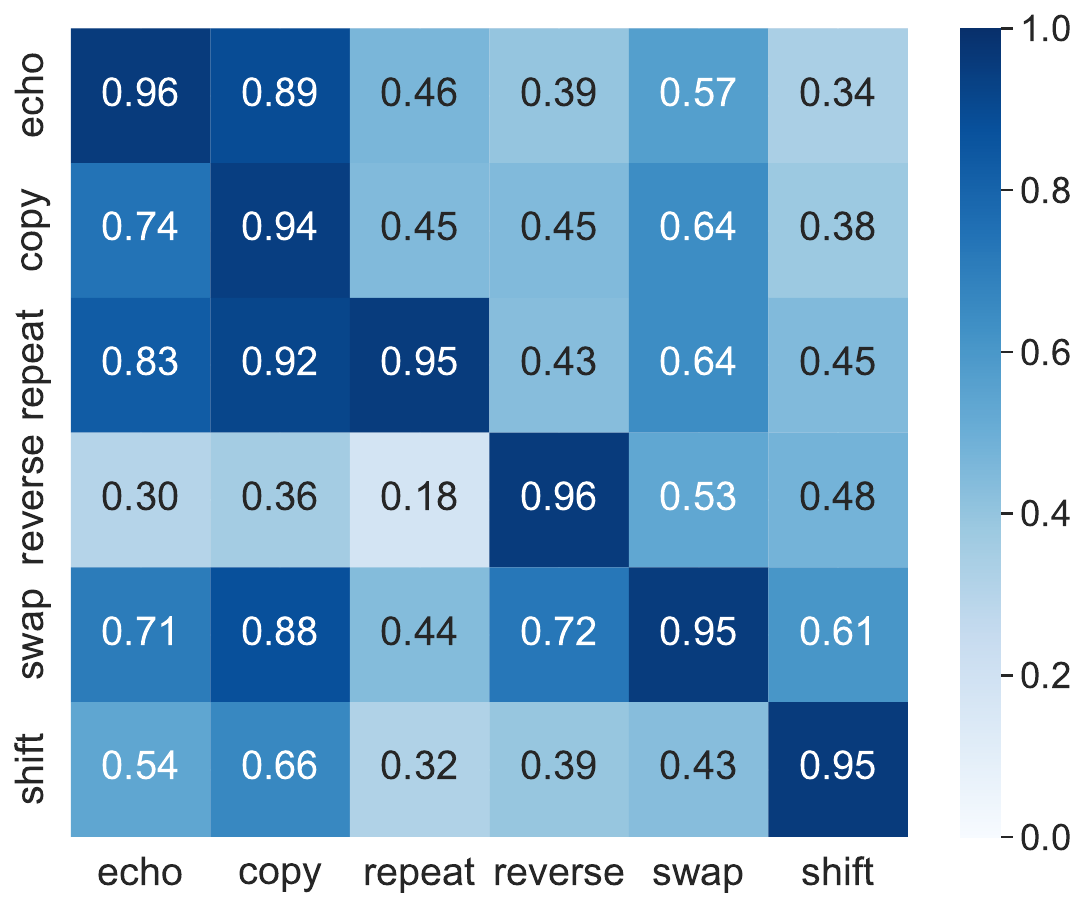}
        \caption{$F_T$ across all token positions.}
        \label{subfig:unary-jsd-all-tokens}
    \end{subfigure}
    \hfill
    \begin{subfigure}[b]{0.3\textwidth}
    \centering
        \includegraphics[width=\textwidth]{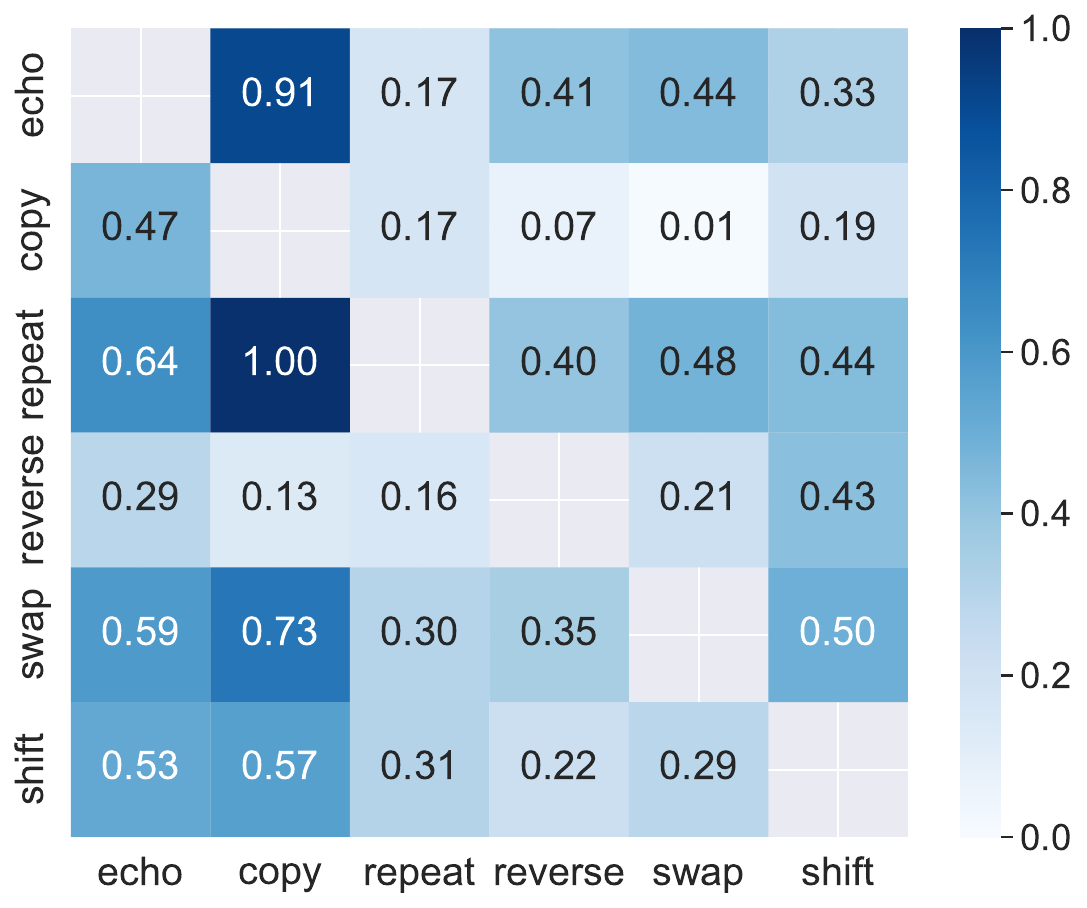}
        \caption{$F_T$ for selected token positions.}
        \label{subfig:unary-jsd-different-tokens}
    \end{subfigure}
    \hfill
    \begin{subfigure}[b]{0.3\textwidth}
        \centering
        \includegraphics[width=\textwidth]{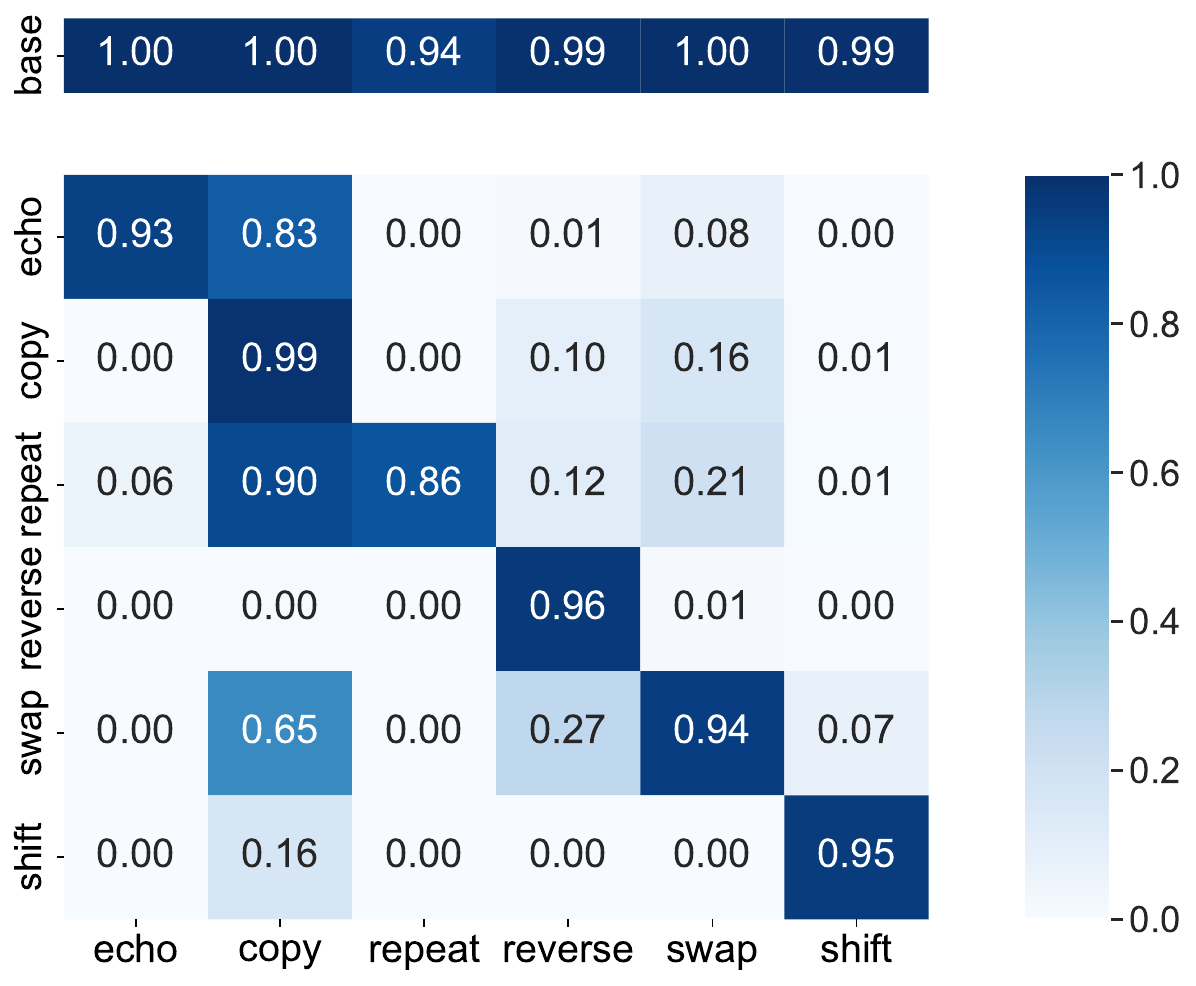}
        \caption{Accuracy}
        \label{subfig:unary-accuray}
    \end{subfigure}
    \caption{Task faithfulness $F_T$ and accuracy for \textbf{unary} tasks. The y-axis represents the circuit, while the x-axis denotes the evaluation task. The diagonal is omitted for selected positions due to a lack of applicable tokens.}
    \label{fig:unary-performance}
\end{figure*}

\paragraph{Mask Training.} Next, we employ \emph{activation pruning through continuous sparsification} (Section~\ref{sec:activation_pruning}) to identify circuits corresponding to each operation. For this, we minimize the loss as described in Equation~\ref{eq:final_loss} by training a mask $\vm $ on the respective subtask dataset (see Section~\ref{subsubsec:dataset}). We consider individual output activations from the feed-forward and multi-head attention modules as mediators $\vz = \eva_i \in \va$. For the ablation value $\tilde{\vz}$, we conduct experiments with both zero and mean ablations. We use the same ablation value across all token positions and adopt an approach similar to node patching, where interventions target the model’s residual stream. When employing mean ablations, we follow the approach proposed by \citet{wang2023interpretability} and use the mediator’s mean value across a reference distribution, specifically the subtask dataset. We optimize for hyperparameters using random search, as detailed in Appendix~\ref{appendix:section:mask-training-hyperparameters}. A sensitivity analysis of the parameter $\lambda$ is provided in Appendix~\ref{appendix:section:lambda-sensitivity}. Further details on the mask training procedure can be found in Appendix~\ref{appendix:section:mask-training}.

\subsubsection{Evaluation}\label{subsubsec:evaluation}
We evaluate circuits based on two main criteria: performance\textemdash{}measured in terms of faithfulness and accuracy\textemdash{}and node overlap. Since the KL-divergence $D_{KL}\left(\vy^{\vm} \parallel \vy^{\mathcal{M}}\right)$ is unbounded and unsuitable for cross-task comparisons, we also compute a normalized version of the Jensen–Shannon divergence, $0\leq\operatorname{JSD}_{\text{norm}}\leq1$, to measure faithfulness (see Equation~\ref{eq:jsd-faithfulness}, Appendix~\ref{appendix:section:mask-evaluation}). We define a circuit's task \emph{faithfulness performance} as $F_T = 1 - \operatorname{JSD}_{\text{norm}}$, where values closer to 1 indicate higher task faithfulness.  Accuracy is determined by the exact match between the circuit's prediction and the ground truth. When evaluating a circuit from task $T$ on a different task $\hat{T}$, we retain the mean values from task $T$ as ablation values for nodes with $\evm_i = 0$ when using mean ablations. For node overlap, we compare circuits using the Intersection over Union (IoU) and Intersection over Minimum (IoM), as defined in Equation~\ref{appendix:equation:iou_iom} in Appendix~\ref{appendix:section:mask-evaluation}.

\subsection{\textsc{Tracr} Experiments}\label{subsubsec:tracr_evaluation}
To validate our method's capacity to identify both \emph{minimal} and \emph{faithful} circuits, we implement four unary PCFG SET functions (\texttt{copy}, \texttt{echo}, \texttt{reverse}, \texttt{swap}) in RASP \citep{weiss2021thinking}. Each function is compiled into transformer model weights using \textsc{Tracr} \citep{lindner2024tracr}, serving as ground truth for evaluation\textemdash{}a common validation approach in prior studies~\citep{conmy2023towards, bhaskar2024finding}. Our results show that \emph{activation pruning through continuous sparsification} perfectly recovers all individual neurons in the compiled circuits while maintaining 100\% faithfulness. For further details, please refer to Appendix~\ref{appendix:section:tracr_validation}.

\subsection{Results}\label{subsec:results}
We begin by presenting the performance results of the base model on PCFG SET. After training, the encoder-decoder model demonstrates strong performance across all ten string-edit operations, with accuracy exceeding 95\% on unary tasks and ranging from 83\% to 99\% on binary tasks. The lowest accuracies of 83\% and 84\% are for \texttt{prepend} and \texttt{append}, which turn out as the most challenging operations (a detailed task-specific breakdown is available in Appendix~\ref{appendix:section:base-model-training}). In the subsequent sections, we focus on circuits identified through mean ablation. For additional results on circuits discovered via zero ablation, see Appendix~\ref{appendix:subsec:zero_ablation}.

\begin{figure*}[tbp]
    \centering
    \begin{subfigure}[b]{0.3\textwidth}
    \centering
        \includegraphics[width=\textwidth]{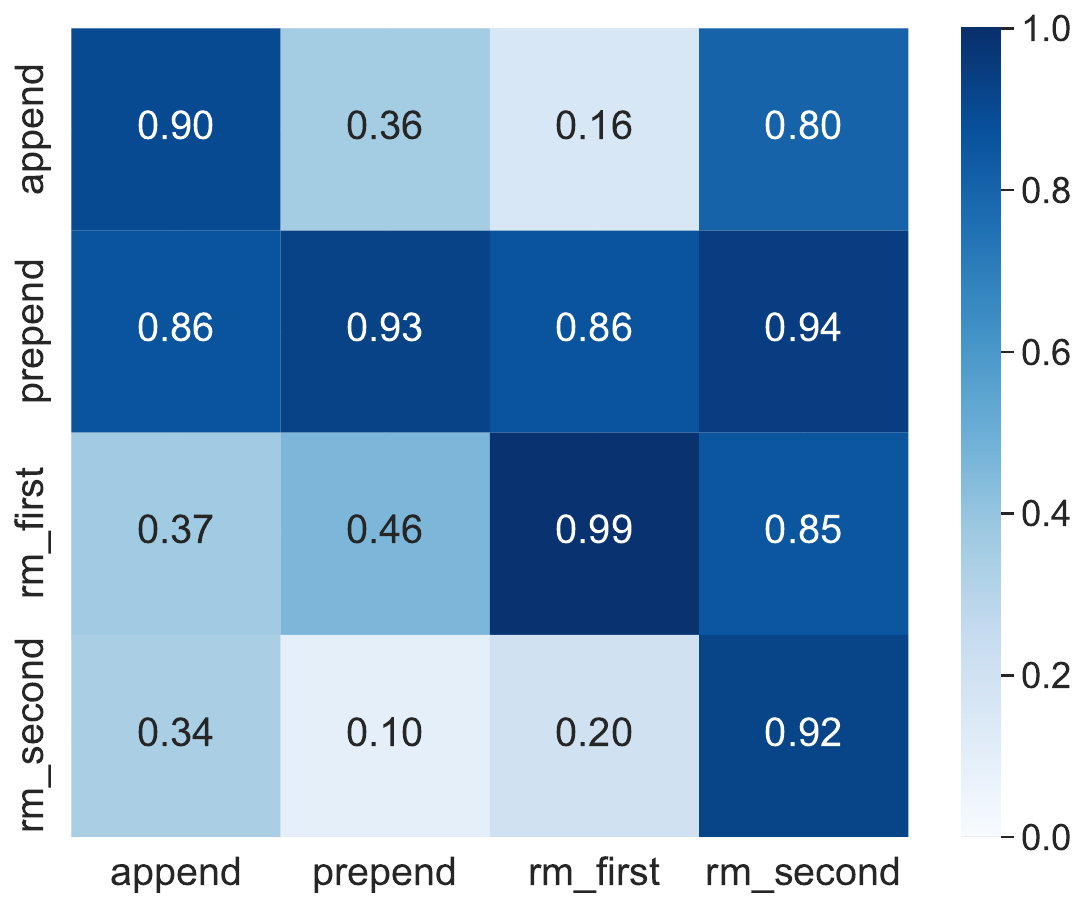}
        \caption{$F_T$ across all token positions.}
        \label{subfig:binary-jsd-all-tokens}
    \end{subfigure}
    \hfill
    \begin{subfigure}[b]{0.3\textwidth}
    \centering
        \includegraphics[width=\textwidth]{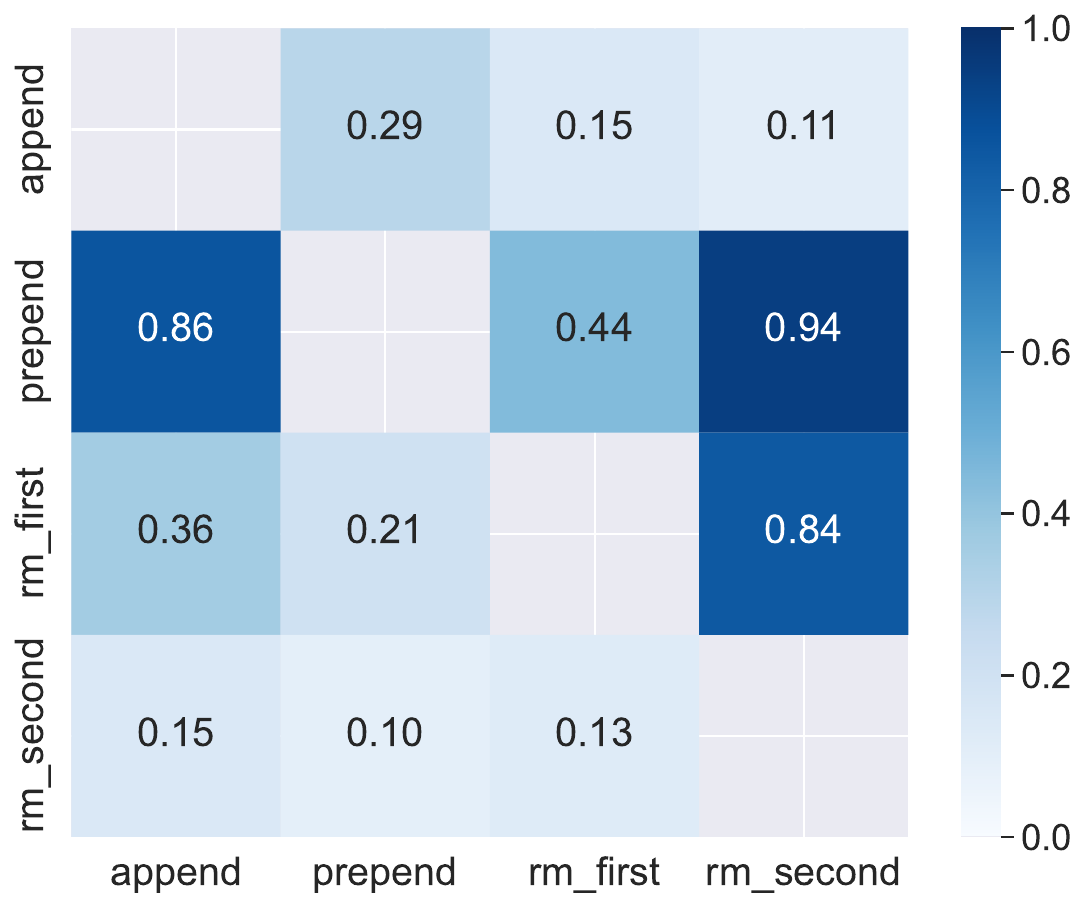}
        \caption{$F_T$ for selected token positions.}
        \label{subfig:binary-jsd-different-tokens}
    \end{subfigure}
    \hfill
    \begin{subfigure}[b]{0.3\textwidth}
        \centering
        \includegraphics[width=\textwidth]{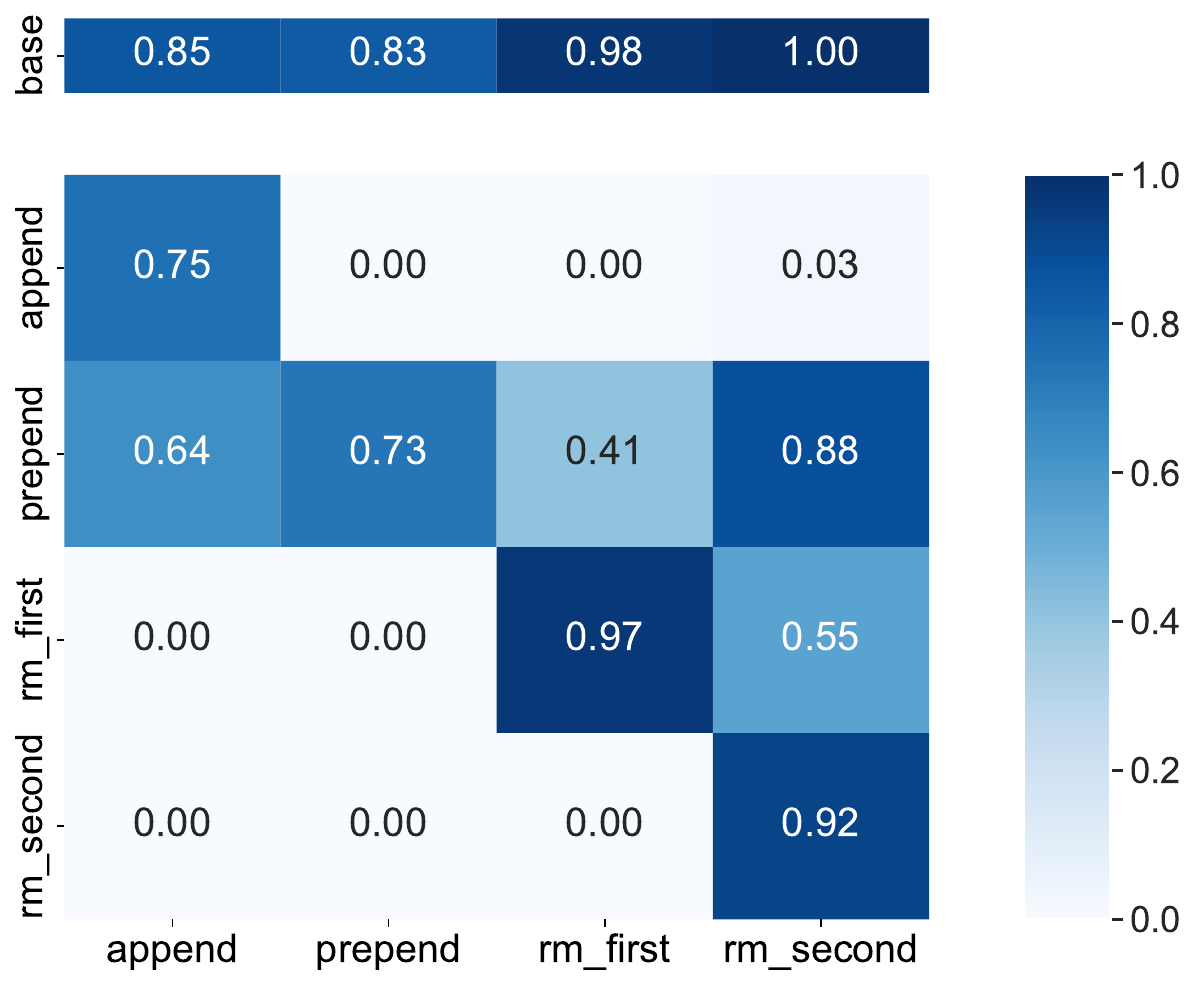}
        \caption{Accuracy.}
        \label{subfig:binary-accuray}
    \end{subfigure} 
    \caption{Task faithfulness $F_T$ and accuracy for \textbf{binary} tasks. The y-axis represents the circuit, while the x-axis denotes the evaluation task. The diagonal is omitted for selected positions due to a lack of applicable tokens.}
    \label{fig:binary-performance}
\end{figure*}

\subsubsection{Circuit Performance}\label{subsubsec:circuit_performance}
We first analyze the circuits' performance across different string-edit operations. Figure~\ref{fig:unary-performance} illustrates both the task faithfulness performance, $F_T$, as defined in Section~\ref{subsubsec:evaluation}, as well as the accuracy of each circuit. Due to functional similarities between the string-edit operations, multiple operators may produce the same output tokens at various positions. Therefore, we assess faithfulness in two ways: \emph{i)} averaged across all output tokens (Figure~\ref{subfig:unary-jsd-all-tokens}), and \emph{ii)} at positions where the ground truth output sequences of the circuit-task and evaluation-task differ (Figure~\ref{subfig:unary-jsd-different-tokens}). For example, when evaluating the \texttt{copy} circuit on the \texttt{echo} task, we assess task faithfulness at the final token of the target output, focusing on the additional $x_n$ present in \texttt{echo}, which replaces the end-of-sequence token in \texttt{copy}.

\paragraph{Unary Circuits.} The diagonal in Figure~\ref{subfig:unary-jsd-all-tokens} shows that all unary circuits exhibit strong faithfulness on their respective tasks, with rates exceeding $0.94$. Several circuits, such as \texttt{echo}, \texttt{repeat}, and \texttt{swap}, also demonstrate high cross-task faithfulness and accuracy on the \texttt{copy} task. This aligns with our intuition as the \texttt{copy} operation is either a significant component of these functions or can be effectively performed by them. In contrast, operators like \texttt{reverse} and \texttt{shift}, which substantially alter the input sequence, show lower functional similarity to \texttt{copy}. This is reflected in both the reduced cross-task faithfulness (Figures~\ref{subfig:unary-jsd-all-tokens};~\ref{subfig:unary-jsd-different-tokens}) and lower accuracy (Figure~\ref{subfig:unary-accuray}) on the \texttt{copy} task. Interestingly, while the \texttt{repeat} and \texttt{swap} circuits perform well on the \texttt{copy} task, the reverse is not true. This occurs despite the theoretical possibility of completing the \texttt{repeat} operation by applying the \texttt{copy} operator twice. Overall, circuits tend to show high cross-task faithfulness performance when evaluated on all tokens (Figure~\ref{subfig:unary-jsd-all-tokens}), but scores drop for tokens that differ between the circuit’s task and the evaluation task (Figure~\ref{subfig:unary-jsd-different-tokens}). For instance, the \texttt{copy} circuit achieves a cross-task faithfulness score of 0.64 on the \texttt{swap} task across the full sequence, but this drops to 0.01 when evaluated only on tokens that differ between the ground truth output sequences of \texttt{copy} and \texttt{swap}. This suggests that the \texttt{copy} circuit largely adheres to its original function, even when the input demands a \texttt{swap} operation.

\begin{figure}[b!]
    \centering
    \includegraphics[width=0.9\columnwidth]{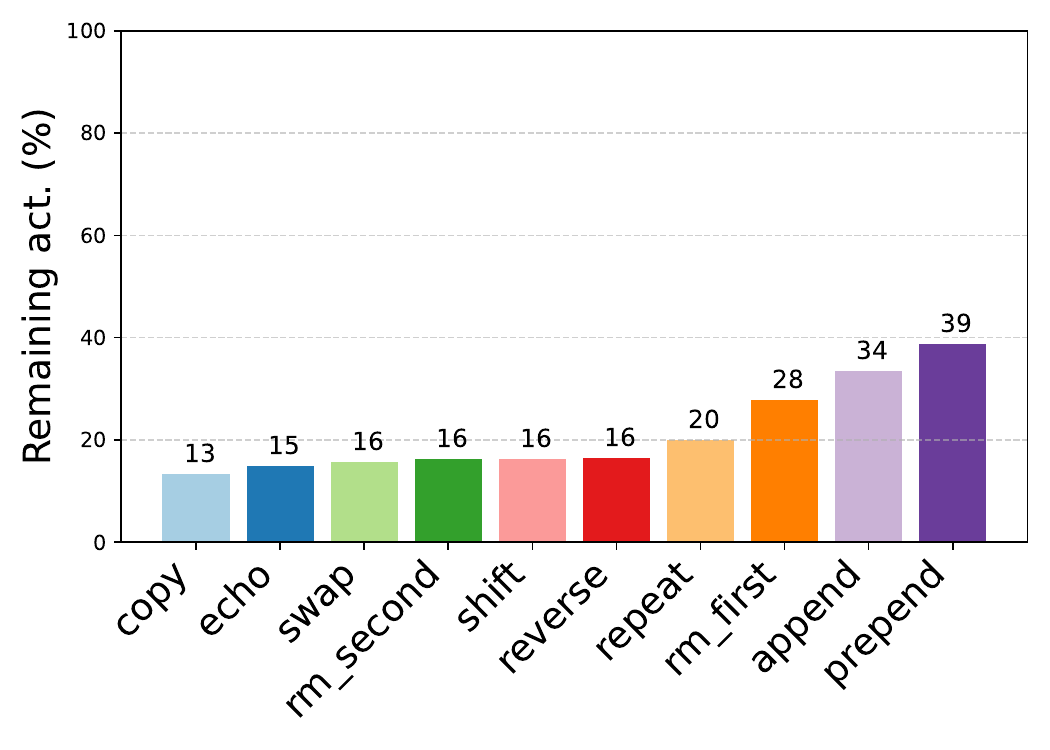}
    \caption{Sparsity of the mean ablated circuits as fraction of a circuit's remaining activations (in \%).}
    \label{fig:global-sparsity}
\end{figure}

\paragraph{Binary Circuits.} Figure~\ref{fig:binary-performance} highlights the performance of circuits associated with the binary operations. Similar to the unary circuits, these circuits demonstrate high faithfulness for their respective tasks, with performance values ranging from $0.90$ to $0.99$ (Figure~\ref{subfig:binary-jsd-all-tokens}). Notably, the \texttt{remove\_first} circuit shows strong cross-task faithfulness and accuracy on the \texttt{remove\_second} task, although the reverse is not true. Additionally, the \texttt{prepend} circuit shows consistently strong performance across all binary tasks. When analyzing the size of each circuit\textemdash{}the fraction of nodes with causal relevance to the model's task behavior, expressed as a fraction of remaining activations (Figure~\ref{fig:global-sparsity})\textemdash{}we observe that the \texttt{prepend} circuit is the largest among all circuits, retaining $39\%$ of its neuron activations. Similarly, the \texttt{remove\_first} circuit is notably larger than the \texttt{remove\_second} circuit. This pattern suggests that more complex operations (\texttt{prepend}, \texttt{append}, \texttt{remove\_first}) engage a greater portion of the model’s activation space compared to simpler tasks. For instance, \texttt{remove\_second} may rely on a straightforward \texttt{copy} operation with an end-of-sequence token at the position of the separator “,”, whereas \texttt{remove\_first} likely requires more intricate processing.

\begin{figure}[tbp]
    \centering
    \resizebox{0.75\columnwidth}{!}{%
        \input{figures/tikz/local_sparsity_copy}
    }
    \caption{The local sparsity of the \texttt{copy} circuit. For each layer, the fraction of remaining activations in the multi-head self-attention (MHSA), multi-head cross-attention (MHCA), and feed-forward (FF) modules are shown.}
    \label{fig:local-sparsity}
\end{figure}

\paragraph{Local sparsity.} To zoom in more, Figure~\ref{fig:local-sparsity} illustrates the local sparsity of the \texttt{copy} circuit, which is the smallest among all circuits (Figure~\ref{fig:global-sparsity}). It shows the percentage of remaining neuron activations in the multi-head self-attention (MHSA), multi-head cross-attention (MHCA), and feed-forward (FF) modules at each layer. Notably, in the decoder, MHCA modules remain mostly active, whereas FF and MHSA activations are almost entirely pruned. This aligns with our expectations, as the copy task primarily involves transferring encoder-processed input to the output. Additional visualizations for other circuits can be found in Appendix~\ref{appendix:section:circuit-sparsity}, specifically Figures~\ref{appendix:figure:local-sparsity-unary} and~\ref{appendix:figure:local-sparsity-binary}.

\begin{figure}[b!]
    \centering
    \begin{subfigure}[b]{0.94\linewidth}
        \centering
        \includegraphics[width=\linewidth]{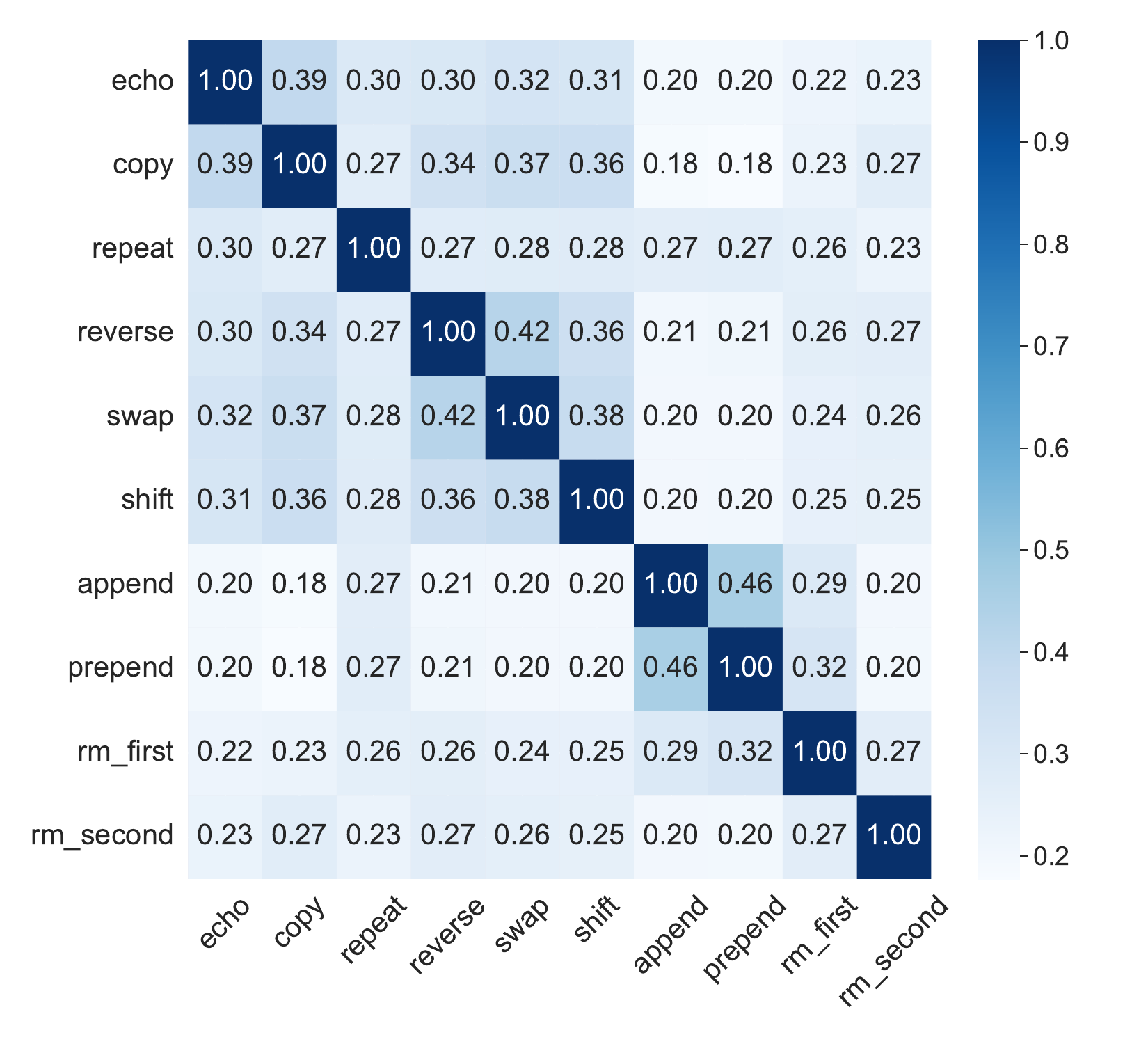}
        \caption{Intersection over Union (IoU).}
        \label{figure:cross-task-overlap-iou}
    \end{subfigure}
    \vspace{0.5cm}
    \begin{subfigure}[b]{0.94\linewidth}
        \centering
        \includegraphics[width=\linewidth]{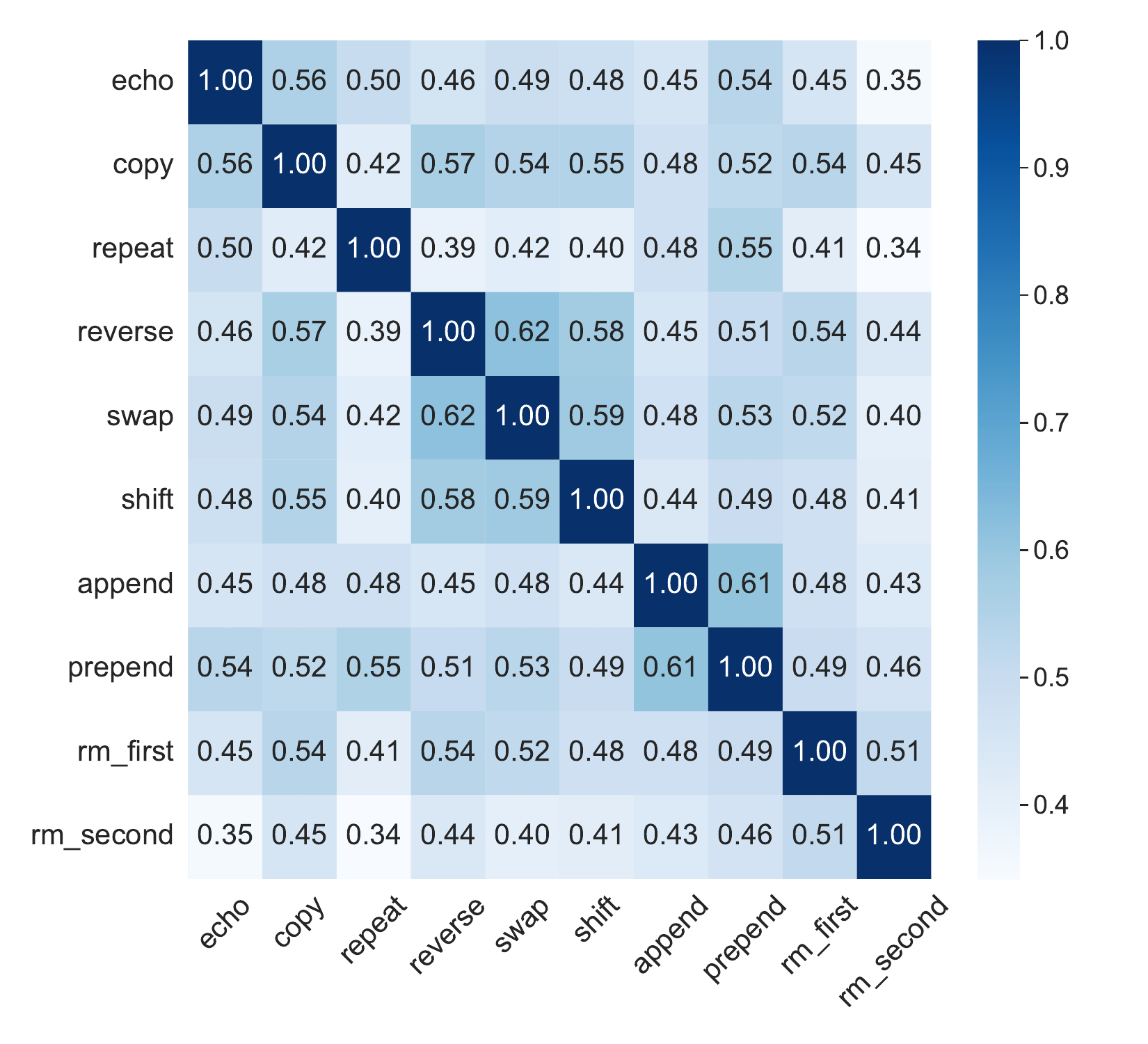}
        \caption{Intersection over Minimum (IoM).}
        \label{figure:cross-task-overlap-iom}
    \end{subfigure}
    \caption{Node overlap for circuits identified via activation pruning with mean ablation.}
    \label{fig:node-overlap}
\end{figure}

\paragraph{Zero vs. mean ablations.} The results so far relate to circuits identified via mean ablation. Consistent with prior work \citep{miller2024transformer}, we find zero ablations to be less faithful. For an overview of respective results, see Appendix~\ref{appendix:subsec:zero_ablation}.

\subsubsection{Circuit Overlap}\label{subsubsec:circuit_overlap}
So far, we evaluated the performance of circuits and their cross-task generalizability. Next, we show their respective node overlap in Figure~\ref{fig:node-overlap}. Specifically, we assess the IoU and IoM between all circuit pairs, as described in Section~\ref{subsubsec:evaluation}. The IoU captures the overlap of circuits relative to their combined size, whereas the IoM measures how much of the smaller circuit is contained within the larger one. Most pairs exhibit IoU values between $0.20$ and $0.30$, though some clusters show higher overlap. For instance, the circuits for \texttt{reverse}, \texttt{swap}, and \texttt{shift} have IoU values between $0.36$ and $0.42$, and IoM values around $0.60$, indicating significant shared activations across these tasks. Given the functional similarities of these operations, this finding aligns with our intuition. However, it is worth noting that the cross-task performance of \texttt{reverse}, \texttt{swap}, and \texttt{shift} is lower than that of other circuits (see Figure~\ref{fig:unary-performance}). A similarly high node overlap is observed for the \texttt{append} and \texttt{prepend} circuits\textemdash{}both associated with tasks that the base model $\mathcal{M}$ struggles with, as shown by the model's comparatively lower accuracy (see Table~\ref{appendix:table:base-model-acc} in Appendix~\ref{appendix:section:base-model-training}).

\begin{figure*}[tbp]
    \centering
    \begin{subfigure}{\textwidth}
        \centering
       \includegraphics[width=0.7\linewidth]{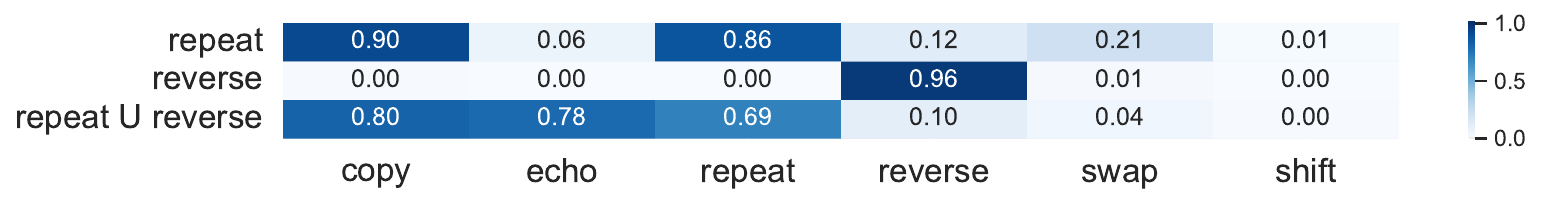}
    \end{subfigure}
    \vspace{1em}
    \begin{subfigure}{\textwidth}
        \centering
       \includegraphics[width=0.7\linewidth]{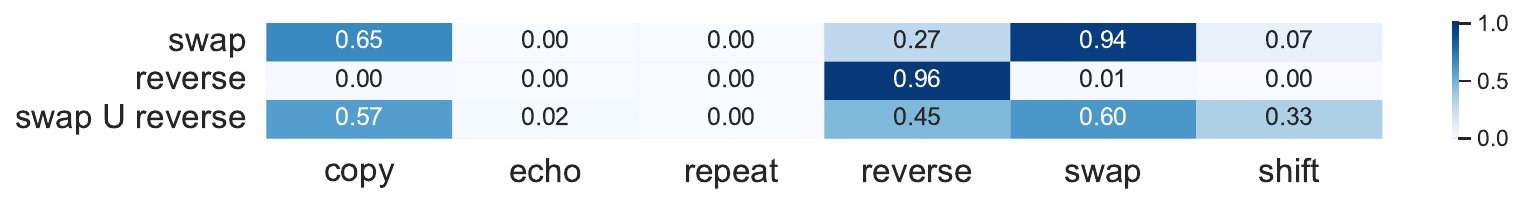}
    \end{subfigure}
    \caption{The results of combining circuits through a union operation on their respective binary masks.}
    \label{fig:exp:set-comparison}
\end{figure*}

\subsubsection{Circuit Compositions: Subnetwork Set Operations}\label{sec:set-operations}
Given that we can define a circuit as binary mask $\vm \in \{0, 1\}^N$ over the model's mediator space, we extend our analysis to study \emph{cross-circuit relationships} by creating compositions through basic set operations, as we want to gauge to what degree composite circuits emerge. Specifically, for a circuit-pair $\left(\vm^{T_1}, \vm^{T_2}\right)$, we define their composite as the union $\vm^{T_1, T_2} = \vm^{T_1} \cup \vm^{T_2}$. Importantly, this union is not symmetric: we use the ablation values $\tilde{\vz}^{T_1}$ of the first circuit for nodes outside the union ($\evm_i^{T_1, T_2} = 0$). For additional details on how this union is applied, please refer to Appendix~\ref{appendix:section:compositions}. 

Figure~\ref{fig:exp:set-comparison} shows the cross-task accuracy of various circuits and their composites, derived from the union operation previously described. The results demonstrate the existence of composite circuits which acquire functional capacities for subtasks that the original base circuits could not handle individually. For example, while neither the \texttt{repeat} nor the \texttt{reverse} circuit can solve the \texttt{echo} task alone, their union achieves a notable accuracy of $78\%$. Similarly, the composite of \texttt{swap} and \texttt{reverse} shows a performance improvement on the \texttt{shift} task, reaching an accuracy of $33\%$. For tasks that can be solved by at least one of the original circuits, the performance of their composite generally reflects a blend of their individual accuracies. For instance, the \texttt{repeat} circuit achieves $90\%$ accuracy on the \texttt{copy} task, while the \texttt{reverse} circuit yields $0.0\%$ accuracy. Their union achieves an intermediate accuracy of $80\%$ on the task. Similarly, the \texttt{swap} circuit achieves $27\%$ accuracy on the \texttt{reverse} task, whereas the \texttt{reverse} circuit reaches up to $96\%$. Their composite circuit achieves a balanced performance of $45\%$. This implies that improvements in accuracy for novel tasks may be accompanied by a decline in performance on previously mastered skills. 

A possible explanation for these observations is that the union operation $\vm^{T_1, T_2} = \vm^{T_1} \cup \vm^{T_2}$ shifts the activation space of $\vm^{T_1}$ towards $\vm^{T_2}$, effectively transforming the input into a feature space that aligns with one of the subtasks the model $\mathcal{M}$ can solve. For instance, the union $\vm^{\texttt{repeat}} \cup \vm^{\texttt{reverse}}$ might shift the activation space of $\vm^{\texttt{repeat}}$ towards $\vm^{\texttt{reverse}}$, essentially allowing the new circuit to perform the \texttt{echo} operation, while sacrificing some performance on the \texttt{repeat} task. Further experiments and results on subnetwork set operations can be found in Appendix~\ref{appendix:section:set-operations}.

Our results show that circuits responsible for tasks like \texttt{echo}, \texttt{repeat}, and \texttt{swap} can perform the \texttt{copy} task, whereas the \texttt{copy} circuit struggles to perform other tasks. This raises the question of whether the \texttt{echo}, \texttt{repeat}, and \texttt{swap} circuits leverage the \texttt{copy} operation in their respective functions. Our analysis on node overlap indicates that the \texttt{copy} circuit \textemdash{}which is notably the smallest among those identified\textemdash{}is largely embedded within the other circuits. Moreover, circuits with similar functions exhibit greater node overlap than functionally distinct ones. Additionally, we show that circuits combined through the union operator can represent novel functional capabilities, suggesting some degree of modularity within the network.

\section{Related Work}\label{sec:related_work}
Techniques that discover circuits through causal mediation analysis have been successfully applied across various domains and tasks, including the study of gender bias in language models \citep{jeoung-diesner-2022-changed, chintam-etal-2023-identifying}, different forms of factual recall \citep{meng2022locating, geva-etal-2023-dissecting}, subject-verb agreement \citep{chintam-etal-2023-identifying}, and arithmetic operations \citep{nanda2023progress, hanna2024does}. Finding subnetworks has also been of interest outside of the mechanistic interpretability literature. Previous work has studied task-specific subnetworks through clustering approaches \citep{casper2022graphical, watanabe2019interpreting}, pruning \citep{csordas2020neural, cao-etal-2021-low, lepori2023break}, sparse fine-tuning \citep{ansell-etal-2022-composable}, and adapters \citep{pfeiffer2021adapterfusion, ruckle2021adapterdrop}. However, few studies have used the circuits paradigm to examine relationships between circuits in functionally related tasks \citep{hanna2024have}.

\section{Conclusion}\label{sec:conclusion}
This study explores modularity in neural networks by analyzing circuits in a transformer-based language model for highly compositional subtasks. Specifically, we train an encoder-decoder model using the PCFG SET dataset and identify circuits responsible for ten modular string-edit operations. To achieve this, we introduce \emph{activation pruning through continuous sparsification}, a method that allows us to formulate circuit identification as minimization problem. Our results demonstrate that this approach successfully identifies both faithful and sparse circuits for each subtask within PCFG SET. Additionally, we assess all circuits by examining their cross-task performance and node overlap. Finally, we show promising first steps of circuit compositions, where new functional circuits can be composed through set operations, such as the union of two circuits.

\section{Limitations}\label{sec:Limitations}
In the following section, we mention limitations of our work that could be addressed by future studies.

\paragraph{Completeness.}  We introduce \emph{activation pruning through continuous sparsification}, a method that formulates the circuit discovery process as an optimization problem balancing \emph{task faithfulness} and \emph{minimality}. However, as discussed in Section~\ref{subsec:circuit_properties}, circuits are further expected to be \emph{complete}, meaning they should include all nodes involved in the model's computations for a given task. We emphasize that our current approach does not guarantee \emph{completeness}. While some methods attempt to evaluate completeness by analyzing circuit behavior under ablations \citep{wang2023interpretability}, these approaches are computationally expensive, especially when dealing with large circuit sizes. Inspired by recent studies~\citep{yu2024functional, chen2025rethinking}, future research could explore more efficient ways to quantify completeness and incorporate this objective into the optimization framework described in Equation~\ref{eq:final_loss}.

\paragraph{Impact of ablation paradigm.} We observe that the circuits identified are influenced by the ablation value $\tilde{\vz}$ used during the identification process. For example, when comparing circuits based on mean versus zero ablation, we find notable differences in the nodes identified, even for zero-ablated circuits that demonstrate high task faithfulness (see results in Appendix~\ref{appendix:subsec:zero_ablation}, specifically Figure~\ref{appendix:figure:ablation-value-overlap}). Essentially, the behavior of a circuit is determined not merely by what it includes but also by what it excludes, as similarly noted by \citet{miller2024transformer}. We believe this is a general characteristic of methods that rely on constant perturbations, such as zero and mean ablations \citep{olsson2022context, wang2023interpretability}.

\paragraph{Generalization of findings.} It is important to note that our study focuses on a small encoder-decoder model trained on a single dataset like PCFG SET. Naturally, the question arises as to whether these findings will generalize to larger models and more diverse tasks. This touches on a broader concern within the field of mechanistic interpretability, where many studies are conducted on smaller models and datasets \citep{elhage2022toy}. We argue that, for our purposes, a controlled dataset like PCFG SET is well-suited to study modularity in language models. Nonetheless, the method we propose in Section~\ref{sec:activation_pruning} is easily extendable to other models and tasks. We believe that scaling our findings to larger models and more complex tasks represents a compelling area for future research.

\section*{Acknowledgments}
We would like to thank the members of the MaiNLP lab for their valuable feedback, with special recognition to Robert Litschko, Michael Hedderich, Florian Echin, and Shijia Zhou. Furthermore, we would like to express our gratitude to Lucas Georges Gabriel Charpentier at LTG for valuable feedback. Our appreciation extends to the anonymous reviewers for their comments and suggestions. We further acknowledge Sigma2, Norway, for providing access to the LUMI supercomputer, part of the EuroHPC Joint Undertaking, hosted by CSC (Finland) and the LUMI consortium. Lastly, we acknowledge the support provided to BP through the ERC Consolidator Grant DIALECT 101043235. For parts of the project, SW acknowledges funding for a research stay at MaiNLP at LMU, funded by Integreat, the Norwegian Centre for Knowledge-driven Machine Learning.

\bibliography{custom}

\appendix

\section{Reproducibility Statement}
\label{appendix:reproducability_statement}
To ensure the reproducibility of our experiments, we make all code publicly available at:~\href{https://github.com/mainlp/circuit-compositions}{https://github.com/mainlp/circuit-compositions}. Details of the training process, including the computational setup, model implementation, and hyperparameter selection, are thoroughly documented in Section~\ref{subsubsec:training} and Appendix~\ref{appendix:section:experimental-details}. Similarly, a detailed account of the evaluation procedure can be found in Section~\ref{subsubsec:evaluation} and Appendix~\ref{appendix:section:mask-evaluation}. Furthermore, all data used in this work is either publicly available or accompanied by a detailed description of the data generation process in Appendix~\ref{appendix:section:dataset}.

\begin{table*}[tbp]
    \centering
    \small
    \begin{tabular}{@{}llll@{}}
    \toprule
    Unary operation & Input & Output & Example\\
    \midrule
    \texttt{copy} & $x_1 \dots x_n$ & $x_1 \dots x_n$  & \texttt{copy K1 Y1 W1 K1}  $\rightarrow$ \texttt{K1 Y1 W1 K1}\\
    \texttt{echo} & $x_1 \dots x_n$ & $x_1 \dots x_n x_n$ & \texttt{echo E1 K1 A1 X1 J}  $\rightarrow$ \texttt{E1 K1 A1 X1 J1 J1} \\
    \texttt{repeat} & $x_1 \dots x_n$ & $x_1 \dots x_n x_1 \dots x_n$ & \texttt{repeat J1 F1 S1}  $\rightarrow$ \texttt{J1 F1 S1 J1 F1 S1} \\
    \texttt{reverse} & $x_1 \dots x_n$ & $x_n \dots x_1$ & \texttt{reverse G1 T1 X1 J1}  $\rightarrow$ \texttt{J1 X1 T1 G1} \\
    \texttt{swap} & $x_1 \dots x_n$ & $x_n x_2 \dots x_{n-1} x_1$ & \texttt{swap B1 Z1 V1 I1 W1}  $\rightarrow$ \texttt{W1 Z1 V1 I1 B1} \\
    \texttt{shift} & $x_1 \dots x_n$ & $x_2 \dots x_n x_1$ & \texttt{shift Y1 I1 D1 H1 K1}  $\rightarrow$ \texttt{I1 D1 H1 K1 Y1} \\
    \midrule
    Binary operation & Input & Output & Example\\
    \midrule
    \texttt{append} & $x, y$ & $x \, y$  & \texttt{append F1 B1, U1 A1 G1}  $\rightarrow$ \texttt{F1 B1 U1 A1 G1} \\
    \texttt{prepend} & $x, y$  & $y \, x$  & \texttt{prepend F1 B1, U1 A1 G1}  $\rightarrow$ \texttt{U1 A1 G1 F1 B1} \\
    \texttt{remove\_first} & $x, y$  & $y$ & \texttt{remove\_first Z1 P1 N1, A1 D1}  $\rightarrow$ \texttt{A1 D1} \\
    \texttt{remove\_second} & $x, y$  & $x$ & \texttt{remove\_second F1 B1, U1 A1 G1}  $\rightarrow$ \texttt{F1 B1} \\
    \bottomrule
    \end{tabular}
    \caption{The string-edit operations in PCFG SET from \citet{hupkes2020compositionality}.}
    \label{appendix:tab:set_operations_examples}
\end{table*}

\section{Dataset} \label{appendix:section:dataset}
\subsection{PCFG SET}\label{appendix:section:dataset:examples}
\citet{hupkes2020compositionality} construct PCFG SET in such a way that compositionality is a salient feature of the dataset, while aligning its statistical properties with those of natural language corpora, specifically English. We present examples for each string-edit operation in Table~\ref{appendix:tab:set_operations_examples} to help the reader get more familiar with the functions employed in PCFG SET \citet{hupkes2020compositionality}.

The PCFG SET dataset is released under a MIT license and used in accordance to the licensing agreements.

\subsection{Generating Isolated Function Data}\label{appendix:section:dataset:generating-isolated-function-data}
We use the probabilistic context-free grammar proposed by~\citet{hupkes2020compositionality} to generate datasets for each of the string-edit operations originally used to construct PCFG SET. Using the same vocabulary, consisting of letters $[A-Z]$ combined with numbers in $(0,\infty] \in \mathcal{Z}^+$, we restrict the grammar to produce samples that use only specific string-edit operation for each subtask dataset, e.g, \texttt{copy W1 O1 Z5 G1}. For each SET operation, we generate 16\,000 samples for training and 4\,000 for validation. Each subtask dataset includes samples of both single-use and composed applications of that function, such as $\texttt{remove\_first}\, \texttt{remove\_first}\, A1\, \, , B1 \, , \,  B1\, A1 \rightarrow B1\, A1$.

\section{Experimental Details}\label{appendix:section:experimental-details}
In this section, we outline key details regarding our experiments, such as our computational setup and the hyperparameters selected.

\subsection{Software and Computing} \label{appendix:section:computing}
All experiments are implemented using PyTorch \citep{paszke2019pytorch} as the primary framework. Detailed information about supporting software and specific versions can be found in our code repository in the supplementary materials. Experiments were run on AMD Instinct MI250X and NVIDIA A100 GPUs.

\subsubsection{Computational Budget}
For training base models, we consumed approximately 12 GPU hours in total. For the mask training, including hyperparameter selection, failed runs, and debugging, we consumed approximately 4700 GPU hours.

\subsection{Base Model Architecture}\label{appendix:section:base-model-architecture}

The base model $\mathcal{M}$ is a transformer-based encoder-decoder, as proposed by \citet{Vaswani@2017attention}, but with the Gated Linear Unit (GLU) variant from \citet{shazeer2020glu}. We use sinusoidal positional encoding, delimit tokens on whitespace, and disjoint embedding matrices for the input and output sequences. The base model $\mathcal{M}$ consists of six encoder and six decoder layers, with a hidden size of dimension $512$, and eight attention heads per layer.

\subsection{Base Model Training}\label{appendix:section:base-model-training}

We train the base model $\mathcal{M}$ using a learning rate of $5 \cdot 10^{-7}$, a batch size of $64$ without gradient accumulation, a gradient clipping value of $15$, and a dropout rate of $0.2$. Furthermore, the model is trained using the original datasets from \citet{hupkes2020compositionality}. In Table~\ref{appendix:table:base-model-acc}, we report the accuracy of the base model on each of the isolated subtask datasets. The base model $\mathcal{M}$ achieves final accuracy of $87.8\%$ on the official PCFG test set, which is slightly lower than the performance of $92.0\%$ reported by \citet{hupkes2020compositionality}.

\begin{table}[tbp]
    \centering
    \small
    \begin{tabular}{@{}c@{\hskip 10pt}lll@{}}
        \toprule
        & Operation & Accuracy \\
        \midrule
        \multirow{6}{*}{\rotatebox[origin=c]{90}{Unary}} 
        & \texttt{copy} & 1.00 \\
        & \texttt{echo} & 0.999 \\
        & \texttt{repeat} & 0.943 \\
        & \texttt{reverse} &  0.989 \\
        & \texttt{swap} & 0.996 \\
        & \texttt{shift} & 0.992 \\
        \midrule
        \multirow{4}{*}{\rotatebox[origin=c]{90}{Binary}} 
        & \texttt{append} & 0.848 \\
        & \texttt{prepend} & 0.832 \\
        & \texttt{remove\_first} & 0.975 \\
        & \texttt{remove\_second} & 0.999 \\
        \bottomrule
    \end{tabular}
    \caption{The generation accuracy of the base model on the different PCFG SET subtasks~\citep{hupkes2020compositionality}.}
    \label{appendix:table:base-model-acc}
\end{table}

\subsection{Mask Training}\label{appendix:section:mask-training}
In this study, we conduct experiments for both mean and zero ablation. Following the approach of \citet{wang2023interpretability}, the mean ablation value $\tilde{\vz}$ is derived from a reference distribution. Specifically, when training a mask for task $T_l$ on dataset $D_l$, we assign the ablated value $\tilde{\vz}_i$ for each potential mediator $\vz_i$ in $\mathcal{M}$ as the average activation of $\vz_i$ across samples in $D_l$, which only includes samples corresponding to task $T_l$. For instance, when training a mask for the \texttt{copy} task, the ablated value $\tilde{\vz}_i$ for each mediator $\vz_i$ is set to the average activation of $\vz_i$ over samples exclusively containing the \texttt{copy} function, i.e., $D_{\texttt{copy}}^{\text{train}}$, averaged across all token positions.

The final optimization problem, described in Equation~\ref{eq:final_loss}, requires the predicted output probabilities $\vy^{\mathcal{M}}$ of the base model $\mathcal{M}$ for each dataset sample across all of the subtask datasets. In practice, this is achieved through caching: a forward pass is performed on all datasets using $\mathcal{M}$, and $\vy^{\mathcal{M}}$ is stored for later use during mask training.

\subsection{Method Validation with \textsc{Tracr}}\label{appendix:section:tracr_validation}
The \textsc{Tracr} models employ a decoder-only architecture with bidirectional attention. Table \ref{appendix:table:tracr-sizes} summarizes the model configurations for each task. All models execute their respective RASP functions on sequences of length four. For mask training, we use a learning rate of 0.001, maximum temperature $\beta_{max} = 200$, $\lambda = 1 \cdot 10^{-4}$, and an initial mask value of $\vs_{\text{initial}} = 1.0$. We train 50 epochs for \texttt{copy}, \texttt{reverse}, and \texttt{swap}, and 200 epochs for \texttt{echo}.

\begin{figure*}
    \centering
    \begin{subfigure}[t]{0.45\textwidth}
        \centering
        \includegraphics[width=0.6\linewidth]{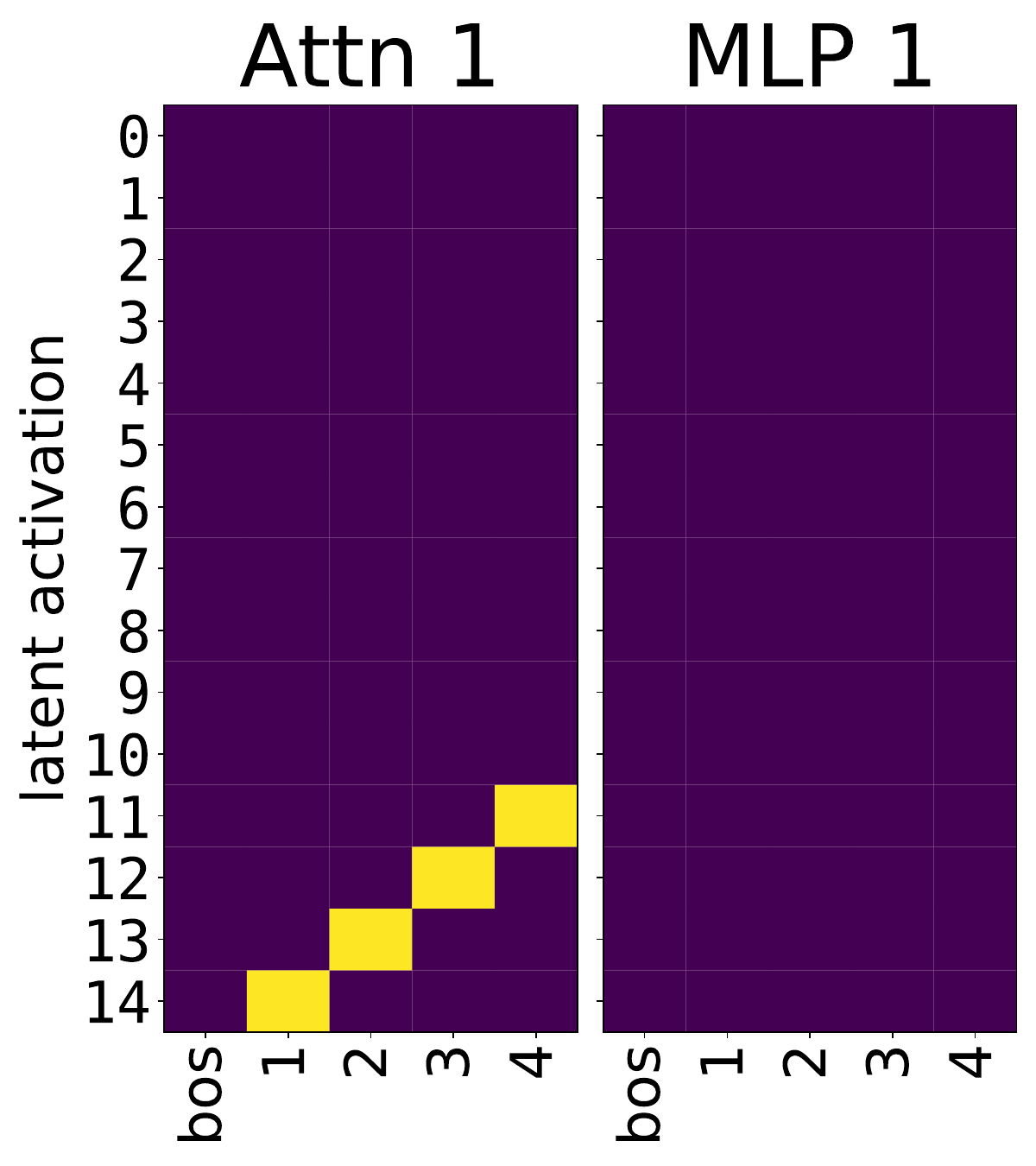}
        \caption{The active nodes extracted from the compiled \texttt{copy} \textsc{Tracr} program.}
    \end{subfigure}
    \hfill
    \begin{subfigure}[t]{0.45\textwidth}
        \centering
        \input{figures/tikz/tracr/tracr_local_sparsity_copy}
        \caption{The local sparsity of the \texttt{copy} \textsc{Tracr} circuit.}
    \end{subfigure}
    \par\bigskip
    \begin{subfigure}[t]{0.45\textwidth}
        \centering
        \includegraphics[width=0.8\linewidth]{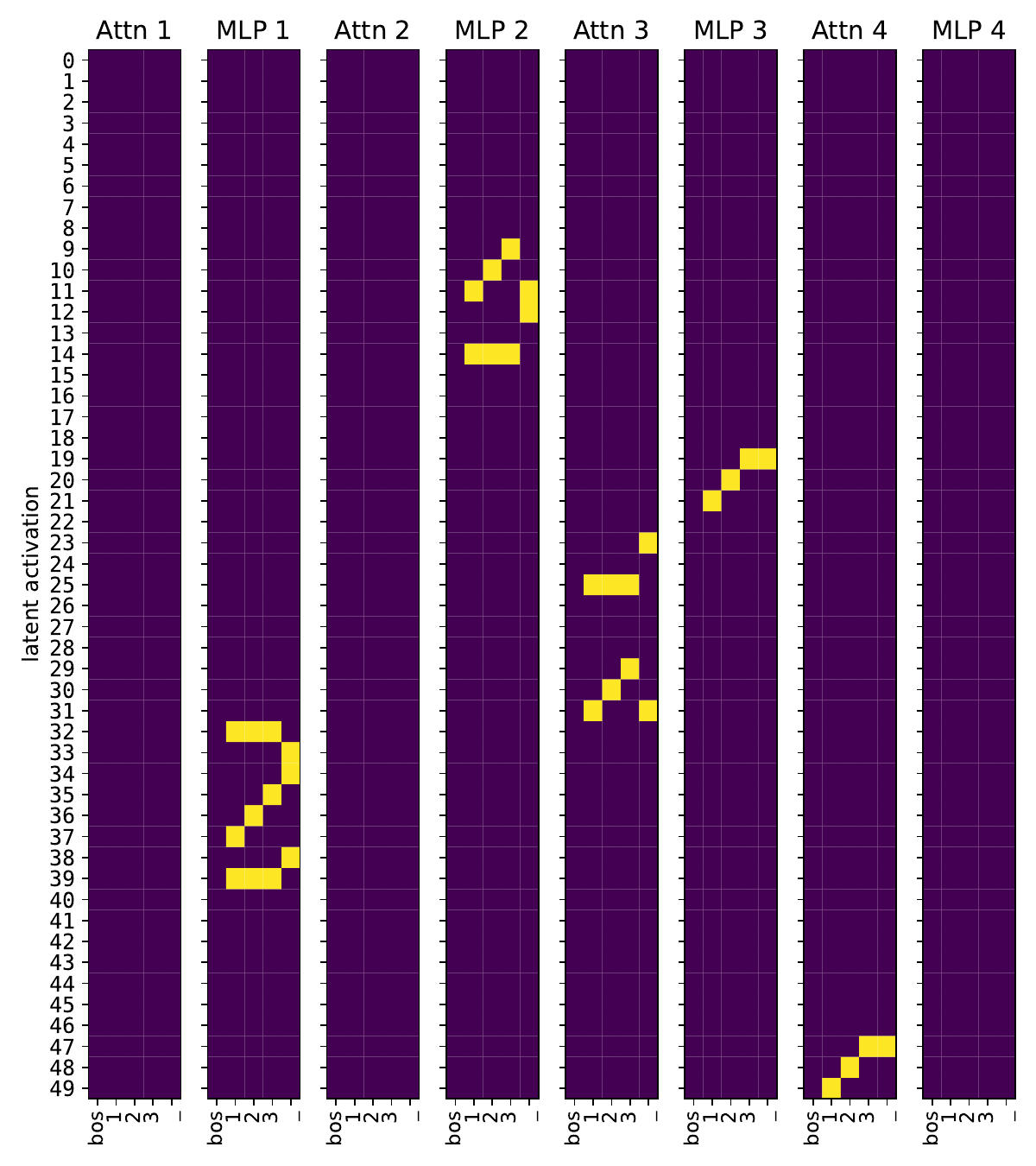}
        \caption{The active nodes extracted from the compiled \texttt{echo} \textsc{Tracr} program.}
    \end{subfigure}
    \hfill
    \begin{subfigure}[t]{0.45\textwidth}
        \centering
        \input{figures/tikz/tracr/tracr_local_sparsity_echo}
        \caption{The local sparsity of the \texttt{echo} \textsc{Tracr} circuit.}
    \end{subfigure}
    \caption{The activation patterns from the compiled \texttt{copy} and \texttt{echo} \textsc{Tracr} programs along the corresponding circuits we identify. We display the fraction of the circuit's remaining activations for each layer and module.}
    \label{appendix:figure:local-sparsity-tracr}
\end{figure*}

\begin{figure*}
    \centering
    \begin{subfigure}[t]{0.45\textwidth}
        \centering
        \includegraphics[width=0.8\linewidth]{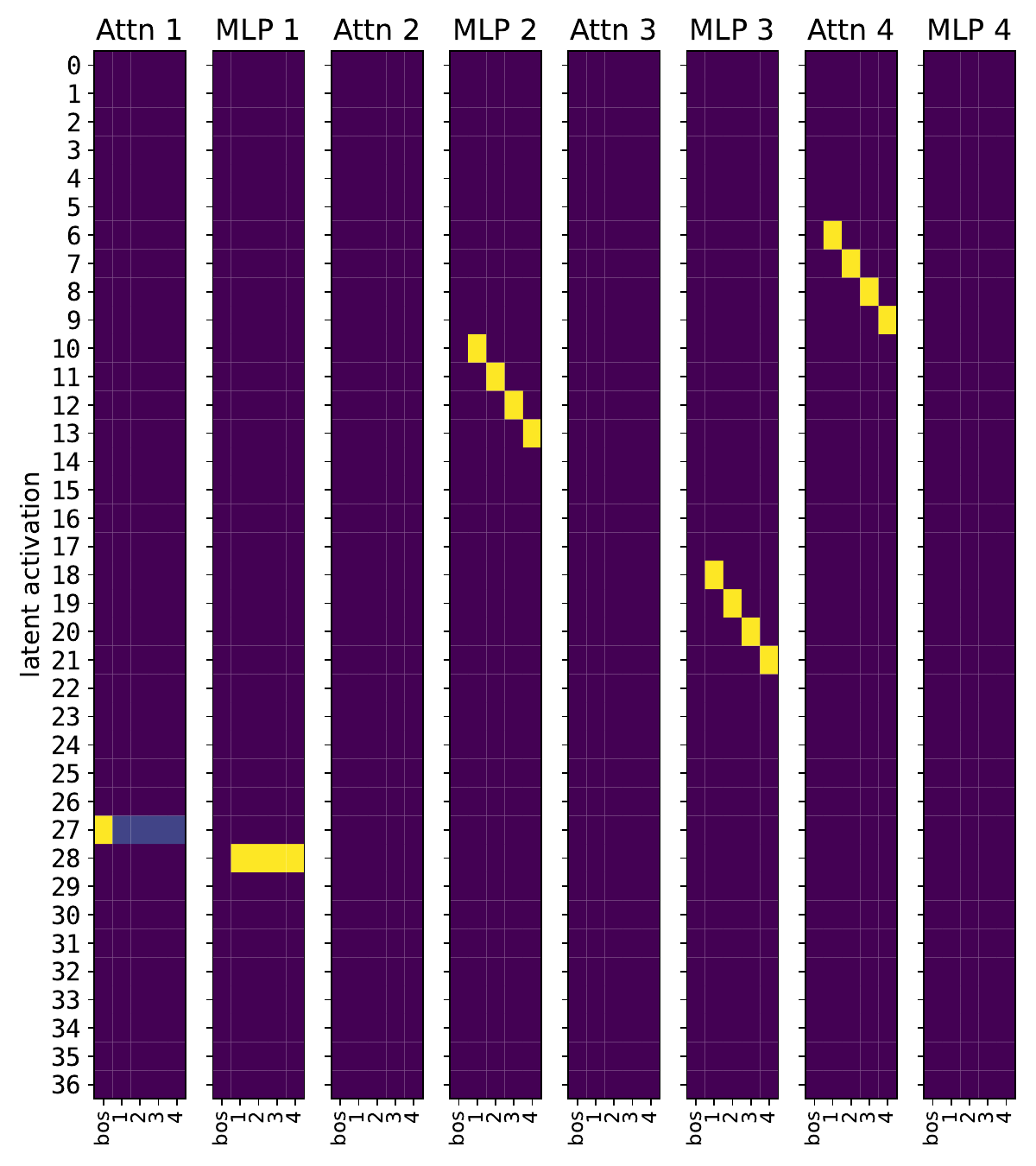}
        \caption{The active nodes extracted from the compiled \texttt{reverse} \textsc{Tracr} program.}
    \end{subfigure}
    \hfill
    \begin{subfigure}[t]{0.45\textwidth}
        \centering
        \input{figures/tikz/tracr/tracr_local_sparsity_reverse}
        \caption{The local sparsity of the \texttt{reverse} \textsc{Tracr} circuit.}
    \end{subfigure}
    \par\bigskip
    \begin{subfigure}[t]{0.45\textwidth}
        \centering
        \includegraphics[width=0.8\linewidth]{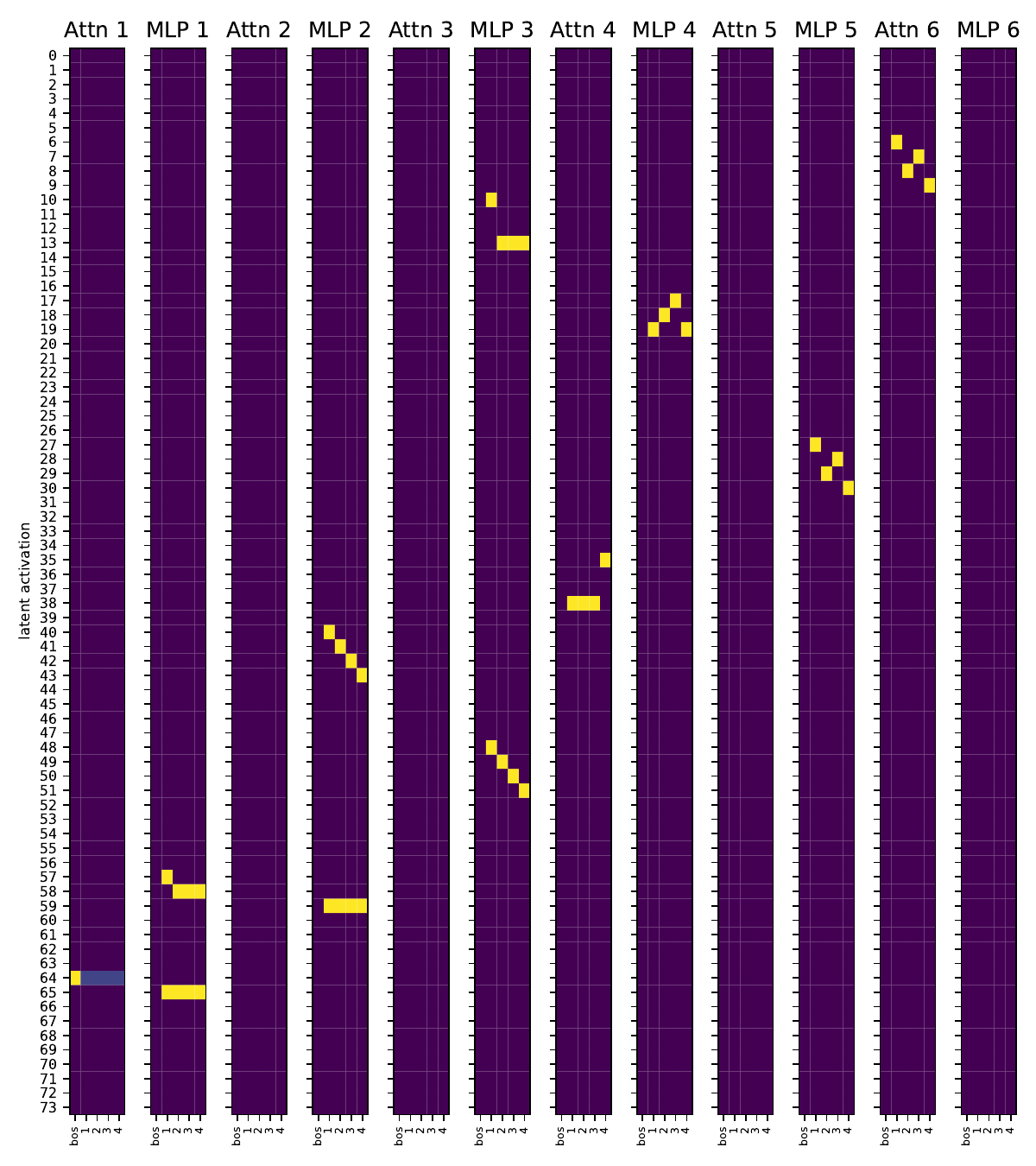}
        \caption{The active nodes extracted from the compiled \texttt{swap} \textsc{Tracr} program.}
    \end{subfigure}
    \hfill
    \begin{subfigure}[t]{0.45\textwidth}
        \centering
        \input{figures/tikz/tracr/tracr_local_sparsity_swap}
        \caption{The local sparsity of the \texttt{swap} \textsc{Tracr} circuit.}
    \end{subfigure}
    \caption{The activation patterns from the compiled \texttt{reverse} and \texttt{swap} \textsc{Tracr} programs along the corresponding circuits we identify. We display the fraction of the circuit's remaining activations for each layer and module. We note that for these two circuits, we prune the first attention module because we do not consider the BOS token during our pruning, which is the only activated position in this layer.}
    \label{appendix:figure:local-sparsity-tracr-2}
\end{figure*}

In Figures \ref{appendix:figure:local-sparsity-tracr} and \ref{appendix:figure:local-sparsity-tracr-2}, we report the sparsity of the pruned circuits along with their respective ground truth circuit in the compiled model. We note that in our experiments, we do not consider the beginning-of-sequence (BOS) token position. Thus, for layers that contain an active neuron only for the BOS token, we prune the whole component. For \texttt{copy}, \texttt{reverse}, and \texttt{swap}, our method recovers all individual neurons in the compiled model. For these tasks, the compiled model achieves a 100\% accuracy on the target task. For \texttt{echo}, our RASP implementation achieves 42\% accuracy when compiled, which our method also recovers with full faithfulness. This shows that our method finds faithful circuits even when the base model's performance is not perfect.

\begin{table}[tbp]
\small
\centering
\begin{tabular}{lcccc}
\toprule
Circuit & Hidden size & QKV size & Layers & Heads \\
\midrule
\texttt{copy} & 15 & 6 & 1 & 1 \\
\texttt{reverse} & 37 & 10 & 4 & 1 \\
\texttt{echo} & 46 & 9 & 4 & 2 \\
\texttt{swap} & 74 & 13 & 6 & 1 \\
\bottomrule
\end{tabular}
\caption{The model configurations of the \textsc{Tracr} programs.}
\label{appendix:table:tracr-sizes}
\end{table}

\begin{table}[b!]
\centering
\small
\begin{tabular}{@{}lcccccc@{}}
\toprule
Circuit & lr &  $\lambda$ & $\vs_{\text{initial}}$ & $\beta_{max}$ & Epochs \\
\midrule
\texttt{echo} & $10^{-4}$ & $10^{-4}$  & 0.05 & 200 &  500 \\
\texttt{copy} & $10^{-4}$ & $10^{-4}$ & 0.05  & 200 &  500 \\
\texttt{repeat} & $10^{-4}$  & $10^{-4}$  & 0.05  & 200 & 500 \\
\texttt{reverse} & $10^{-4}$ & $10^{-4}$  & 0.05 & 200 &  500\\
\texttt{swap} & $10^{-4}$ & $10^{-4}$  & 0.05 & 200 &  500 \\
\texttt{shift} & $10^{-4}$  & $10^{-4}$  & 0.05 & 200 &  500 \\
\texttt{append} & $10^{-3}$ & $10^{-5}$ & 0 & 200  &  500 \\
\texttt{prepend} & $10^{-3}$ & $10^{-5}$ & 0 & 100 &  500 \\
\texttt{rm\_first} & $10^{-4}$ & $10^{-5}$  & 0.05 & 300 & 500  \\
\texttt{rm\_second} & $10^{-3}$ & $10^{-5}$  & 0.05 & 100 & 300 \\
\bottomrule
\end{tabular}
\caption{Final hyperparameters used for mask training.}
\label{appendix:table:circuit-parameters}
\end{table}

\subsubsection{Hyperparameters}\label{appendix:section:mask-training-hyperparameters}
In Table~\ref{appendix:table:circuit-parameters}, we report the final hyperparameters for the mask training for each subtask, using the same notation as in Section~\ref{sec:activation_pruning}. These were selected based on a random search over the following hyperparameters: learning rate: \,\{$1\cdot10^{-3}$, $1\cdot10^{-4}$, $1\cdot10^{-5}$\}, \, $\lambda$: \, \{$1\cdot10^{-3}$, $1\cdot10^{-4}$, $1\cdot10^{-5}$\},\, $\vs_{\text{initial}}$:\, \{0.2, 0.05, 0\},\, $\beta_{max}$: \, \{100, 200, 300\}.

\begin{table*}[tbp]
\centering
\small
\begin{minipage}[t]{0.49\textwidth}
\centering
\begin{tabular}{lccc}
\toprule
$\lambda$ & \multicolumn{2}{c}{Faithfulness} & Sparsity \\
\cmidrule(lr){2-3}
 & $F_T$ & $D_{KL}$ & \\
\midrule
$1\times10^{-2}$ & 0.228 & 4.180 & 0.022 \\
$1\times10^{-3}$ & 0.833 & 0.687 & 0.105 \\
$1\times10^{-4}$ & 0.943 & 0.120 & 0.133 \\
$1\times10^{-5}$ & 0.992 & 0.015 & 0.162 \\
$1\times10^{-6}$ & 0.999 & 0.001 & 0.298 \\
\bottomrule
\end{tabular}
\end{minipage}
\hfill
\begin{minipage}[t]{0.49\textwidth}
\centering
\begin{tabular}{lccc}
\toprule
$\lambda$ & \multicolumn{2}{c}{Faithfulness} & Sparsity \\
\cmidrule(lr){2-3}
 & $F_T$ & $D_{KL}$ & \\
\midrule
$1\times10^{-2}$ & 0.106 & 4.478 & 0.017 \\
$1\times10^{-3}$ & 0.734 & 0.737 & 0.135 \\
$1\times10^{-4}$ & 0.951 & 0.121 & 0.199 \\
$1\times10^{-5}$ & 0.982 & 0.026 & 0.324 \\
$1\times10^{-6}$ & 0.994 & 0.013 & 0.648 \\
\bottomrule
\end{tabular}
\end{minipage}
\caption{Effect of the sparsity regularization for different values of $\lambda$ on the \texttt{copy} (left) and \texttt{repeat} (right) tasks.}
\label{tab:lambda_sensitivity_analysis}
\end{table*}

\subsection{Evaluation}\label{appendix:section:mask-evaluation}
We evaluate circuits according to two primary criteria, performance and node overlap, as described in Section~\ref{subsubsec:evaluation}. Specifically, we quantify the performance of a circuit by its \textit{faithfulness} score, which is calculated via the KL divergence, $\operatorname{D}_{KL}$, between the output distribution of the circuit $\vy^{\vm}$ and the base model $\vy^{\mathcal{M}}$. However, to get a bounded metric, we instead use the normalized version of the Jensen-Shannon divergence between the two output distributions:

\begin{align}
     &\operatorname{JSD}_{\text{norm}}\left(\vy^{\vm} \parallel \vy^{\mathcal{M}}\right) = \frac{1}{2\log(2)} \big( \label{eq:jsd-faithfulness} \\
    &D_{KL}(\vy^{\vm} \parallel \vy^{\vm\mathcal{M}}) + D_{KL}(\vy^{\mathcal{M}} \parallel \vy^{\vm\mathcal{M}}) \notag \\
    &{\big), \quad \text{where} \quad \vy^{\vm\mathcal{M}} = \frac{\vy^{\vm} + \vy^{\mathcal{M}}}{2} \notag}
\end{align}

where $\vy^{\vm\mathcal{M}}$ is a mixture distribution of $\vy^{\vm}$ and $\vy^{\mathcal{M}}$. Note that the Jensen-Shannon divergence is symmetric. Furthermore, it is bounded, i.e., $0 \leq \operatorname{JSD}_{\text{norm}} \leq 1$.

To measure the circuits' node overlap we compute their Intersection over Union (IoU) and Intersection over Minimum (IoM). Given two circuits $\vm^{T_1} \in \{0, 1\}^N$ and $\vm^{T_2} \in \{0, 1\}^N$ defined over the same node space, the IoU and IoM are computed as follows:

\begin{align}
    \operatorname{IoU} \label{appendix:equation:iou_iom}  &= \frac{\|\vm^{T_1} \cap \vm^{T_2}\|_1}
    {\|\vm^{T_1} \cup \vm^{T_2}\|_1} \\
    \operatorname{IoM} &= \frac{\|\vm^{T_1} \cap \vm^{T_2}\|_1}
    {\min\left(\| \vm^{T_1} \|_1, \| \vm^{T_2} \|_1\right)} \notag
\end{align}

where $\vm^{T_1} \cap \vm^{T_2}$ represents the intersection of the two binary masks, while $\vm^{T_1} \cup \vm^{T_2}$ denotes their union. Specifically, the intersection between two circuits $\vm^{T_1, T_2} = \vm^{T_1} \cup \vm^{T_2}$ is defined as:

\begin{align}
    \evm^{T_1, T_2}_i &= \evm^{T_1}_i \cap \evm^{T_2}_i 
    = \evm^{T_1}_i \land \evm^{T_2}_i \notag \\
    &= \begin{cases}
        1, &\text{if} \quad \evm^{T_1}_i = 1 \;\text{and}\; \evm^{T_2}_i = 1, \\
        0, &\text{otherwise.}
    \end{cases}
\end{align}

Similarly, the union between two circuits $\vm^{T_1, T_2} = \vm^{T_1} \cup \vm^{T_2}$ can be computed as:

\begin{align}
    \evm^{T_1, T_2}_i &= \evm^{T_1}_i \cup \evm^{T_2}_i 
    = \evm^{T_1}_i \lor \evm^{T_2}_i \notag \\
    &= \begin{cases}
        0, &\text{if} \quad \evm^{T_1}_i = 0 \;\text{and}\; \evm^{T_2}_i = 0, \\
        1, &\text{otherwise.}
    \end{cases}
    \label{appendix:eq:union}
\end{align}

\subsection{Subnetwork Compositions }\label{appendix:section:compositions}
When creating circuit compositions, we apply the union operator between the binary masks of two circuits, as illustrated in Equation~\ref{appendix:eq:union}. For zero-ablated circuits, this operation remains symmetric. However, in the case of mean-ablated circuits, symmetry is not preserved. This is primarily due to the fact that the mediator value, $\vz_i$, for which $\evm_i = 0$ after the union, is ablated by the mean value across the reference distribution associated with the first circuit's task, $\tilde{\vz}^{T_1}_i$. Specifically, for the union $\vm^{T_1, T_2} = \vm^{T_1} \cup \vm^{T_2}$, the mediator value $\vz_i$ is determined as follows:

\begin{equation}\label{appendix:eq:union-mean}
    \vz_i = \begin{cases}
    \tilde{\vz}^{T_1}_i &\text{if} \quad \evm^{T_1}_i=0 \; \text{and} \; \evm^{T_2}_i=0,\\
    \vz_{x_i} &\text{otherwise,}
    \end{cases}
\end{equation}

This means that $\vm^{T_1} \cup \vm^{T_2} \neq \vm^{T_2} \cup \vm^{T_1}$ for mean ablations if $\tilde{\vz}^{T_1}_i \neq \tilde{\vz}^{T_2}_i$. For zero ablation, the ablation value remains the same for both cases $\tilde{\vz}^{T_1}_i = \tilde{\vz}^{T_2}_i = 0$, thus yielding symmetry of the operator.

\section{Supplemental Results}\label{appendix:section:supplemental-results}
In this section, we report supplemental results that complement the paper's key findings.

\subsection{Mean Ablation}
We first report additional results related to circuits identified via mean ablation.

\subsubsection{Circuit Performance}
\begin{figure*}[tbp]
    \centering
    \begin{subfigure}[b]{0.40\textwidth}
        \centering
        \includegraphics[width=\textwidth]{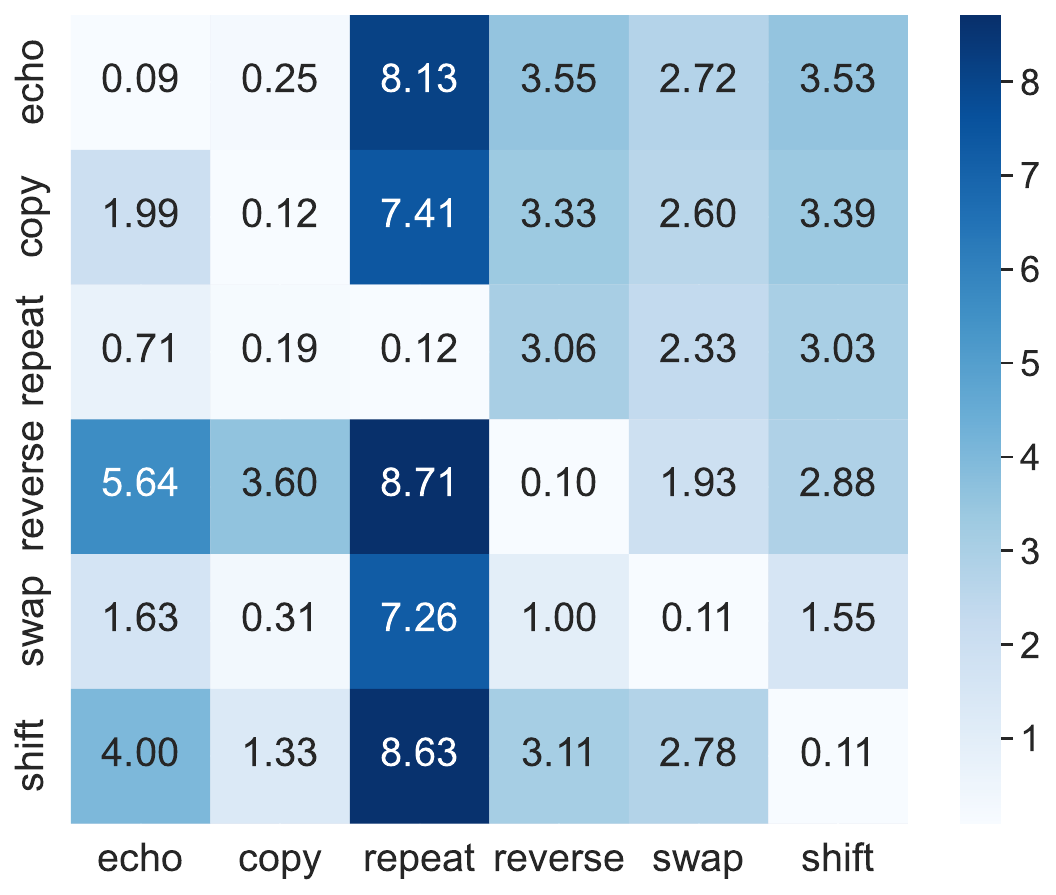}
        \caption{All token positions.}
        \label{appendix:figure:mean-kl-unary}
        \end{subfigure}
    \hfill
    \begin{subfigure}[b]{0.40\textwidth}
        \centering
        \includegraphics[width=\textwidth]{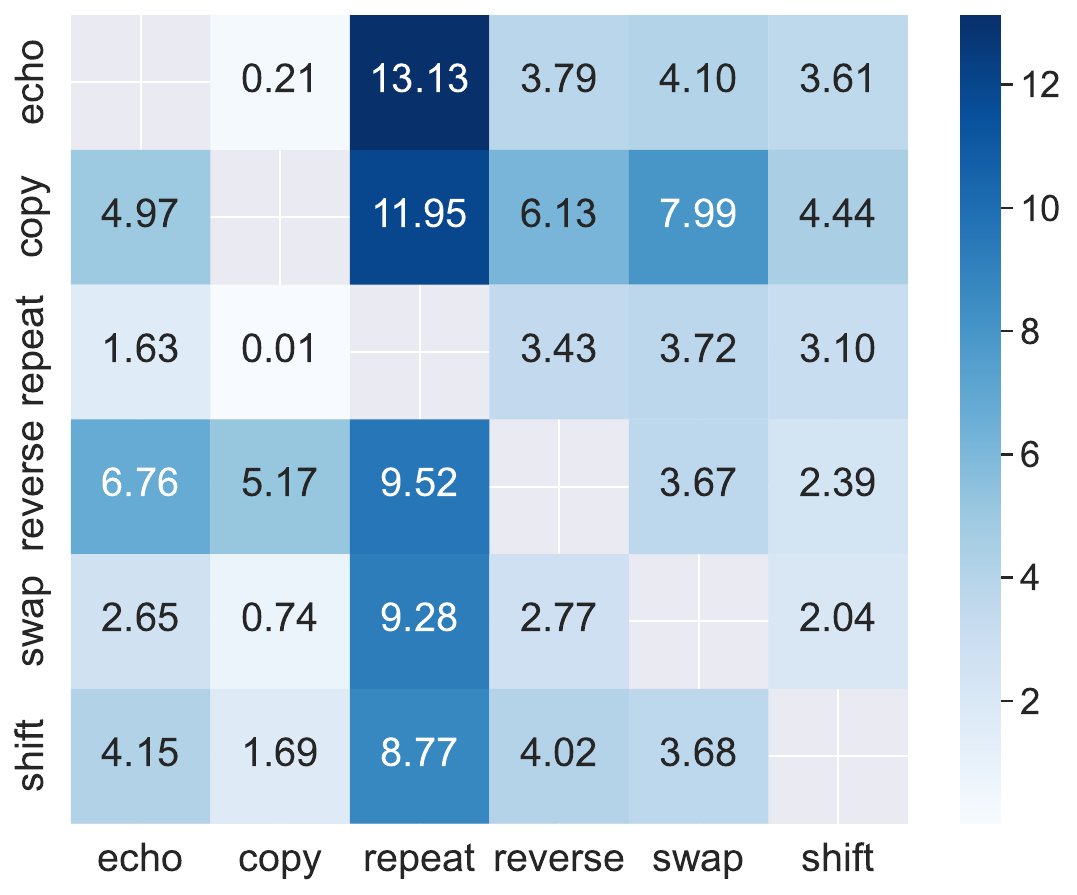}
        \caption{Differing token positions.}
        \label{appendix:figure:mean-kl-unary-diff}
    \end{subfigure}
    \caption{Task faithfulness measured via $\operatorname{D}_{KL}\left(\vy^{\vm} \parallel \vy^{\mathcal{M}}\right)$ for the \textbf{mean-ablated unary} circuits. The y-axis corresponds to the circuit, while the x-axis represents the evaluation task. When we only evaluate selected positions, we omit the diagonal, as there are no applicable tokens for comparison.}
    \label{appendix:figure:mean-ablaton-kl-divergence-scores}
\end{figure*}

\begin{figure*}[tbp]
    \centering
    \begin{subfigure}[b]{0.40\textwidth}
        \centering
        \includegraphics[width=\textwidth]{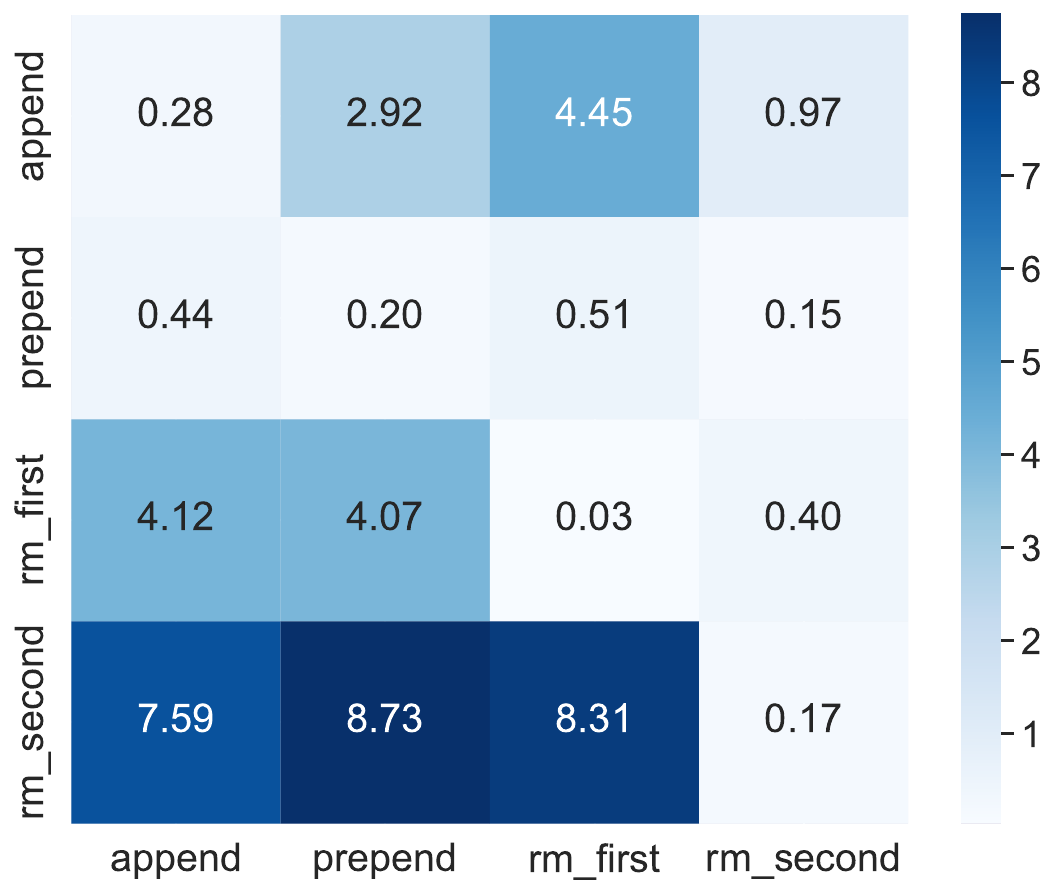}
        \caption{All token positions.}
        \label{appendix:figure:mean-kl-binary}
    \end{subfigure}
    \hfill
    \begin{subfigure}[b]{0.40\textwidth}
        \centering
        \includegraphics[width=\textwidth]{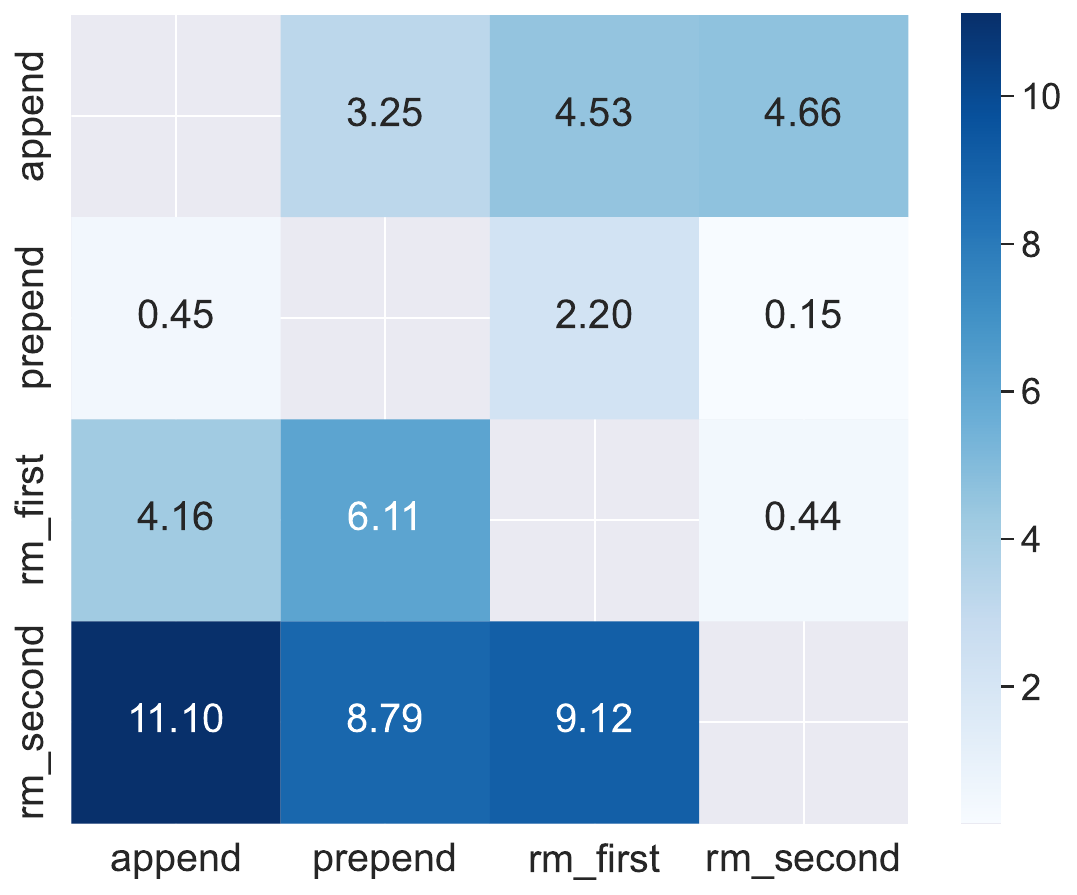}
        \caption{Differing token positions.}
        \label{appendix:figure:mean-kl-binary-diff}
    \end{subfigure}
    \caption{Task faithfulness measured via $\operatorname{D}_{KL}\left(\vy^{\vm} \parallel \vy^{\mathcal{M}}\right)$ for the \textbf{mean-ablated binary} circuits. The y-axis corresponds to the circuit, while the x-axis represents the evaluation task. When we only evaluate selected positions, we omit the diagonal, as there are no applicable tokens for comparison.}
    \label{appendix:figure:mean-ablaton-kl-divergence-scores-diff}
\end{figure*}

In Figure~\ref{appendix:figure:mean-ablaton-kl-divergence-scores} and Figure~\ref{appendix:figure:mean-ablaton-kl-divergence-scores-diff} we report the KL divergence between the predicted output distributions of the mean-ablated circuits, $\vm$, and the base model, $\mathcal{M}$, for both unary and binary string-edit operations. The results exhibit a trend consistent with the observations discussed in Section~\ref{subsubsec:circuit_performance}.

\subsubsection{Local Sparsity}\label{appendix:section:circuit-sparsity}
Figure~\ref{appendix:figure:local-sparsity-unary} illustrates the local sparsity of each unary circuit identified via mean-ablations, while \ref{appendix:figure:local-sparsity-binary} presents the local sparsity of each binary circuit, respectively. It is evident that unary circuits are highly sparse, with no module retaining more than 45\% of activations. Notably, the feed-forward (FF) and multi-head self-attention (MHSA) modules within the decoder demonstrate pronounced sparsity across all layers, typically retaining only $0.0\%$ to $10\%$ of activations. In contrast, the multi-head cross-attention (MHCA) modules retain a higher proportion of activations. As shown in Figure~\ref{appendix:figure:local-sparsity-binary}, binary circuits generally retain a greater number of activations. Specifically, the \texttt{append} and \texttt{prepend} circuits show substantial remaining activations in both the encoder and decoder.

\subsubsection{Lambda Sensitivity Analysis}\label{appendix:section:lambda-sensitivity}
To assess the impact of the hyperparameter $\lambda$ in Equation~\ref{eq:final_loss}, we perform a parameter sensitivity analysis for the two unary operations \texttt{copy} and \texttt{repeat}. Specifically, we consider six different $\lambda$ values: $\lambda \in {1\times10^{-2}, 1\times10^{-3}, 1\times10^{-4}, 1\times10^{-5}, 1\times10^{-6}}$, and train a mask for each, following the procedure outlined in Sections~\ref{sec:activation_pruning} and~\ref{sec:experiments}. We report task faithfulness $F_T$, KL-divergence $D_{KL}$ (as defined in Section~\ref{subsubsec:evaluation}), and the resulting circuit's sparsity for each task in Table~\ref{tab:lambda_sensitivity_analysis}.

As expected, we observe that the $\lambda$ parameter governs the trade-off between task faithfulness and circuit sparsity: higher values of $\lambda$ place greater emphasis on the regularization term in Equation~\ref{eq:final_binary_loss}, thereby promoting sparsity in the resulting circuit while sacrificing some faithfulness. In contrast, lower values of $\lambda$ result in less sparse but more faithful models. For instance, a value of $\lambda = 1 \times 10^{-6}$ yields a high faithfulness performance of $F_{\text{copy}} = 0.999$ on the copy task and a sparsity level of $0.298$, whereas a value of $\lambda = 1 \times 10^{-2}$ results in a lower faithfulness performance of $F_{\text{copy}} = 0.228$ and higher sparsity level of $0.022$.

\subsubsection{Deterministic Mask Approximation}\label{appendix:section:deterministic_approx}
As discussed in Section~\ref{sec:activation_pruning}, a key advantage of learning a binary mask through continuous sparsification is the deterministic nature of the approach. To assess whether our method consistently converges to the same circuits despite the stochastic elements of the training process (e.g., dataset shuffling), we examine the node overlap of circuits trained using different random seeds. Our findings demonstrate that the method reliably converges to the same mask for identical subtasks, regardless of the random seed. This holds true for both mean and zero ablation. For example, we report an IoU and IoM of $1.0$ when comparing five \texttt{prepend} circuits trained with varying random seeds.

\begin{figure}[b!]
    \centering
    \begin{subfigure}[b]{0.8\linewidth}
        \centering
        \includegraphics[width=\linewidth]{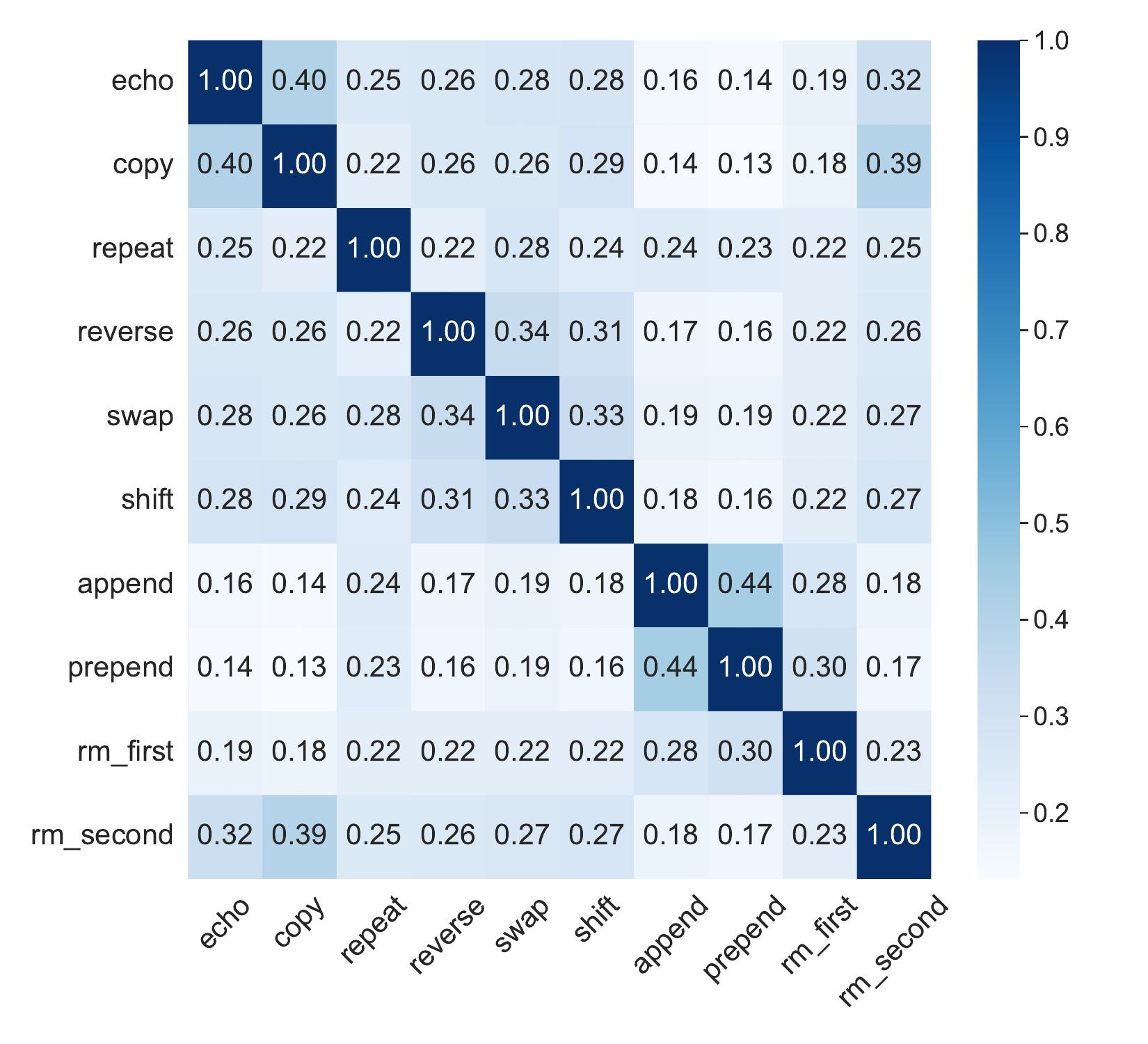}
        \caption{Intersection over Union (IoU).}
        \label{appendix:figure:cross-task-overlap-iou-zero}
        \end{subfigure}
    \vspace{0.5cm}
    \begin{subfigure}[b]{0.8\linewidth}
        \centering
        \includegraphics[width=\linewidth]{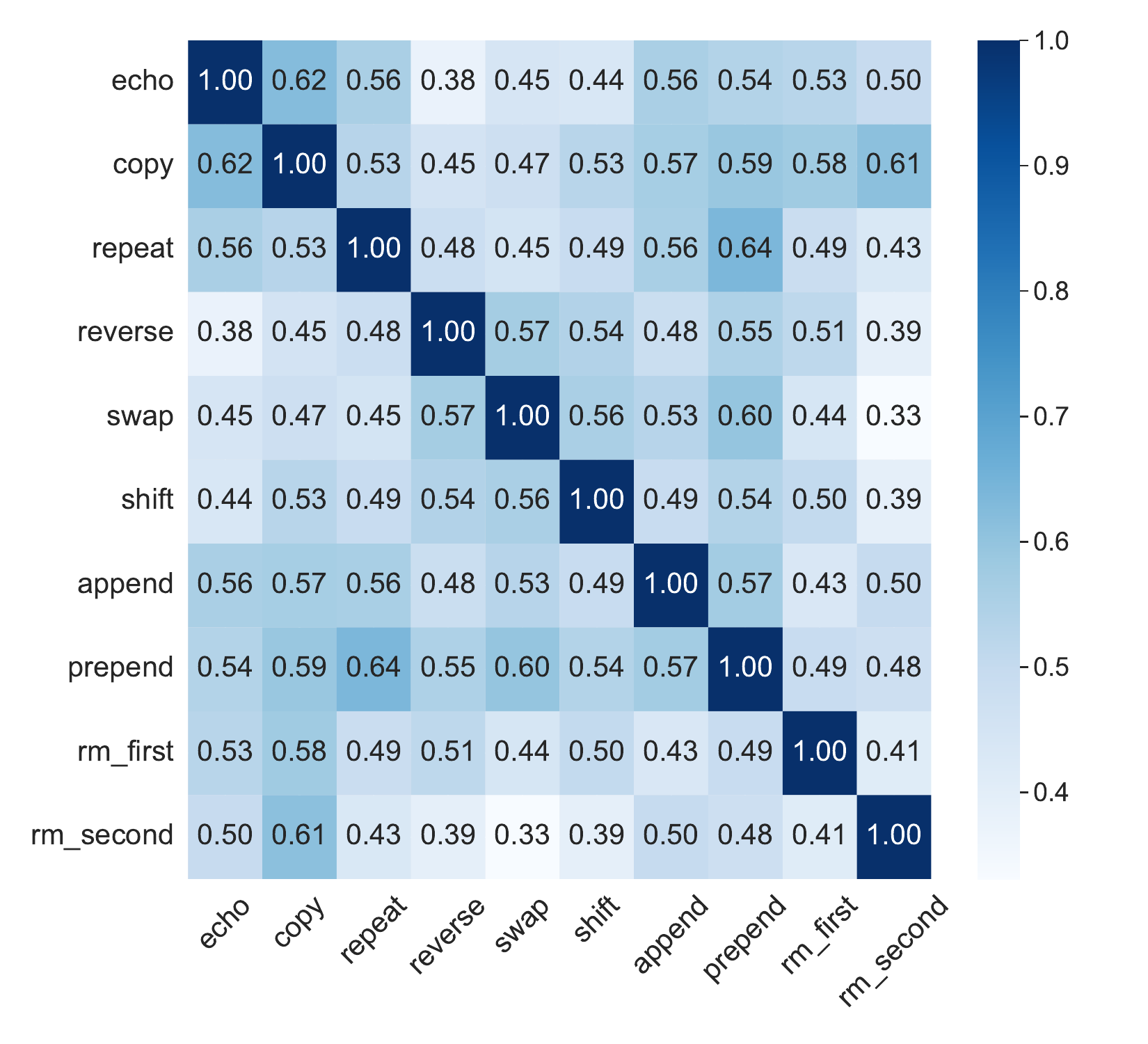}
        \caption{Intersection over Minimum (IoM).}
        \label{appendix:figure:cross-task-overlap-iom-zero}
    \end{subfigure}
    \caption{Node overlap for circuits identified via zero ablation.}
    \label{appendix:figure:cross-task-overlap-zero}
\end{figure}

\subsubsection{Set Operations}\label{appendix:section:set-operations}
\begin{figure*}[tbp]
    \centering
    \begin{subfigure}{\textwidth}
        \centering
       \includegraphics[width=0.85\linewidth]{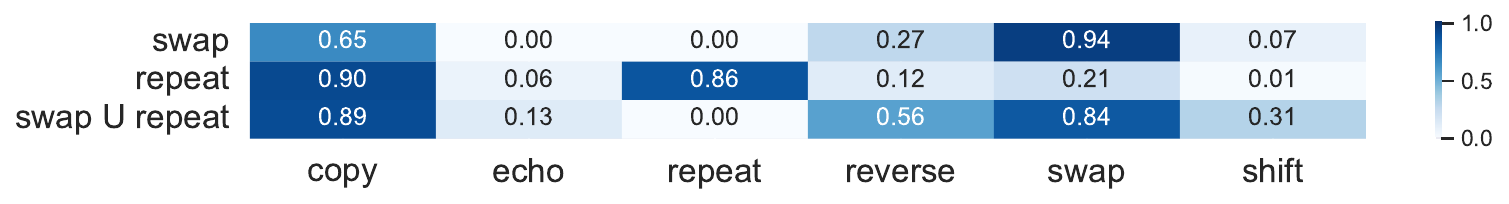}
    \end{subfigure}
    \vspace{1em}
    \begin{subfigure}{\textwidth}
       \centering
      \includegraphics[width=0.85\linewidth]{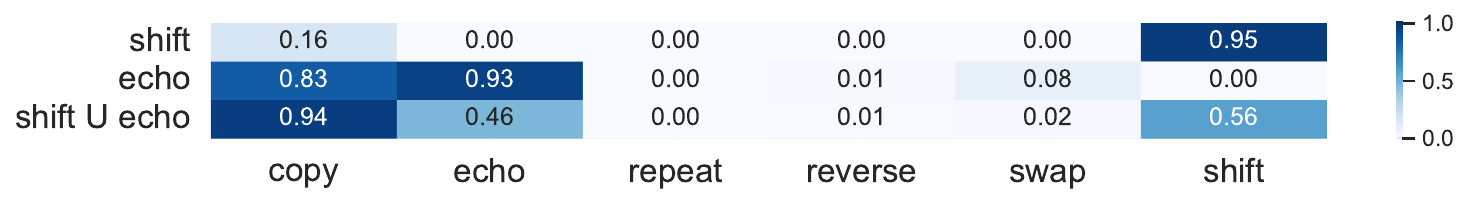}
    \end{subfigure}
    \vspace{1em}
    \begin{subfigure}{\textwidth}
       \centering
      \includegraphics[width=0.85\linewidth]{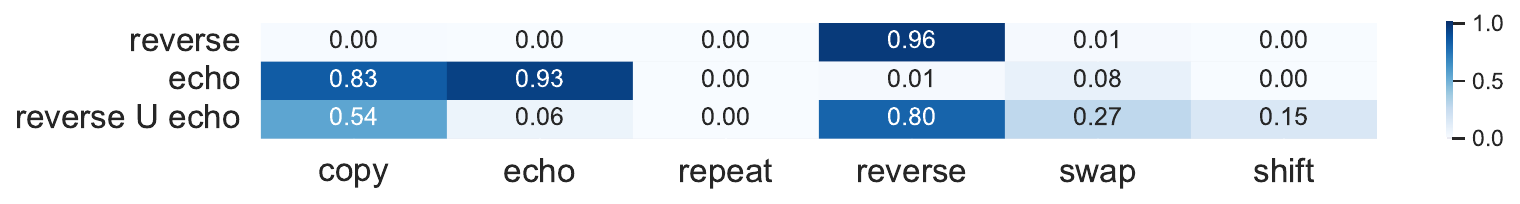}
    \end{subfigure}
    \caption{The results of combining circuits through a union operation on their respective binary masks.}
    \label{appendix:figure:set-comparison-experiments}
\end{figure*}

In Figure~\ref{appendix:figure:set-comparison-experiments} we present the cross-task accuracy of additional composite circuits as supplementary results to Section~\ref{sec:set-operations}. Consistent with previous observations, the union of two circuits acquires functional capacities for subtasks that the original base circuits cannot perform independently. For example, the union $\vm^{\texttt{swap}} \cup \vm^\texttt{repeat}$ achieves a $56\%$ accuracy on the \texttt{reverse} task and $31\%$ on \texttt{shift}, representing significant improvements over the performance of these circuits in isolation. Similarly, the union $\vm^{\texttt{reverse}} \cup \vm^\texttt{echo}$ reaches $31\%$ accuracy on \texttt{swap}. In the case of $\vm^{\texttt{shift}} \cup \vm^\texttt{echo}$, we observe that while the composite retains approximately half of the performance on the individual tasks, it also enhances performance on the \texttt{copy} task.

\subsection{Zero Ablation}\label{appendix:subsec:zero_ablation}
While the main paper focuses on circuits obtained through mean ablation, this section presents the results of experiments conducted using zero ablation.

\begin{figure}[b!]
    \centering
    \includegraphics[width=0.8\linewidth]{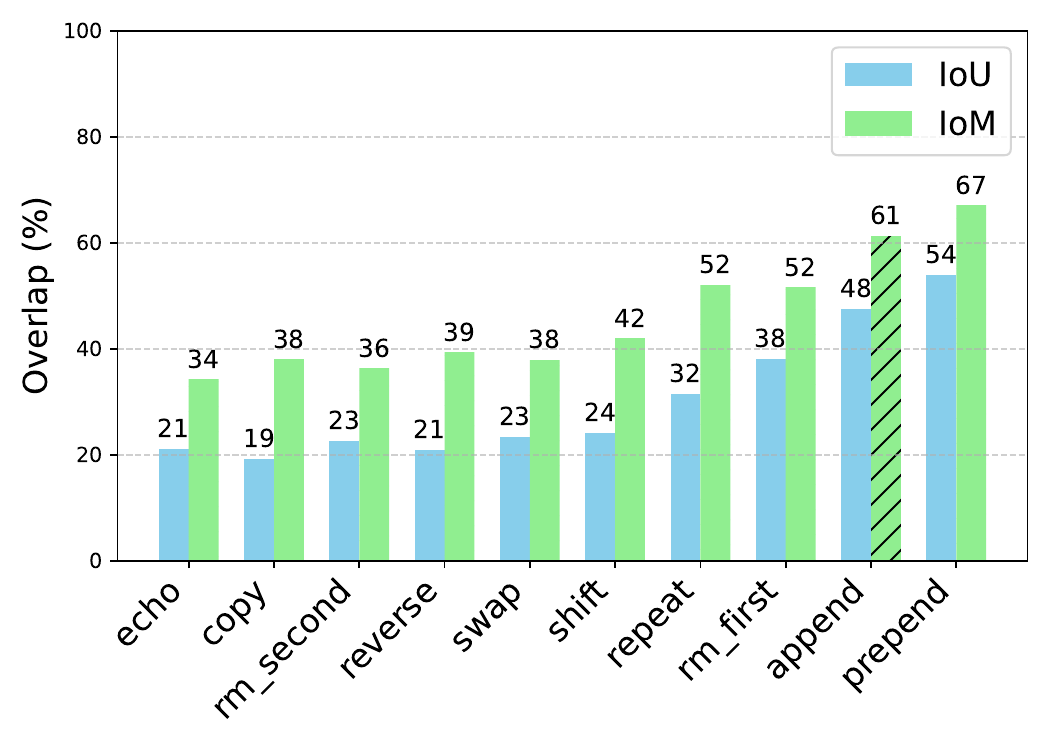}
    \caption{The overlap between mean and zero ablated circuits for all subtasks.}
    \label{appendix:figure:ablation-value-overlap}
\end{figure}

\subsubsection{Circuit Performance}
\begin{figure*}[h]
    \centering
    \begin{subfigure}[b]{0.329\textwidth}
    \centering
        \includegraphics[width=\textwidth]{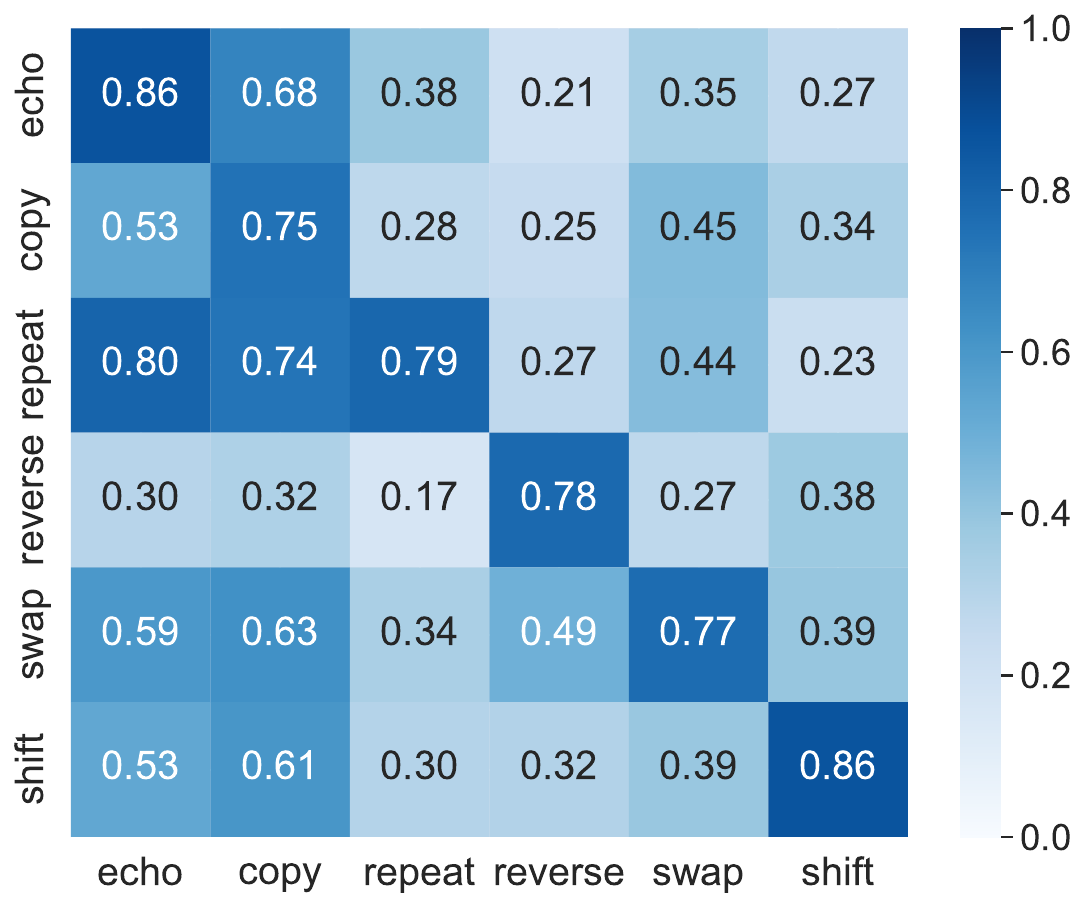}
        \caption{$F_T$ across all token positions.}
        \label{appendix:subfig:unary-jsd-all-tokens-zero}
    \end{subfigure}
    \hfill
    \begin{subfigure}[b]{0.329\textwidth}
    \centering
        \includegraphics[width=\textwidth]{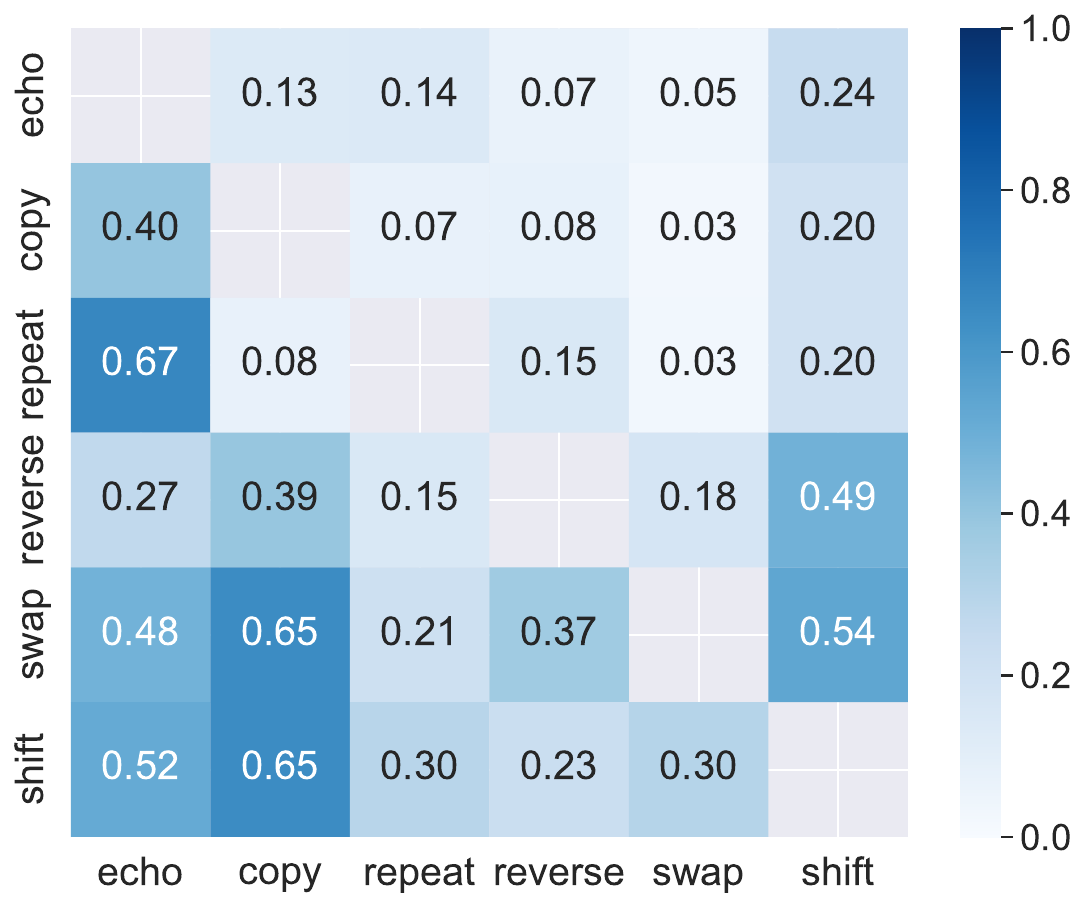}
        \caption{$F_T$ for selected token positions.}
        \label{appendix:subfig:unary-jsd-different-tokens-zero}
    \end{subfigure}
    \hfill
    \begin{subfigure}[b]{0.329\textwidth}
        \centering
        \includegraphics[width=\textwidth]{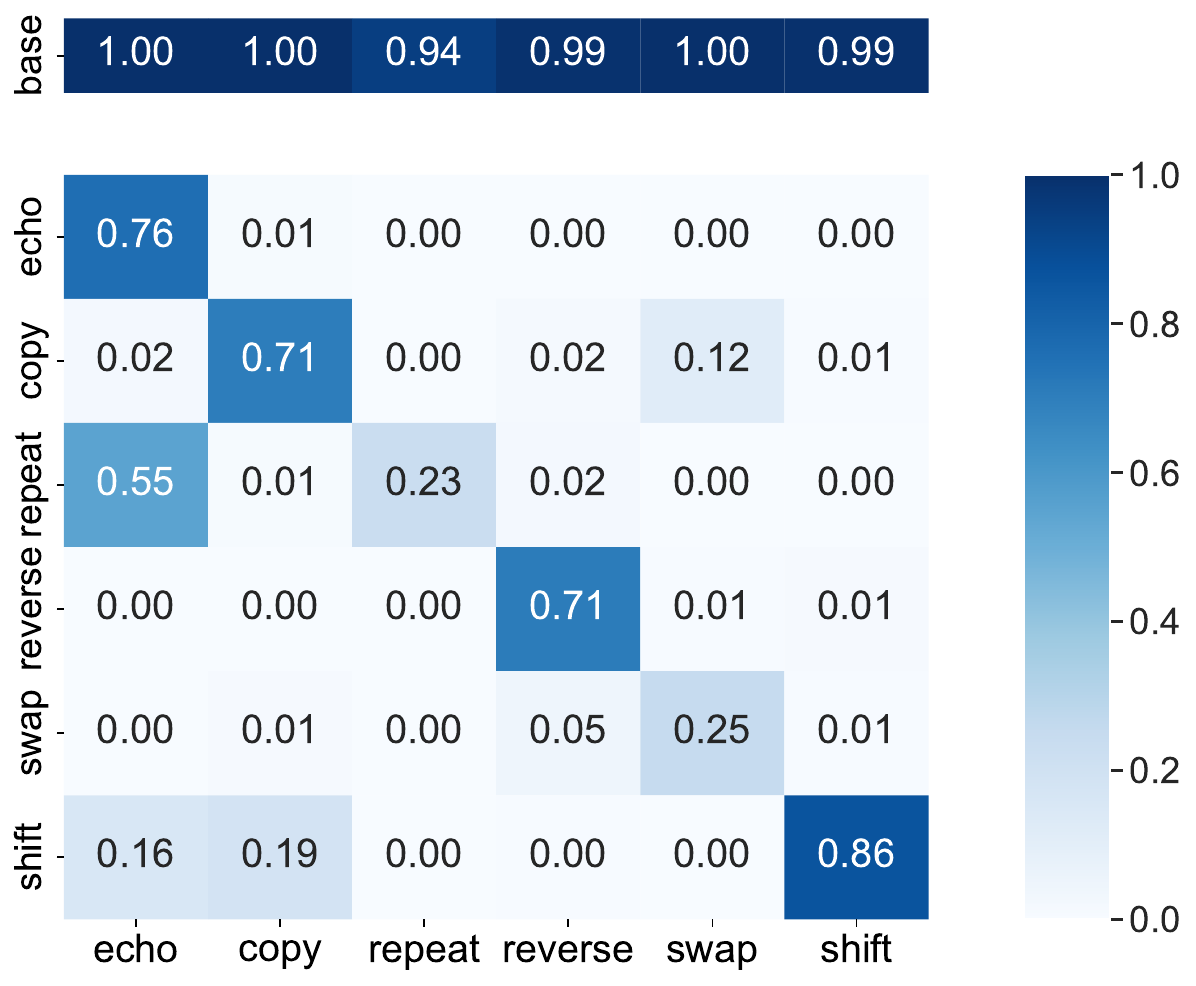}
        \caption{Accuracy.}
        \label{appendix:subfig:unary-accuray-zero}
    \end{subfigure}
    \caption{Task faithfulness performance $F_T$ and accuracy for the \textbf{unary} tasks for the \textbf{zero} ablated circuits. The y-axis corresponds to the circuit, while the x-axis represents the evaluation task. For task faithfulness with respect to positions where the ground truth tokens differ between the circuit and the evaluation task, the diagonal is omitted, as there are no applicable token positions for comparison.}
    \label{appendix:figure:unary-performance-zero}
\end{figure*}

\begin{figure*}[h]
    \centering
    \begin{subfigure}[b]{0.329\textwidth}
    \centering
        \includegraphics[width=\textwidth]{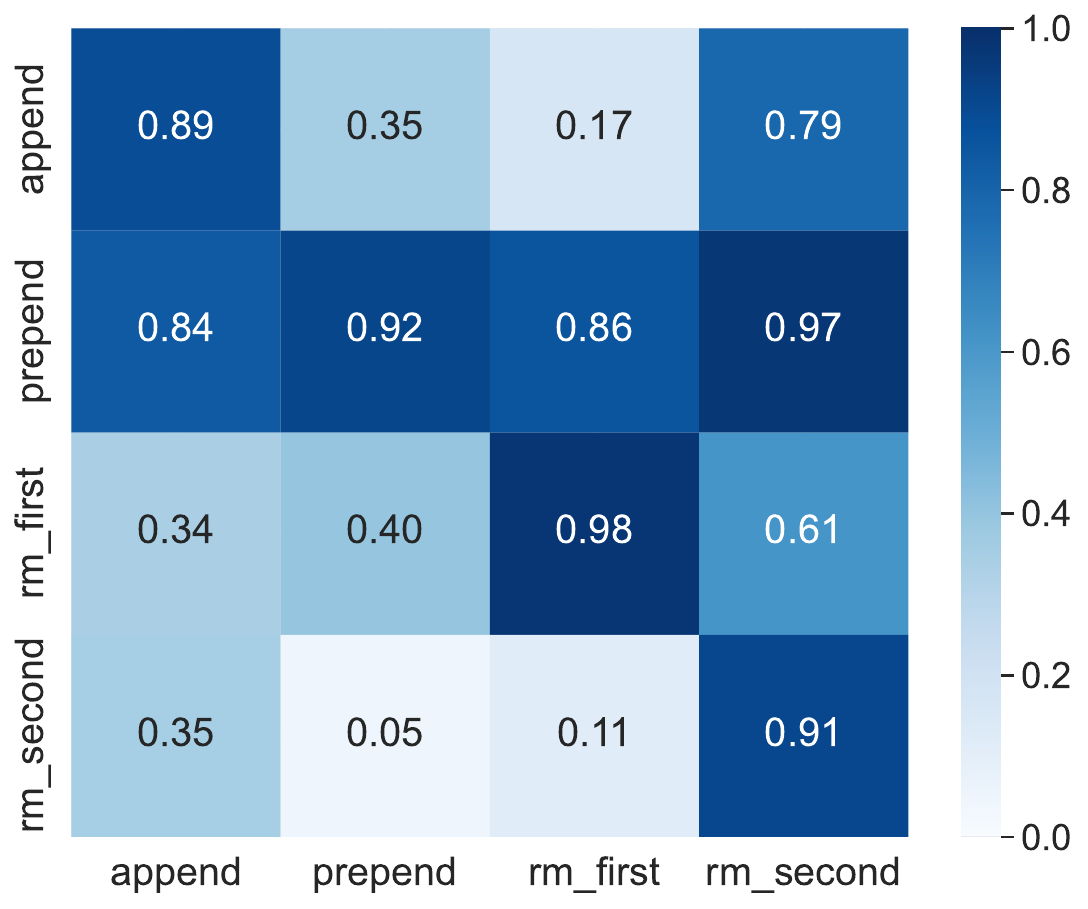}
        \caption{$F_T$ for all token positions.}
        \label{appendix:subfig:binary-jsd-all-tokens}
    \end{subfigure}
    \hfill
    \begin{subfigure}[b]{0.329\textwidth}
    \centering
        \includegraphics[width=\textwidth]{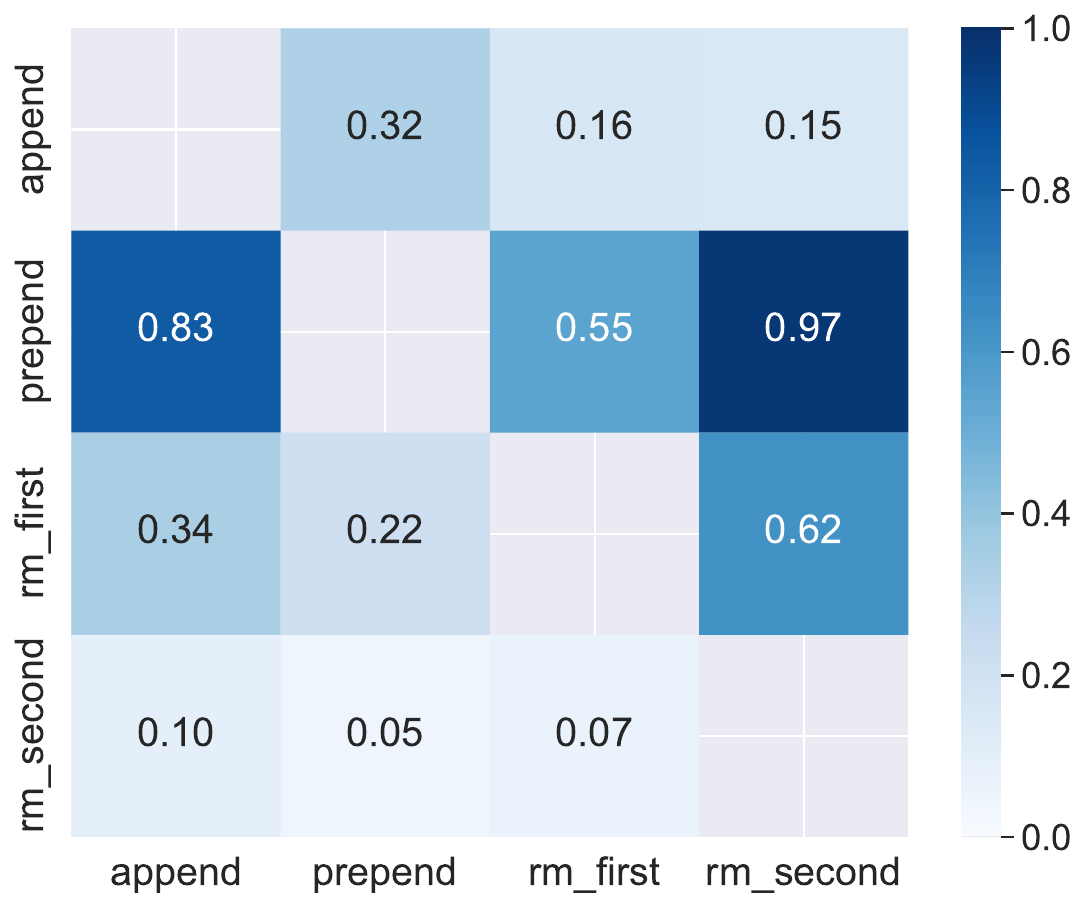}
        \caption{$F_T$ for selected token positions.}
        \label{appendix:subfig:binary-jsd-different-tokens}
    \end{subfigure}
    \hfill
    \begin{subfigure}[b]{0.329\textwidth}
        \centering
        \includegraphics[width=\textwidth]{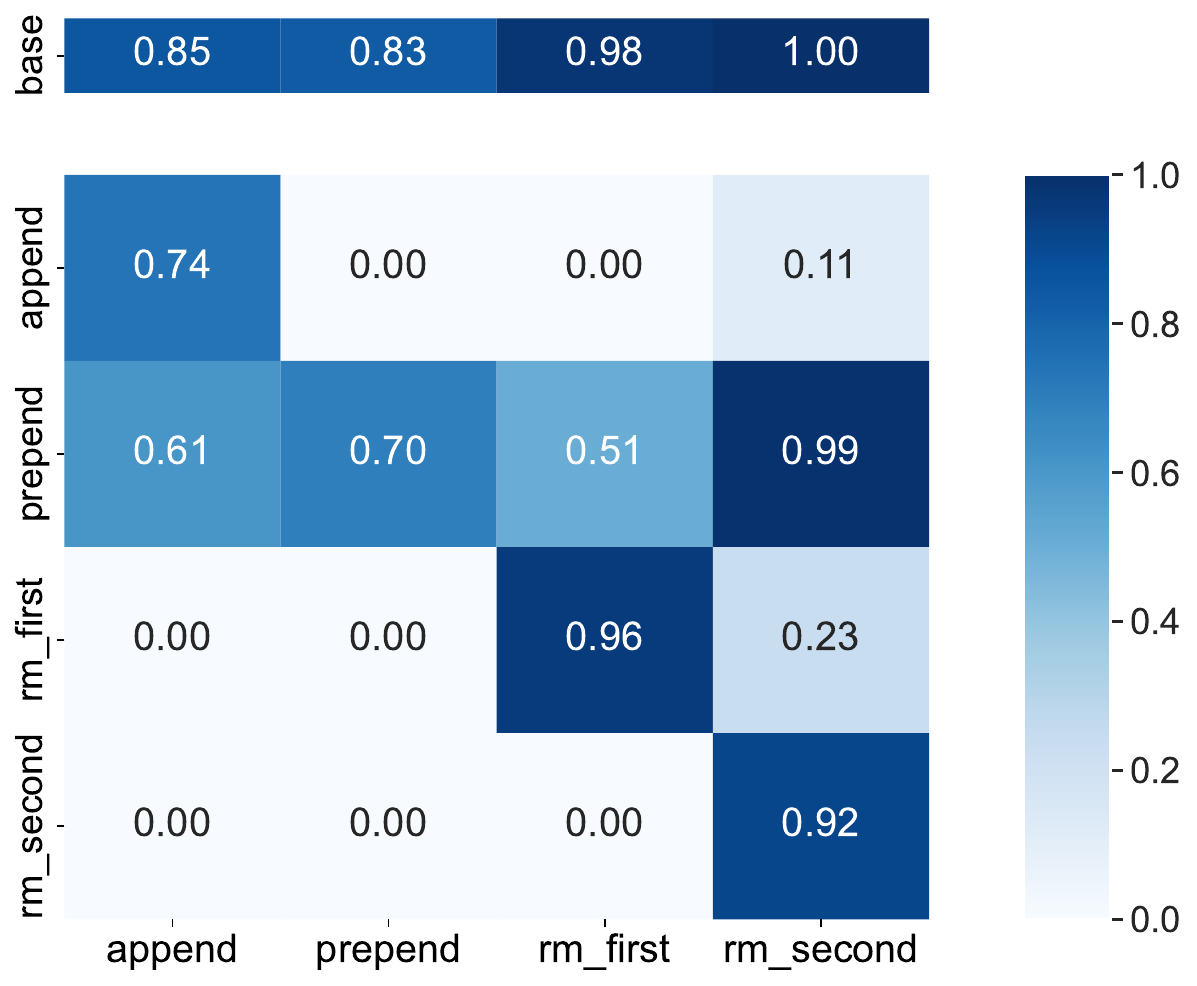}
        \caption{Accuracy.}
        \label{appendix:subfig:binary-accuray-zero}
    \end{subfigure} 
    \caption{Task faithfulness performance $F_T$ and accuracy for the \textbf{binary} tasks for the \textbf{zero} ablated circuits. The y-axis corresponds to the circuit, while the x-axis represents the evaluation task. For task faithfulness with respect to positions where the ground truth tokens differ between the circuit and the evaluation task, the diagonal is omitted, as there are no applicable token positions for comparison.}
    \label{appendix:figure:binary-performance-zero}
\end{figure*}

\begin{figure*}[h]
    \centering
    \begin{subfigure}[b]{0.40\textwidth}
        \centering
        \includegraphics[width=\textwidth]{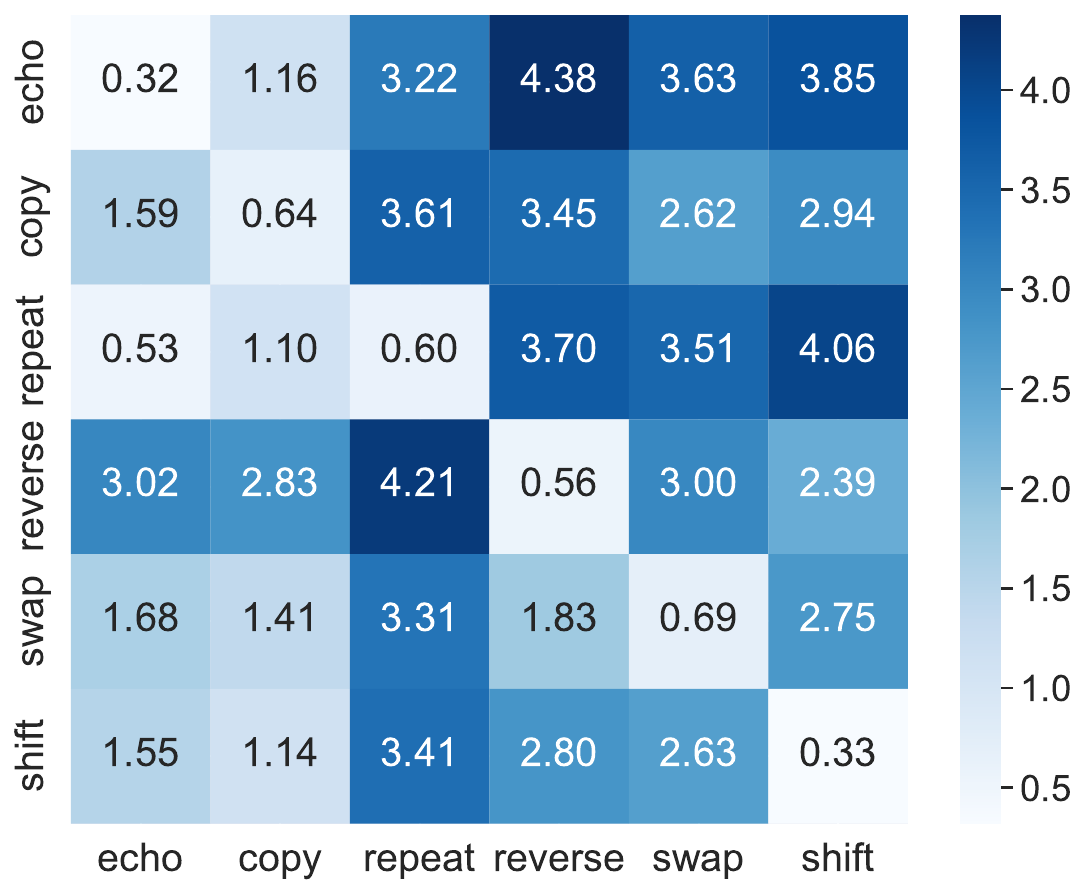}
        \caption{All token positions.}
        \label{appendix:figure:zero-kl-unary}
        \end{subfigure}
    \hfill
    \begin{subfigure}[b]{0.40\textwidth}
        \centering
        \includegraphics[width=\textwidth]{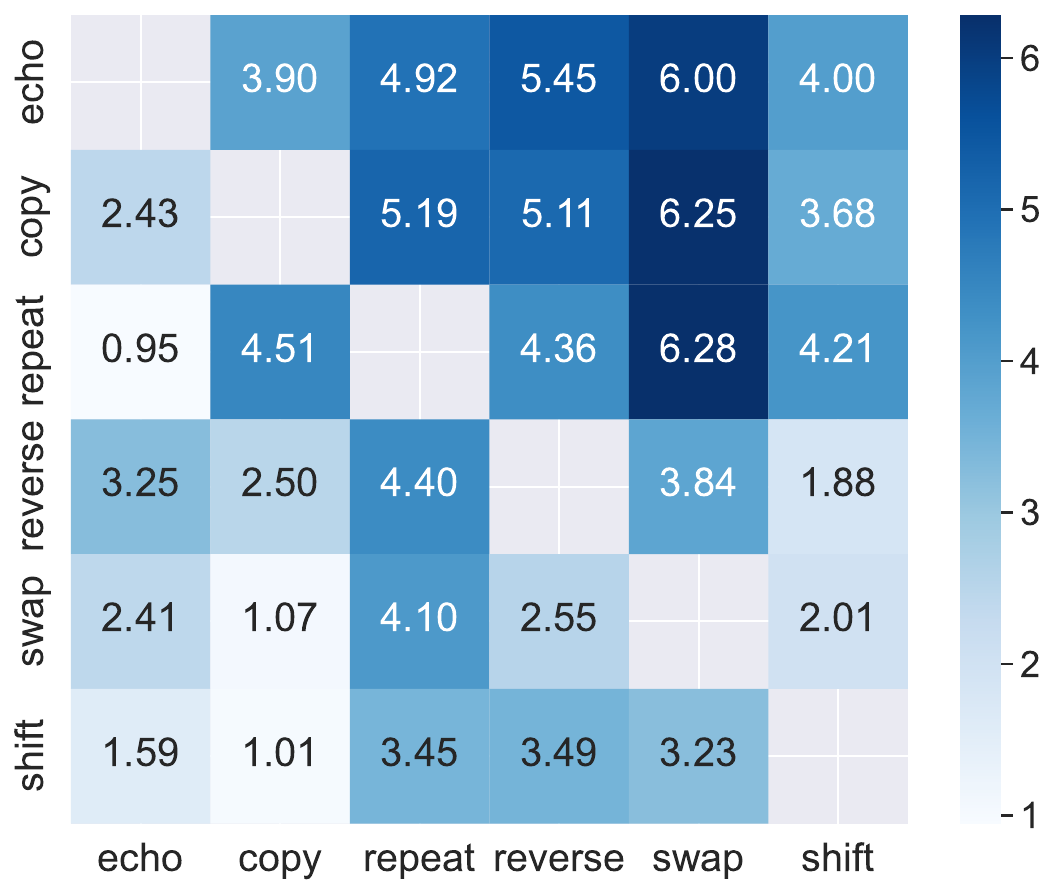}
        \caption{Differing token positions.}
        \label{appendix:figure:zero-kl-unary-diff}
    \end{subfigure}
    \caption{Task faithfulness measured via $\operatorname{D}_{KL}\left(\vy^{\vm} \parallel \vy^{\mathcal{M}}\right)$ for the \textbf{zero-ablated unary} circuits. The y-axis corresponds to the circuit, while the x-axis represents the evaluation task. When we only evaluate selected positions, we omit the diagonal, as there are no applicable tokens for comparison.}
    \label{appendix:figure:zero-ablaton-kl-divergence-scores-unary}
\end{figure*}

\begin{figure*}[h]
    \centering
    \begin{subfigure}[b]{0.40\textwidth}
        \centering
        \includegraphics[width=\textwidth]{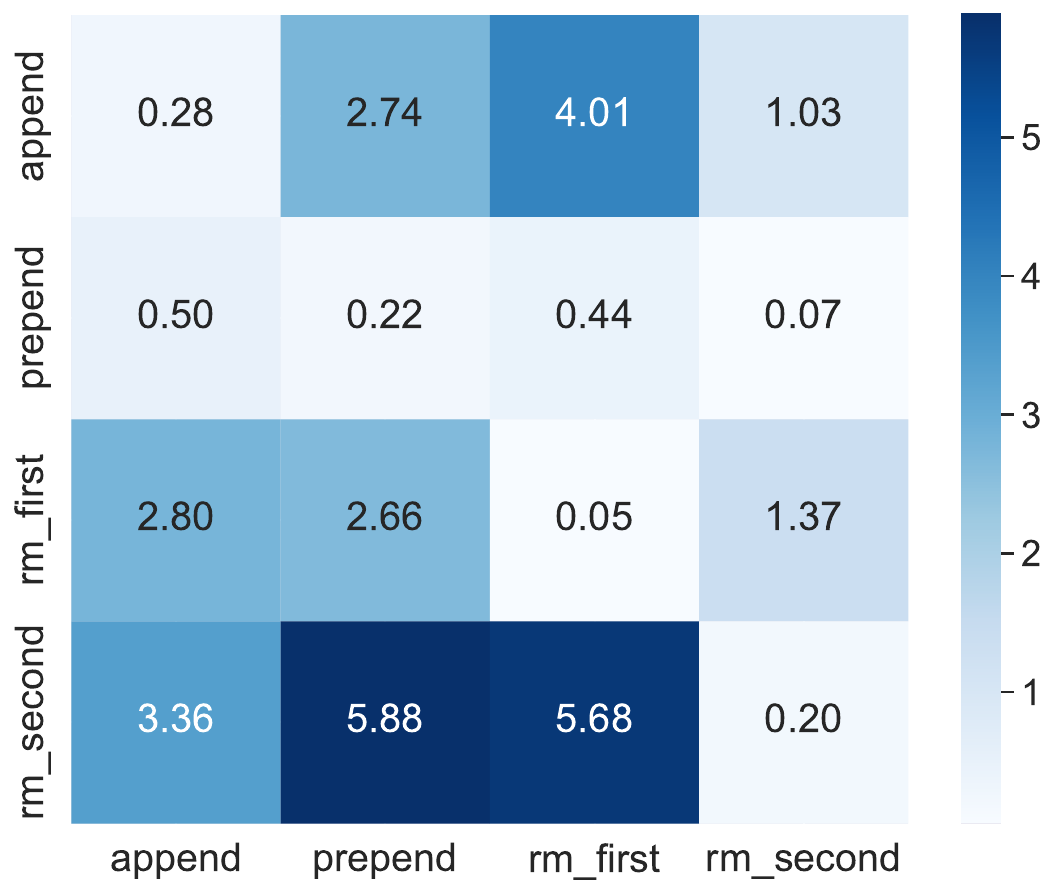}
        \caption{All token positions.}
        \label{appendix:figure:zero-kl-binary}
    \end{subfigure}
    \hfill
    \begin{subfigure}[b]{0.40\textwidth}
        \centering
        \includegraphics[width=\textwidth]{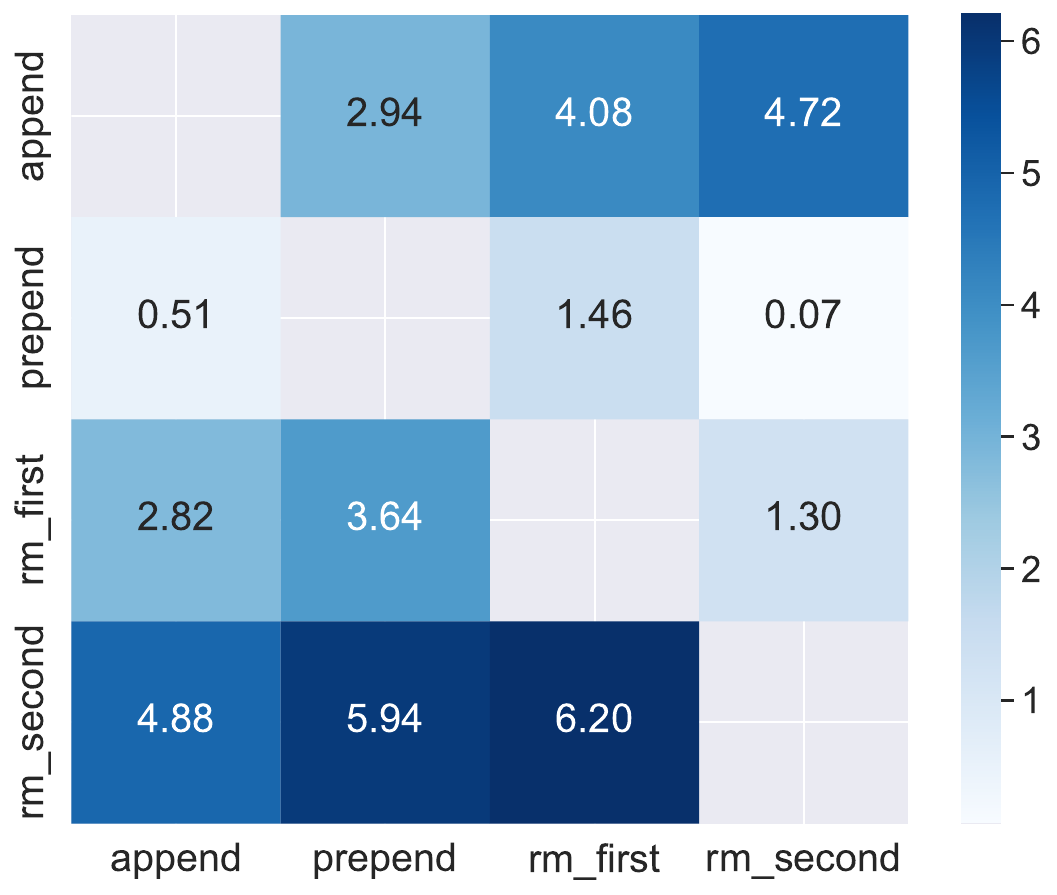}
        \caption{Differing token positions.}
        \label{appendix:figure:zero-kl-binary-diff}
    \end{subfigure}
    \caption{Task faithfulness measured via $\operatorname{D}_{KL}\left(\vy^{\vm} \parallel \vy^{\mathcal{M}}\right)$ for the \textbf{zero-ablated binary} circuits. The y-axis corresponds to the circuit, while the x-axis represents the evaluation task. When we only evaluate selected positions, we omit the diagonal, as there are no applicable tokens for comparison.}
    \label{appendix:figure:zero-ablaton-kl-divergence-scores-binary}
\end{figure*}

In Figure~\ref{appendix:figure:unary-performance-zero} and Figure~\ref{appendix:figure:binary-performance-zero}, we present the task faithfulness performance $F_T$ and generation accuracy for the zero-ablated circuits. Similarly, Figure~\ref{appendix:figure:zero-ablaton-kl-divergence-scores-unary} and Figure~\ref{appendix:figure:zero-ablaton-kl-divergence-scores-binary} illustrate task faithfulness in terms of the KL divergence between the predicted output distributions of the zero-ablated circuits, $\vm$, and the base model, $\mathcal{M}$, for both unary and binary string-edit operations. The key finding from these experiments is that zero ablation yields results closely aligned with those of mean ablation for binary operations, but not for unary operations. Similarly, patterns observed in mean-ablated unary circuits are not evident here. As noted in the literature, zero ablation can significantly shift the distribution, leading to degraded performance, which is one of the reasons for adopting mean ablation \citep{miller2024transformer}.

\subsubsection{Node Overlap}
Figure~\ref{appendix:figure:cross-task-overlap-zero} illustrates the IoU and IoM between all circuit-pairs under zero ablation.  The observed patterns are consistent with those identified under mean ablation (Figure~\ref{fig:node-overlap}), where unary circuits tend to exhibit higher node overlap than binary circuits.

\begin{figure}[tbp]
    \centering
    \begin{subfigure}[b]{0.8\linewidth}
        \centering
        \includegraphics[width=\linewidth]{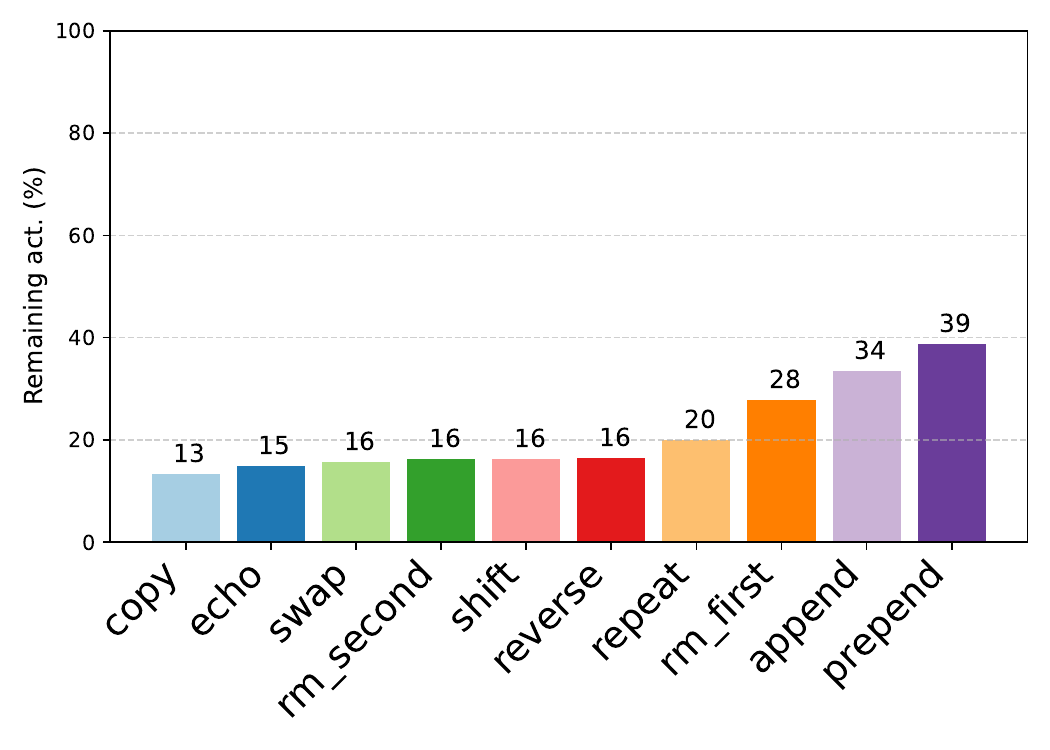}
        \caption{Mean ablation.}
    \end{subfigure}
    \vspace{0.5cm}
    \begin{subfigure}[b]{0.8\linewidth}
        \centering
        \includegraphics[width=\linewidth]{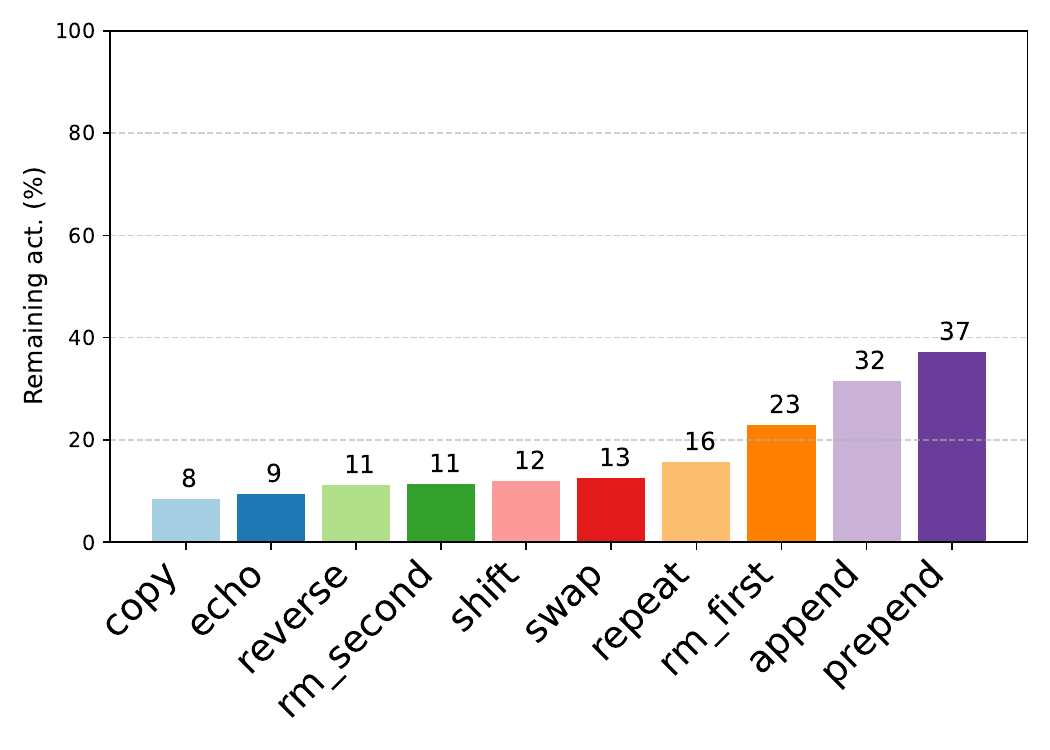}
        \caption{Zero ablation.}
    \end{subfigure}
    \caption{Overall remaining nodes after activation pruning.}
    \label{appendix:figure:global_sparsity}
\end{figure}

\subsection{Comparison Between Mean and Zero Ablation}
Figure~\ref{appendix:figure:ablation-value-overlap} illustrates the overlap between the mean and zero ablated circuits for identical subtasks. As discussed in Section~\ref{sec:Limitations}, we find significant differences in the circuits identified, specifically for unary circuits that tend to engage a smaller portion of the base model's activation space (refer to Figure~\ref{appendix:figure:global_sparsity} for comparison).

\subsection{Global Sparsity}\label{appendix:section:sparsity}
In Figure~\ref{appendix:figure:global_sparsity} we present the global sparsity of the circuits for both mean and zero ablation. As touched upon in Section~\ref{sec:Limitations}, we observe that while the two ablation strategies identify circuits of comparable sizes, they do not necessarily yield identical circuits. Additionally, the ordering of circuits in terms of remaining activations is nearly the same for both methods, with the only notable difference being the internal ordering of \texttt{swap} and \texttt{reverse}.

\section{Use of AI Assistants}
\label{appendix:section:ai}
For parts of our project’s source code, we used GitHub Copilot as an assistant tool.

\begin{figure*}
    \centering
    \begin{subfigure}[b]{0.45\textwidth}
        \centering
        \input{figures/tikz/local_sparsity_copy}
        \caption{The local sparsity of the \texttt{copy} circuit.}
    \end{subfigure}
    \begin{subfigure}[b]{0.45\textwidth}
        \centering
        \input{figures/tikz/local_sparsity_echo}
        \caption{The local sparsity of the \texttt{echo} circuit.}
    \end{subfigure}
    \par\bigskip
    \begin{subfigure}[b]{0.45\textwidth}
        \centering
        \input{figures/tikz/local_sparsity_repeat}
        \caption{The local sparsity of the \texttt{repeat} circuit.}
    \end{subfigure}
    \begin{subfigure}[b]{0.45\textwidth}
        \centering
        \input{figures/tikz/local_sparsity_reverse}
        \caption{The local sparsity of the \texttt{reverse} circuit.}
    \end{subfigure}
    \par\bigskip
    \begin{subfigure}[b]{0.45\textwidth}
        \centering
        \input{figures/tikz/local_sparsity_swap}
        \caption{The local sparsity of the \texttt{swap} circuit.}
    \end{subfigure}
    \begin{subfigure}[b]{0.45\textwidth}
        \centering
        \input{figures/tikz/local_sparsity_shift}
        \caption{The local sparsity of the \texttt{shift} circuit.}
    \end{subfigure}
    \caption{The local sparsity of the \textbf{unary} circuits achieved via mean ablation. Considered are feed-forward (FF), multi-head self-attention (MHSA), and multi-head cross-attention (MHCA) modules for each layer.}
    \label{appendix:figure:local-sparsity-unary}
\end{figure*}

\begin{figure*}
    \centering
    \begin{subfigure}[b]{0.45\textwidth}
        \centering
        \input{figures/tikz/local_sparsity_append}
        \caption{The local sparsity of the \texttt{append} circuit.}
    \end{subfigure}
    \begin{subfigure}[b]{0.45\textwidth}
        \centering
        \input{figures/tikz/local_sparsity_prepend}
        \caption{The local sparsity of the \texttt{prepend} circuit.}
    \end{subfigure}
    \par\bigskip
    \begin{subfigure}[b]{0.45\textwidth}
        \centering
        \input{figures/tikz/local_sparsity_remove_first}
        \caption{The local sparsity of the \texttt{remove\_first} circuit.}
    \end{subfigure}
    \begin{subfigure}[b]{0.45\textwidth}
        \centering
        \input{figures/tikz/local_sparsity_remove_second}
        \caption{The local sparsity of the \texttt{remove\_second} circuit.}
    \end{subfigure}
    \caption{The local sparsity of the \textbf{binary} circuits achieved via mean ablation. Considered are feed-forward (FF), multi-head self-attention (MHSA), and multi-head cross-attention (MHCA) modules for each layer.}
    \label{appendix:figure:local-sparsity-binary}
\end{figure*}

\end{document}

%% file: math_commands.tex
\usepackage{amsmath,amsfonts,bm}

\def\eqref#1{equation~\ref{#1}}

\def\1{\bm{1}}

\def\va{{\bm{a}}}

\def\vm{{\bm{m}}}

\def\vs{{\bm{s}}}

\def\vw{{\bm{w}}}
\def\vx{{\bm{x}}}
\def\vy{{\bm{y}}}
\def\vz{{\bm{z}}}

\def\eva{{a}}

\def\evm{{m}}

\DeclareMathAlphabet{\mathsfit}{\encodingdefault}{\sfdefault}{m}{sl}
\SetMathAlphabet{\mathsfit}{bold}{\encodingdefault}{\sfdefault}{bx}{n}

\def\gE{{\mathcal{E}}}

\def\gG{{\mathcal{G}}}

\def\gZ{{\mathcal{Z}}}

\newcommand{\R}{\mathbb{R}}



%% file: figures/tikz/intro_figure.tex
\begin{tikzpicture}[
    font=\sffamily\scriptsize,
    node distance=1.3cm and 1.6cm,
    >=LaTeX,
    line cap=round,
    line join=round,
    connection/.style={->, semithick},
    inputNode/.style={
      circle, draw=black!70, fill=yellow!60, 
      thick, minimum size=0.7em, inner sep=2pt
    },
    hiddenNode/.style={
      circle, draw=black!70, fill=blue!30,
      thick, minimum size=0.7em, inner sep=2pt
    },
    hidden2Node/.style={
      circle, draw=black!70, fill=green!50!gray,
      thick, minimum size=0.7em, inner sep=2pt
    },
    outputNode/.style={
      circle, draw=black!70, fill=pink!50,
      thick, minimum size=0.7em, inner sep=2pt
    },
    fadedNode/.style={
      circle, draw=black!15, fill=black!10,
      thick, minimum size=0.7em, inner sep=2pt
    }
]


\node[inputNode] (tI1) at (0,1) {};
\node[inputNode] (tI2) at (0,0.5) {};
\node[inputNode] (tI3) at (0,0) {};

\node[hiddenNode] (tH1) at (1.,1.25) {};
\node[fadedNode] (tH2) at (1.,0.75) {};
\node[hiddenNode] (tH3) at (1.,0.25) {};
\node[fadedNode] (tH4) at (1.,-0.25) {};

\node[fadedNode] (tG1) at (2,1.25) {};
\node[fadedNode] (tG2) at (2,0.75) {};
\node[hidden2Node] (tG3) at (2,0.25) {};
\node[fadedNode] (tG4) at (2,-0.25) {};

\node[outputNode] (tO1) at (3,0.5) {};

\node[anchor=east, align=center] 
  at ($(tI1)!0.5!(tI3) + (-0.175,0)$) {
    \texttt{Reverse}\\
    $x_1, x_2, \ldots, x_n$
};
  
\node[anchor=west] (topLabel) 
  at ($(tO1) + (0.175,0)$) {$x_n, x_{n\text{\tiny-}1}, \ldots x_2, x_1$};

\foreach \i in {1,2,3} {
  \foreach \j in {2,4} {
    \draw[fadedNode, ->] (tI\i) -- (tH\j);
  }
}
\foreach \i in {1,2,3,4} {
    \foreach \j in {1,2,4} {
    \draw[fadedNode, ->] (tH\i) -- (tG\j);
    }
}
\foreach \i in {1,2,4} {
    \draw[fadedNode, ->] (tG\i) -- (tO1);
}

\foreach \i in {1,2,3} {
  \foreach \j in {1,3} {
    \draw[connection] (tI\i) -- (tH\j);
  }
}
\foreach \i in {1,3} {
    \foreach \j in {3} {
    \draw[connection] (tH\i) -- (tG\j);
    }
}
\foreach \i in {3} {
    \draw[connection] (tG\i) -- (tO1);
}


\begin{scope}[yshift=-2.75cm] 

\node[inputNode] (bI1) at (0,1.0) {};
\node[inputNode] (bI2) at (0,0.5) {};
\node[inputNode] (bI3) at (0,0) {};

\node[fadedNode] (bH1) at (1,1.25) {};
\node[hiddenNode] (bH2) at (1,0.75) {};
\node[hiddenNode] (bH3) at (1,0.25) {};
\node[fadedNode] (bH4) at (1,-0.25) {};

\node[hidden2Node] (bG1) at (2,1.25) {};
\node[fadedNode] (bG2) at (2,0.75) {};
\node[fadedNode] (bG3) at (2,0.25) {};
\node[hidden2Node] (bG4) at (2,-0.25) {};

\node[outputNode] (bO1) at (3,0.5) {};

\node[anchor=east, align=center] 
  at ($(bI1)!0.5!(bI3) + (-0.175,0)$) {
    \texttt{Swap}\\
    $x_1, x_2, \ldots, x_n$
};

\node[anchor=west] (botLabel) 
  at ($(bO1) + (0.175,0)$) {$x_n, x_2, \ldots x_{n\text{\tiny-}1}, x_1$};

\foreach \i in {1,2,3} {
  \foreach \j in {1,4} {
    \draw[fadedNode, ->] (bI\i) -- (bH\j);
  }
}
\foreach \i in {1,2,3,4} {
    \foreach \j in {1,2,3,4} {
    \draw[fadedNode, ->] (bH\i) -- (bG\j);
    }
}
\foreach \i in {2,3} {
    \draw[fadedNode, ->] (bG\i) -- (bO1);
}

\foreach \i in {1,2,3} {
  \foreach \j in {2,3} {
    \draw[connection] (bI\i) -- (bH\j);
  }
}
\foreach \i in {2,3} {
    \foreach \j in {1,4} {
    \draw[connection] (bH\i) -- (bG\j);
    }
}
\foreach \i in {1,4} {
    \draw[connection] (bG\i) -- (bO1);
}

\end{scope}

\node[font=\bfseries] (qm) at (1.25, -0.85) {?};
\draw[<->, dashed, thick] (tH4) -- (bH1);

\end{tikzpicture}

%% file: figures/tikz/local_sparsity_copy.tex
\begin{tikzpicture}
    \tikzstyle{standard} = [rounded corners=3pt];
    \tikzstyle{ffwnode} = [minimum height=1.5em,minimum width=2.0em,inner sep=1pt,rounded corners=3pt,draw,fill=cyan!30];
    \tikzstyle{msanode} = [minimum height=1.5em,minimum width=2.0em,inner sep=1pt,rounded corners=3pt,draw,fill=green!30];
    \tikzstyle{mcanode} = [minimum height=1.5em,minimum width=2.0em,inner sep=1pt,rounded corners=3pt,draw,fill=purple!30];

    \node [msanode,anchor=west, fill=green!4!white] (msa1) at (0,0) {\tiny \textbf{4.0\%}};
    \node [ffwnode,anchor=west, fill=cyan!36!white] (ffw1) at ([xshift=0.75em]msa1.east) {\tiny \textbf{36\%}};
    \node [anchor=south west] at ([xshift=-1.25em,yshift=0.0em]msa1.south west) {\textbf{1}};
    
    \node [ffwnode,anchor=south, fill=cyan!25!white] (ffw2) at ([yshift=0.4em]ffw1.north) {\tiny \textbf{25\%}};
    \node [msanode,anchor=south, fill=green!19!white] (msa2) at ([yshift=0.4em]msa1.north) {\tiny \textbf{19\%}};
    \node [anchor=south west] at ([xshift=-1.25em,yshift=0.0em]msa2.south west) {\textbf{2}};
    
    \node [ffwnode,anchor=south, fill=cyan!21!white] (ffw3) at ([yshift=0.4em]ffw2.north) {\tiny \textbf{21\%}};
    \node [msanode,anchor=south, fill=green!13!white] (msa3) at ([yshift=0.4em]msa2.north) {\tiny \textbf{13\%}};
    \node [anchor=south west] at ([xshift=-1.25em,yshift=0.0em]msa3.south west) {\textbf{3}};
    
    \node [ffwnode,anchor=south, fill=cyan!25!white] (ffw4) at ([yshift=0.4em]ffw3.north) {\tiny \textbf{25\%}};
    \node [msanode,anchor=south, fill=green!6!white] (msa4) at ([yshift=0.4em]msa3.north) {\tiny \textbf{6.0\%}};
    \node [anchor=south west] at ([xshift=-1.25em,yshift=0.0em]msa4.south west) {\textbf{4}};
    
    \node [ffwnode,anchor=south, fill=cyan!23!white] (ffw5) at ([yshift=0.4em]ffw4.north) {\tiny \textbf{23\%}};
    \node [msanode,anchor=south, fill=green!9!white] (msa5) at ([yshift=0.4em]msa4.north) {\tiny \textbf{9.0\%}};
    \node [anchor=south west] at ([xshift=-1.25em,yshift=0.0em]msa5.south west) {\textbf{5}};
    
    \node [ffwnode,anchor=south, fill=cyan!22!white] (ffw6) at ([yshift=0.4em]ffw5.north) {\tiny \textbf{22\%}};
    \node [msanode,anchor=south, fill=green!11!white] (msa6) at ([yshift=0.4em]msa5.north) {\tiny \textbf{11\%}};
    \node [anchor=south west] at ([xshift=-1.25em,yshift=0.0em]msa6.south west) {\textbf{6}};
    
    \node [anchor=south] (encoder) at ([xshift=0.5em,yshift=0.5em]msa6.north) {\textbf{Encoder}};
    
    \draw[rounded corners=10pt, thick, draw=black] ([xshift=-1.25em,yshift=-0.5em]msa1.south west) rectangle ([xshift=.5em,yshift=0.5em]ffw6.north east);
        
    \node [msanode,anchor=west, fill=green!0!white] (msa_d1) at ([xshift=6em]msa1.east) {\tiny \textbf{0.0\%}};
    \node [mcanode,anchor=west, fill=purple!20!white] (mca_d1) at ([xshift=0.75em]msa_d1.east) {\tiny \textbf{20\%}};
    \node [ffwnode,anchor=west, fill=cyan!18!white] (ffw_d1) at ([xshift=0.75em]mca_d1.east) {\tiny \textbf{18\%}};

    
    \node [ffwnode,anchor=south, fill=cyan!12!white] (ffw_d2) at ([yshift=0.4em]ffw_d1.north) {\tiny \textbf{12\%}};
    \node [msanode,anchor=south, fill=green!2!white] (msa_d2) at ([yshift=0.4em]msa_d1.north) {\tiny \textbf{2.0\%}};
    \node [mcanode,anchor=south, fill=purple!19!white] (mca_d2) at ([yshift=0.4em]mca_d1.north) {\tiny \textbf{19\%}};
        
    \node [ffwnode,anchor=south, fill=cyan!3!white] (ffw_d3) at ([yshift=0.4em]ffw_d2.north) {\tiny \textbf{3.0\%}};
    \node [msanode,anchor=south, fill=green!2!white] (msa_d3) at ([yshift=0.4em]msa_d2.north) {\tiny \textbf{2.0\%}};
    \node [mcanode,anchor=south, fill=purple!17!white] (mca_d3) at ([yshift=0.4em]mca_d2.north) {\tiny \textbf{17\%}};
    
    \node [ffwnode,anchor=south, fill=cyan!1!white] (ffw_d4) at ([yshift=0.4em]ffw_d3.north) {\tiny \textbf{1.0\%}};
    \node [msanode,anchor=south, fill=green!1!white] (msa_d4) at ([yshift=0.4em]msa_d3.north) {\tiny \textbf{1.0\%}};
    \node [mcanode,anchor=south, fill=purple!25!white] (mca_d4) at ([yshift=0.4em]mca_d3.north) {\tiny \textbf{25\%}};
    
    \node [ffwnode,anchor=south, fill=cyan!1!white] (ffw_d5) at ([yshift=0.4em]ffw_d4.north) {\tiny \textbf{1.0\%}};
    \node [msanode,anchor=south, fill=green!2!white] (msa_d5) at ([yshift=0.4em]msa_d4.north) {\tiny \textbf{2.0\%}};
    \node [mcanode,anchor=south, fill=purple!21!white] (mca_d5) at ([yshift=0.4em]mca_d4.north) {\tiny \textbf{21\%}};
    
    \node [ffwnode,anchor=south, fill=cyan!1!white] (ffw_d6) at ([yshift=0.4em]ffw_d5.north) {\tiny \textbf{1.0\%}};
    \node [msanode,anchor=south, fill=green!2!white] (msa_d6) at ([yshift=0.4em]msa_d5.north) {\tiny \textbf{2.0\%}};
    \node [mcanode,anchor=south, fill=purple!42!white] (mca_d6) at ([yshift=0.4em]mca_d5.north) {\tiny \textbf{42\%}};
    
    \node [anchor=south] (decoder) at ([xshift=1.0em,yshift=0.5em]msa_d6.north) {\textbf{Decoder}};
    
    \draw[rounded corners=10pt, thick, draw=black] ([xshift=-0.5em,yshift=-0.5em]msa_d1.south west) rectangle ([xshift=0.5em,yshift=0.5em]ffw_d6.north east);

    \draw[->,standard] ([xshift=0.5em,yshift=0em]ffw4.east) -- ([xshift=-.5em,yshift=0em]msa_d4.west);

    \node [ffwnode,anchor=south] (legend_ffw_encoder) at ([yshift=2.5em]ffw6.north) {\tiny \textbf{FF}};
    \node [msanode,anchor=south] (legend_msa_encoder) at ([yshift=2.5em]msa6.north) {\tiny \textbf{MHSA}};
    
    \node [ffwnode,anchor=south] (legend_ffw) at ([yshift=2.5em]ffw_d6.north) {\tiny \textbf{FF}};
    \node [msanode,anchor=south] (legend_msa) at ([yshift=2.5em]msa_d6.north) {\tiny \textbf{MHSA}};
    \node [mcanode,anchor=south] (legend_mca) at ([yshift=2.5em]mca_d6.north) {\tiny \textbf{MHCA}};

    \draw[->, standard] ([yshift=-1.5em, xshift=1em]msa1.south) -- ([yshift=-0.5em, xshift=1em]msa1.south);
    \draw[->, standard] ([yshift=-1.5em]mca_d1.south) -- ([yshift=-0.5em]mca_d1.south);
        
\end{tikzpicture}

%% file: figures/tikz/tracr/tracr_local_sparsity_copy.tex
\begin{tikzpicture}
    \tikzstyle{standard} = [rounded corners=3pt];
    \tikzstyle{ffwnode} = [minimum height=1.5em,minimum width=2.0em,inner sep=1pt,rounded corners=3pt,draw,fill=cyan!30];
    \tikzstyle{msanode} = [minimum height=1.5em,minimum width=2.0em,inner sep=1pt,rounded corners=3pt,draw,fill=green!30];
        
    \node [msanode,anchor=west, fill=green!26.67!white] (msa_d1) at ([xshift=6em]ffw1.east) {\tiny 26.67\%};
    \node [ffwnode,anchor=west, fill=cyan!0!white] (ffw_d1) at ([xshift=0.75em]msa_d1.east) {\tiny 0.0\%};
    \node [anchor=south west] at ([xshift=-1.25em,yshift=0.0em]msa_d1.south west) {\textbf{1}};


    \draw[rounded corners=10pt, thick, draw=black] ([xshift=-1.25em,yshift=-0.5em]msa_d1.south west) rectangle ([xshift=0.5em,yshift=0.5em]ffw_d1.north east);

    
    \node [msanode,anchor=south] (legend_msa) at ([yshift=2.5em]msa_d1.north) {\tiny Attn };
    \node [ffwnode,anchor=south] (legend_ffw) at ([yshift=2.5em]ffw_d1.north) {\tiny MLP};

    \draw[->, standard] ([yshift=-1.5em, xshift=-1.5em]ffw_d1.south) -- ([yshift=-0.9em, xshift=-1.5em]ffw_d1.south);
        
\end{tikzpicture}

%% file: figures/tikz/tracr/tracr_local_sparsity_echo.tex
\begin{tikzpicture}
    \tikzstyle{standard} = [rounded corners=3pt];
    \tikzstyle{ffwnode} = [minimum height=1.5em,minimum width=2.0em,inner sep=1pt,rounded corners=3pt,draw,fill=cyan!30];
    \tikzstyle{msanode} = [minimum height=1.5em,minimum width=2.0em,inner sep=1pt,rounded corners=3pt,draw,fill=green!30];

    \node [msanode,anchor=west, fill=green!0!white] (msa_d1) at ([xshift=6em]ffw1.east) {\tiny 0.0\%};
    \node [ffwnode,anchor=west, fill=cyan!17.39!white] (ffw_d1) at ([xshift=0.75em]msa_d1.east) {\tiny 17.39\%};
    \node [anchor=south west] at ([xshift=-1.25em,yshift=0.0em]msa_d1.south west) {\textbf{1}};

    \node [msanode,anchor=south, fill=green!0!white] (msa_d2) at ([yshift=0.4em]msa_d1.north) {\tiny 0.0\%};
    \node [ffwnode,anchor=south, fill=cyan!10.87!white] (ffw_d2) at ([yshift=0.4em]ffw_d1.north) {\tiny 10.87\%};
    \node [anchor=south west] at ([xshift=-1.25em,yshift=0.0em]msa_d2.south west) {\textbf{2}};

    \node [msanode,anchor=south, fill=green!10.87!white] (msa_d3) at ([yshift=0.4em]msa_d2.north) {\tiny 10.87\%};
    \node [ffwnode,anchor=south, fill=cyan!6.52!white] (ffw_d3) at ([yshift=0.4em]ffw_d2.north) {\tiny 6.52\%};
    \node [anchor=south west] at ([xshift=-1.25em,yshift=0.0em]msa_d3.south west) {\textbf{3}};

    
    \node [msanode,anchor=south, fill=green!6.52!white] (msa_d4) at ([yshift=0.4em]msa_d3.north) {\tiny 6.52\%};
    \node [ffwnode,anchor=south, fill=cyan!0!white] (ffw_d4) at ([yshift=0.4em]ffw_d3.north) {\tiny 0.0\%};
    \node [anchor=south west] at ([xshift=-1.25em,yshift=0.0em]msa_d4.south west) {\textbf{4}};
    
    \draw[rounded corners=10pt, thick, draw=black] ([xshift=-1.25em,yshift=-0.5em]msa_d1.south west) rectangle ([xshift=0.5em,yshift=0.5em]ffw_d4.north east);

    
    
    \node [msanode,anchor=south] (legend_msa) at ([yshift=2.5em]msa_d4.north) {\tiny Attn };
    \node [ffwnode,anchor=south] (legend_ffw) at ([yshift=2.5em]ffw_d4.north) {\tiny MLP};

    \draw[->, standard] ([yshift=-1.5em, xshift=-1.5em]ffw_d1.south) -- ([yshift=-0.9em, xshift=-1.5em]ffw_d1.south);
        
\end{tikzpicture}

%% file: figures/tikz/tracr/tracr_local_sparsity_reverse.tex
\begin{tikzpicture}
    \tikzstyle{standard} = [rounded corners=3pt];
    \tikzstyle{ffwnode} = [minimum height=1.5em,minimum width=2.0em,inner sep=1pt,rounded corners=3pt,draw,fill=cyan!30];
    \tikzstyle{msanode} = [minimum height=1.5em,minimum width=2.0em,inner sep=1pt,rounded corners=3pt,draw,fill=green!30];

    \node [msanode,anchor=west, fill=green!0.0!white] (msa_d1) at ([xshift=6em]ffw1.east) {\tiny 0.0\%};
    \node [ffwnode,anchor=west, fill=cyan!2.70!white] (ffw_d1) at ([xshift=0.75em]msa_d1.east) {\tiny 2.70\%};
    \node [anchor=south west] at ([xshift=-1.25em,yshift=0.0em]msa_d1.south west) {\textbf{1}};

    
    \node [msanode,anchor=south, fill=green!0!white] (msa_d2) at ([yshift=0.4em]msa_d1.north) {\tiny 0.0\%};
    \node [ffwnode,anchor=south, fill=cyan!10.81!white] (ffw_d2) at ([yshift=0.4em]ffw_d1.north) {\tiny 10.81\%};
    \node [anchor=south west] at ([xshift=-1.25em,yshift=0.0em]msa_d2.south west) {\textbf{2}};

        
    \node [msanode,anchor=south, fill=green!0!white] (msa_d3) at ([yshift=0.4em]msa_d2.north) {\tiny 0.0\%};
    \node [ffwnode,anchor=south, fill=cyan!10.81!white] (ffw_d3) at ([yshift=0.4em]ffw_d2.north) {\tiny 10.81\%};
    \node [anchor=south west] at ([xshift=-1.25em,yshift=0.0em]msa_d3.south west) {\textbf{3}};

    
    \node [msanode,anchor=south, fill=green!10.81!white] (msa_d4) at ([yshift=0.4em]msa_d3.north) {\tiny 10.81\%};
    \node [ffwnode,anchor=south, fill=cyan!0!white] (ffw_d4) at ([yshift=0.4em]ffw_d3.north) {\tiny 0.0\%};
    \node [anchor=south west] at ([xshift=-1.25em,yshift=0.0em]msa_d4.south west) {\textbf{4}};

    \draw[rounded corners=10pt, thick, draw=black] ([xshift=-1.25em,yshift=-0.5em]msa_d1.south west) rectangle ([xshift=0.5em,yshift=0.5em]ffw_d4.north east);

    
    \node [msanode,anchor=south] (legend_ffw) at ([yshift=2.5em]msa_d4.north) {\tiny Attn};
    \node [ffwnode,anchor=south] (legend_msa) at ([yshift=2.5em]ffw_d4.north) {\tiny MLP};

    \draw[->, standard] ([yshift=-1.5em, xshift=-1.5em]ffw_d1.south) -- ([yshift=-0.9em, xshift=-1.5em]ffw_d1.south);
        
\end{tikzpicture}

%% file: figures/tikz/tracr/tracr_local_sparsity_swap.tex
\begin{tikzpicture}
    \tikzstyle{standard} = [rounded corners=3pt];
    \tikzstyle{ffwnode} = [minimum height=1.5em,minimum width=2.0em,inner sep=1pt,rounded corners=3pt,draw,fill=cyan!30];
    \tikzstyle{msanode} = [minimum height=1.5em,minimum width=2.0em,inner sep=1pt,rounded corners=3pt,draw,fill=green!30];

    \node [msanode,anchor=west, fill=green!0.0!white] (msa_d1) at ([xshift=6em]ffw1.east) {\tiny 0.0\%};
    \node [ffwnode,anchor=west, fill=cyan!4.05!white] (ffw_d1) at ([xshift=0.75em]msa_d1.east) {\tiny 4.05\%};
    \node [anchor=south west] at ([xshift=-1.25em,yshift=0.0em]msa_d1.south west) {\textbf{1}};

    
    \node [msanode,anchor=south, fill=green!0.0!white] (msa_d2) at ([yshift=0.4em]msa_d1.north) {\tiny 0.0\%};
    \node [ffwnode,anchor=south, fill=cyan!6.76!white] (ffw_d2) at ([yshift=0.4em]ffw_d1.north) {\tiny 6.76\%};
    \node [anchor=south west] at ([xshift=-1.25em,yshift=0.0em]msa_d2.south west) {\textbf{2}};

        
    \node [msanode,anchor=south, fill=green!0.0!white] (msa_d3) at ([yshift=0.4em]msa_d2.north) {\tiny 0.0\%};
    \node [ffwnode,anchor=south, fill=cyan!8.11!white] (ffw_d3) at ([yshift=0.4em]ffw_d2.north) {\tiny 8.11\%};
    \node [anchor=south west] at ([xshift=-1.25em,yshift=0.0em]msa_d3.south west) {\textbf{3}};

    
    \node [msanode,anchor=south, fill=green!2.70!white] (msa_d4) at ([yshift=0.4em]msa_d3.north) {\tiny 2.70\%};
    \node [ffwnode,anchor=south, fill=cyan!4.05!white] (ffw_d4) at ([yshift=0.4em]ffw_d3.north) {\tiny 4.05\%};
    \node [anchor=south west] at ([xshift=-1.25em,yshift=0.0em]msa_d4.south west) {\textbf{4}};

    
    \node [msanode,anchor=south, fill=green!0.0!white] (msa_d5) at ([yshift=0.4em]msa_d4.north) {\tiny 0.0\%};
    \node [ffwnode,anchor=south, fill=cyan!5.41!white] (ffw_d5) at ([yshift=0.4em]ffw_d4.north) {\tiny 5.41\%};
    \node [anchor=south west] at ([xshift=-1.25em,yshift=0.0em]msa_d5.south west) {\textbf{5}};

    
    \node [msanode,anchor=south, fill=green!4.05!white] (msa_d6) at ([yshift=0.4em]msa_d5.north) {\tiny 4.05\%};
    \node [ffwnode,anchor=south, fill=cyan!0.0!white] (ffw_d6) at ([yshift=0.4em]ffw_d5.north) {\tiny 0.0\%};
    \node [anchor=south west] at ([xshift=-1.25em,yshift=0.0em]msa_d6.south west) {\textbf{6}};

    
    \draw[rounded corners=10pt, thick, draw=black] ([xshift=-1.25em,yshift=-0.5em]msa_d1.south west) rectangle ([xshift=0.5em,yshift=0.5em]ffw_d6.north east);

    
    \node [msanode,anchor=south] (legend_msa) at ([yshift=2.5em]msa_d6.north) {\tiny Attn};
    \node [ffwnode,anchor=south] (legend_ffw) at ([yshift=2.5em]ffw_d6.north) {\tiny MLP};

    \draw[->, standard] ([yshift=-1.5em, xshift=-1.5em]ffw_d1.south) -- ([yshift=-0.9em, xshift=-1.5em]ffw_d1.south);
        
\end{tikzpicture}

%% file: figures/tikz/local_sparsity_echo.tex
\begin{tikzpicture}
    \tikzstyle{standard} = [rounded corners=3pt];
    \tikzstyle{ffwnode} = [minimum height=1.5em,minimum width=2.0em,inner sep=1pt,rounded corners=3pt,draw,fill=cyan!30];
    \tikzstyle{msanode} = [minimum height=1.5em,minimum width=2.0em,inner sep=1pt,rounded corners=3pt,draw,fill=green!30];
    \tikzstyle{mcanode} = [minimum height=1.5em,minimum width=2.0em,inner sep=1pt,rounded corners=3pt,draw,fill=purple!30];
    
    \node [msanode,anchor=west, fill=green!5!white] (msa1) at (0,0) {\tiny \textbf{5.0\%}};
    \node [ffwnode,anchor=west, fill=cyan!38!white] (ffw1) at ([xshift=0.75em]msa1.east) {\tiny \textbf{38\%}};
    \node [anchor=south west] at ([xshift=-1.25em,yshift=0.0em]msa1.south west) {\textbf{1}};
    
    \node [ffwnode,anchor=south, fill=cyan!29!white] (ffw2) at ([yshift=0.4em]ffw1.north) {\tiny \textbf{29\%}};
    \node [msanode,anchor=south, fill=green!23!white] (msa2) at ([yshift=0.4em]msa1.north) {\tiny \textbf{23\%}};
    \node [anchor=south west] at ([xshift=-1.25em,yshift=0.0em]msa2.south west) {\textbf{2}};
    
    \node [ffwnode,anchor=south, fill=cyan!25!white] (ffw3) at ([yshift=0.4em]ffw2.north) {\tiny \textbf{25\%}};
    \node [msanode,anchor=south, fill=green!11!white] (msa3) at ([yshift=0.4em]msa2.north) {\tiny \textbf{11\%}};
    \node [anchor=south west] at ([xshift=-1.25em,yshift=0.0em]msa3.south west) {\textbf{3}};
    
    \node [ffwnode,anchor=south, fill=cyan!25!white] (ffw4) at ([yshift=0.4em]ffw3.north) {\tiny \textbf{25\%}};
    \node [msanode,anchor=south, fill=green!8!white] (msa4) at ([yshift=0.4em]msa3.north) {\tiny \textbf{8.0\%}};
    \node [anchor=south west] at ([xshift=-1.25em,yshift=0.0em]msa4.south west) {\textbf{4}};
    
    \node [ffwnode,anchor=south, fill=cyan!19!white] (ffw5) at ([yshift=0.4em]ffw4.north) {\tiny \textbf{19\%}};
    \node [msanode,anchor=south, fill=green!7!white] (msa5) at ([yshift=0.4em]msa4.north) {\tiny \textbf{7.0\%}};
    \node [anchor=south west] at ([xshift=-1.25em,yshift=0.0em]msa5.south west) {\textbf{5}};
    
    \node [ffwnode,anchor=south, fill=cyan!18!white] (ffw6) at ([yshift=0.4em]ffw5.north) {\tiny \textbf{18\%}};
    \node [msanode,anchor=south, fill=green!13!white] (msa6) at ([yshift=0.4em]msa5.north) {\tiny \textbf{13\%}};
    \node [anchor=south west] at ([xshift=-1.25em,yshift=0.0em]msa6.south west) {\textbf{6}};
    
    \node [anchor=south] (encoder) at ([xshift=0.5em,yshift=0.5em]msa6.north) {\textbf{Encoder}};
    
    \draw[rounded corners=10pt, thick, draw=black] ([xshift=-1.25em,yshift=-0.5em]msa1.south west) rectangle ([xshift=.5em,yshift=0.5em]ffw6.north east);
        
    \node [msanode,anchor=west, fill=green!0!white] (msa_d1) at ([xshift=6em]msa1.east) {\tiny \textbf{0.0\%}};
    \node [mcanode,anchor=west, fill=purple!25!white] (mca_d1) at ([xshift=0.75em]msa_d1.east) {\tiny \textbf{25\%}};
    \node [ffwnode,anchor=west, fill=cyan!18!white] (ffw_d1) at ([xshift=0.75em]mca_d1.east) {\tiny \textbf{18\%}};
    
    \node [ffwnode,anchor=south, fill=cyan!19!white] (ffw_d2) at ([yshift=0.4em]ffw_d1.north) {\tiny \textbf{19\%}};
    \node [msanode,anchor=south, fill=green!2!white] (msa_d2) at ([yshift=0.4em]msa_d1.north) {\tiny \textbf{2.0\%}};
    \node [mcanode,anchor=south, fill=purple!18!white] (mca_d2) at ([yshift=0.4em]mca_d1.north) {\tiny \textbf{18\%}};
        
    \node [ffwnode,anchor=south, fill=cyan!3!white] (ffw_d3) at ([yshift=0.4em]ffw_d2.north) {\tiny \textbf{4.0\%}};
    \node [msanode,anchor=south, fill=green!2!white] (msa_d3) at ([yshift=0.4em]msa_d2.north) {\tiny \textbf{2.0\%}};
    \node [mcanode,anchor=south, fill=purple!19!white] (mca_d3) at ([yshift=0.4em]mca_d2.north) {\tiny \textbf{19\%}};
    
    \node [ffwnode,anchor=south, fill=cyan!3!white] (ffw_d4) at ([yshift=0.4em]ffw_d3.north) {\tiny \textbf{3.0\%}};
    \node [msanode,anchor=south, fill=green!4!white] (msa_d4) at ([yshift=0.4em]msa_d3.north) {\tiny \textbf{4.0\%}};
    \node [mcanode,anchor=south, fill=purple!32!white] (mca_d4) at ([yshift=0.4em]mca_d3.north) {\tiny \textbf{32\%}};
    
    \node [ffwnode,anchor=south, fill=cyan!2!white] (ffw_d5) at ([yshift=0.4em]ffw_d4.north) {\tiny \textbf{2.0\%}};
    \node [msanode,anchor=south, fill=green!2!white] (msa_d5) at ([yshift=0.4em]msa_d4.north) {\tiny \textbf{2.0\%}};
    \node [mcanode,anchor=south, fill=purple!28!white] (mca_d5) at ([yshift=0.4em]mca_d4.north) {\tiny \textbf{28\%}};
    
    \node [ffwnode,anchor=south, fill=cyan!0!white] (ffw_d6) at ([yshift=0.4em]ffw_d5.north) {\tiny \textbf{0.0\%}};
    \node [msanode,anchor=south, fill=green!2!white] (msa_d6) at ([yshift=0.4em]msa_d5.north) {\tiny \textbf{2.0\%}};
    \node [mcanode,anchor=south, fill=purple!45!white] (mca_d6) at ([yshift=0.4em]mca_d5.north) {\tiny \textbf{45\%}};
    
    \node [anchor=south] (decoder) at ([xshift=1.0em,yshift=0.5em]msa_d6.north) {\textbf{Decoder}};
    
    \draw[rounded corners=10pt, thick, draw=black] ([xshift=-0.5em,yshift=-0.5em]msa_d1.south west) rectangle ([xshift=0.5em,yshift=0.5em]ffw_d6.north east);

    \draw[->,standard] ([xshift=0.5em,yshift=0em]ffw4.east) -- ([xshift=-.5em,yshift=0em]msa_d4.west);

    \node [ffwnode,anchor=south] (legend_ffw_encoder) at ([yshift=2.5em]ffw6.north) {\tiny \textbf{FF}};
    \node [msanode,anchor=south] (legend_msa_encoder) at ([yshift=2.5em]msa6.north) {\tiny \textbf{MHSA}};
    
    \node [ffwnode,anchor=south] (legend_ffw) at ([yshift=2.5em]ffw_d6.north) {\tiny \textbf{FF}};
    \node [msanode,anchor=south] (legend_msa) at ([yshift=2.5em]msa_d6.north) {\tiny \textbf{MHSA}};
    \node [mcanode,anchor=south] (legend_mca) at ([yshift=2.5em]mca_d6.north) {\tiny \textbf{MHCA}};

    \draw[->, standard] ([yshift=-1.5em, xshift=1em]msa1.south) -- ([yshift=-0.5em, xshift=1em]msa1.south);
    \draw[->, standard] ([yshift=-1.5em]mca_d1.south) -- ([yshift=-0.5em]mca_d1.south);
        
\end{tikzpicture}

%% file: figures/tikz/local_sparsity_repeat.tex
\begin{tikzpicture}
    \tikzstyle{standard} = [rounded corners=3pt];
    \tikzstyle{ffwnode} = [minimum height=1.5em,minimum width=2.0em,inner sep=1pt,rounded corners=3pt,draw,fill=cyan!30];
    \tikzstyle{msanode} = [minimum height=1.5em,minimum width=2.0em,inner sep=1pt,rounded corners=3pt,draw,fill=green!30];
    \tikzstyle{mcanode} = [minimum height=1.5em,minimum width=2.0em,inner sep=1pt,rounded corners=3pt,draw,fill=purple!30];
    
    \node [msanode,anchor=west, fill=green!14!white] (msa1) at (0,0) {\tiny \textbf{14\%}};
    \node [ffwnode,anchor=west, fill=cyan!47!white] (ffw1) at ([xshift=0.75em]msa1.east) {\tiny \textbf{47\%}};
    \node [anchor=south west] at ([xshift=-1.25em,yshift=0.0em]ffw1.south west) {\textbf{1}};
    
    \node [ffwnode,anchor=south, fill=cyan!36!white] (ffw2) at ([yshift=0.4em]ffw1.north) {\tiny \textbf{36\%}};
    \node [msanode,anchor=south, fill=green!31!white] (msa2) at ([yshift=0.4em]msa1.north) {\tiny \textbf{31\%}};
    \node [anchor=south west] at ([xshift=-1.25em,yshift=0.0em]msa2.south west) {\textbf{2}};
    
    \node [ffwnode,anchor=south, fill=cyan!31!white] (ffw3) at ([yshift=0.4em]ffw2.north) {\tiny \textbf{31\%}};
    \node [msanode,anchor=south, fill=green!10!white] (msa3) at ([yshift=0.4em]msa2.north) {\tiny \textbf{10\%}};
    \node [anchor=south west] at ([xshift=-1.25em,yshift=0.0em]msa3.south west) {\textbf{3}};
    
    \node [ffwnode,anchor=south, fill=cyan!34!white] (ffw4) at ([yshift=0.4em]ffw3.north) {\tiny \textbf{34\%}};
    \node [msanode,anchor=south, fill=green!15!white] (msa4) at ([yshift=0.4em]msa3.north) {\tiny \textbf{15\%}};
    \node [anchor=south west] at ([xshift=-1.25em,yshift=0.0em]msa4.south west) {\textbf{4}};
    
    \node [ffwnode,anchor=south, fill=cyan!35!white] (ffw5) at ([yshift=0.4em]ffw4.north) {\tiny \textbf{35\%}};
    \node [msanode,anchor=south, fill=green!10!white] (msa5) at ([yshift=0.4em]msa4.north) {\tiny \textbf{10\%}};
    \node [anchor=south west] at ([xshift=-1.25em,yshift=0.0em]msa5.south west) {\textbf{5}};
    
    \node [ffwnode,anchor=south, fill=cyan!36!white] (ffw6) at ([yshift=0.4em]ffw5.north) {\tiny \textbf{36\%}};
    \node [msanode,anchor=south, fill=green!2!white] (msa6) at ([yshift=0.4em]msa5.north) {\tiny \textbf{2.0\%}};
    \node [anchor=south west] at ([xshift=-1.25em,yshift=0.0em]msa6.south west) {\textbf{6}};
    
    \node [anchor=south] (encoder) at ([xshift=0.5em,yshift=0.5em]msa6.north) {\textbf{Encoder}};
    
    \draw[rounded corners=10pt, thick, draw=black] ([xshift=-1.25em,yshift=-0.5em]msa1.south west) rectangle ([xshift=.5em,yshift=0.5em]ffw6.north east);
        
    \node [msanode,anchor=west, fill=green!0!white] (msa_d1) at ([xshift=6em]msa1.east) {\tiny \textbf{0.0\%}};
    \node [mcanode,anchor=west, fill=purple!47!white] (mca_d1) at ([xshift=0.75em]msa_d1.east) {\tiny \textbf{47\%}};
    \node [ffwnode,anchor=west, fill=cyan!37!white] (ffw_d1) at ([xshift=0.75em]mca_d1.east) {\tiny \textbf{37\%}};
    
    \node [ffwnode,anchor=south, fill=cyan!27!white] (ffw_d2) at ([yshift=0.4em]ffw_d1.north) {\tiny \textbf{27\%}};
    \node [msanode,anchor=south, fill=green!16!white] (msa_d2) at ([yshift=0.4em]msa_d1.north) {\tiny \textbf{16\%}};
    \node [mcanode,anchor=south, fill=purple!27!white] (mca_d2) at ([yshift=0.4em]mca_d1.north) {\tiny \textbf{27\%}};
        
    \node [ffwnode,anchor=south, fill=cyan!8!white] (ffw_d3) at ([yshift=0.4em]ffw_d2.north) {\tiny \textbf{8.0\%}};
    \node [msanode,anchor=south, fill=green!12!white] (msa_d3) at ([yshift=0.4em]msa_d2.north) {\tiny \textbf{12\%}};
    \node [mcanode,anchor=south, fill=purple!17!white] (mca_d3) at ([yshift=0.4em]mca_d2.north) {\tiny \textbf{17\%}};
    
    \node [ffwnode,anchor=south, fill=cyan!1!white] (ffw_d4) at ([yshift=0.4em]ffw_d3.north) {\tiny \textbf{1.0\%}};
    \node [msanode,anchor=south, fill=green!3!white] (msa_d4) at ([yshift=0.4em]msa_d3.north) {\tiny \textbf{3.0\%}};
    \node [mcanode,anchor=south, fill=purple!32!white] (mca_d4) at ([yshift=0.4em]mca_d3.north) {\tiny \textbf{32\%}};
    
    \node [ffwnode,anchor=south, fill=cyan!1!white] (ffw_d5) at ([yshift=0.4em]ffw_d4.north) {\tiny \textbf{1.0\%}};
    \node [msanode,anchor=south, fill=green!1!white] (msa_d5) at ([yshift=0.4em]msa_d4.north) {\tiny \textbf{1.0\%}};
    \node [mcanode,anchor=south, fill=purple!26!white] (mca_d5) at ([yshift=0.4em]mca_d4.north) {\tiny \textbf{26\%}};
    
    \node [ffwnode,anchor=south, fill=cyan!0!white] (ffw_d6) at ([yshift=0.4em]ffw_d5.north) {\tiny \textbf{0.0\%}};
    \node [msanode,anchor=south, fill=green!1!white] (msa_d6) at ([yshift=0.4em]msa_d5.north) {\tiny \textbf{1.0\%}};
    \node [mcanode,anchor=south, fill=purple!43!white] (mca_d6) at ([yshift=0.4em]mca_d5.north) {\tiny \textbf{43\%}};
    
    \node [anchor=south] (decoder) at ([xshift=1.0em,yshift=0.5em]msa_d6.north) {\textbf{Decoder}};
    
    \draw[rounded corners=10pt, thick, draw=black] ([xshift=-0.5em,yshift=-0.5em]msa_d1.south west) rectangle ([xshift=0.5em,yshift=0.5em]ffw_d6.north east);

    \draw[->,standard] ([xshift=0.5em,yshift=0em]ffw4.east) -- ([xshift=-.5em,yshift=0em]msa_d4.west);

    \node [ffwnode,anchor=south] (legend_ffw_encoder) at ([yshift=2.5em]ffw6.north) {\tiny \textbf{FF}};
    \node [msanode,anchor=south] (legend_msa_encoder) at ([yshift=2.5em]msa6.north) {\tiny \textbf{MHSA}};
    
    \node [ffwnode,anchor=south] (legend_ffw) at ([yshift=2.5em]ffw_d6.north) {\tiny \textbf{FF}};
    \node [msanode,anchor=south] (legend_msa) at ([yshift=2.5em]msa_d6.north) {\tiny \textbf{MHSA}};
    \node [mcanode,anchor=south] (legend_mca) at ([yshift=2.5em]mca_d6.north) {\tiny \textbf{MHCA}};

    \draw[->, standard] ([yshift=-1.5em, xshift=1em]msa1.south) -- ([yshift=-0.5em, xshift=1em]msa1.south);
    \draw[->, standard] ([yshift=-1.5em]mca_d1.south) -- ([yshift=-0.5em]mca_d1.south);
        
\end{tikzpicture}

%% file: figures/tikz/local_sparsity_reverse.tex
\begin{tikzpicture}
    \tikzstyle{standard} = [rounded corners=3pt];
    \tikzstyle{ffwnode} = [minimum height=1.5em,minimum width=2.0em,inner sep=1pt,rounded corners=3pt,draw,fill=cyan!30];
    \tikzstyle{msanode} = [minimum height=1.5em,minimum width=2.0em,inner sep=1pt,rounded corners=3pt,draw,fill=green!30];
    \tikzstyle{mcanode} = [minimum height=1.5em,minimum width=2.0em,inner sep=1pt,rounded corners=3pt,draw,fill=purple!30];
    
    \node [msanode,anchor=west, fill=green!10!white] (msa1) at (0,0) {\tiny \textbf{10\%}};
    \node [ffwnode,anchor=west, fill=cyan!44!white] (ffw1) at ([xshift=0.75em]msa1.east) {\tiny \textbf{44 \%}};
    \node [anchor=south west] at ([xshift=-1.25em,yshift=0.0em]msa1.south west) {\textbf{1}};
    
    \node [ffwnode,anchor=south, fill=cyan!38!white] (ffw2) at ([yshift=0.4em]ffw1.north) {\tiny \textbf{38\%}};
    \node [msanode,anchor=south, fill=green!29!white] (msa2) at ([yshift=0.4em]msa1.north) {\tiny \textbf{29\%}};
    \node [anchor=south west] at ([xshift=-1.25em,yshift=0.0em]msa2.south west) {\textbf{2}};
    
    \node [ffwnode,anchor=south, fill=cyan!31!white] (ffw3) at ([yshift=0.4em]ffw2.north) {\tiny \textbf{31\%}};
    \node [msanode,anchor=south, fill=green!19!white] (msa3) at ([yshift=0.4em]msa2.north) {\tiny \textbf{19\%}};
    \node [anchor=south west] at ([xshift=-1.25em,yshift=0.0em]msa3.south west) {\textbf{3}};
    
    \node [ffwnode,anchor=south, fill=cyan!29!white] (ffw4) at ([yshift=0.4em]ffw3.north) {\tiny \textbf{29\%}};
    \node [msanode,anchor=south, fill=green!19!white] (msa4) at ([yshift=0.4em]msa3.north) {\tiny \textbf{19\%}};
    \node [anchor=south west] at ([xshift=-1.25em,yshift=0.0em]msa4.south west) {\textbf{4}};
    
    \node [ffwnode,anchor=south, fill=cyan!31!white] (ffw5) at ([yshift=0.4em]ffw4.north) {\tiny \textbf{31\%}};
    \node [msanode,anchor=south, fill=green!10!white] (msa5) at ([yshift=0.4em]msa4.north) {\tiny \textbf{10\%}};
    \node [anchor=south west] at ([xshift=-1.25em,yshift=0.0em]msa5.south west) {\textbf{5}};
    
    \node [ffwnode,anchor=south, fill=cyan!31!white] (ffw6) at ([yshift=0.4em]ffw5.north) {\tiny \textbf{31\%}};
    \node [msanode,anchor=south, fill=green!6!white] (msa6) at ([yshift=0.4em]msa5.north) {\tiny \textbf{6.0\%}};
    \node [anchor=south west] at ([xshift=-1.25em,yshift=0.0em]msa6.south west) {\textbf{6}};
    
    \node [anchor=south] (encoder) at ([xshift=0.5em,yshift=0.5em]msa6.north) {\textbf{Encoder}};
    
    \draw[rounded corners=10pt, thick, draw=black] ([xshift=-1.25em,yshift=-0.5em]msa1.south west) rectangle ([xshift=.5em,yshift=0.5em]ffw6.north east);
        
    \node [msanode,anchor=west, fill=green!0!white] (msa_d1) at ([xshift=6em]msa1.east) {\tiny \textbf{0.0\%}};
    \node [mcanode,anchor=west, fill=purple!24!white] (mca_d1) at ([xshift=0.75em]msa_d1.east) {\tiny \textbf{24\%}};
    \node [ffwnode,anchor=west, fill=cyan!18!white] (ffw_d1) at ([xshift=0.75em]mca_d1.east) {\tiny \textbf{18\%}};
    
    \node [ffwnode,anchor=south, fill=cyan!15!white] (ffw_d2) at ([yshift=0.4em]ffw_d1.north) {\tiny \textbf{15\%}};
    \node [msanode,anchor=south, fill=green!4!white] (msa_d2) at ([yshift=0.4em]msa_d1.north) {\tiny \textbf{4.0\%}};
    \node [mcanode,anchor=south, fill=purple!15!white] (mca_d2) at ([yshift=0.4em]mca_d1.north) {\tiny \textbf{15\%}};
        
    \node [ffwnode,anchor=south, fill=cyan!3!white] (ffw_d3) at ([yshift=0.4em]ffw_d2.north) {\tiny \textbf{3.0\%}};
    \node [msanode,anchor=south, fill=green!3!white] (msa_d3) at ([yshift=0.4em]msa_d2.north) {\tiny \textbf{3.0\%}};
    \node [mcanode,anchor=south, fill=purple!16!white] (mca_d3) at ([yshift=0.4em]mca_d2.north) {\tiny \textbf{16\%}};
    
    \node [ffwnode,anchor=south, fill=cyan!2!white] (ffw_d4) at ([yshift=0.4em]ffw_d3.north) {\tiny \textbf{2.0\%}};
    \node [msanode,anchor=south, fill=green!1!white] (msa_d4) at ([yshift=0.4em]msa_d3.north) {\tiny \textbf{1.0\%}};
    \node [mcanode,anchor=south, fill=purple!25!white] (mca_d4) at ([yshift=0.4em]mca_d3.north) {\tiny \textbf{25\%}};
    
    \node [ffwnode,anchor=south, fill=cyan!1!white] (ffw_d5) at ([yshift=0.4em]ffw_d4.north) {\tiny \textbf{1.0\%}};
    \node [msanode,anchor=south, fill=green!1!white] (msa_d5) at ([yshift=0.4em]msa_d4.north) {\tiny \textbf{1.0\%}};
    \node [mcanode,anchor=south, fill=purple!22!white] (mca_d5) at ([yshift=0.4em]mca_d4.north) {\tiny \textbf{22\%}};
    
    \node [ffwnode,anchor=south, fill=cyan!1!white] (ffw_d6) at ([yshift=0.4em]ffw_d5.north) {\tiny \textbf{1.0\%}};
    \node [msanode,anchor=south, fill=green!2!white] (msa_d6) at ([yshift=0.4em]msa_d5.north) {\tiny \textbf{2.0\%}};
    \node [mcanode,anchor=south, fill=purple!42!white] (mca_d6) at ([yshift=0.4em]mca_d5.north) {\tiny \textbf{42\%}};
    
    \node [anchor=south] (decoder) at ([xshift=1.0em,yshift=0.5em]msa_d6.north) {\textbf{Decoder}};
    
    \draw[rounded corners=10pt, thick, draw=black] ([xshift=-0.5em,yshift=-0.5em]msa_d1.south west) rectangle ([xshift=0.5em,yshift=0.5em]ffw_d6.north east);

    \draw[->,standard] ([xshift=0.5em,yshift=0em]ffw4.east) -- ([xshift=-.5em,yshift=0em]msa_d4.west);

    \node [ffwnode,anchor=south] (legend_ffw_encoder) at ([yshift=2.5em]ffw6.north) {\tiny \textbf{FF}};
    \node [msanode,anchor=south] (legend_msa_encoder) at ([yshift=2.5em]msa6.north) {\tiny \textbf{MHSA}};
    
    \node [ffwnode,anchor=south] (legend_ffw) at ([yshift=2.5em]ffw_d6.north) {\tiny \textbf{FF}};
    \node [msanode,anchor=south] (legend_msa) at ([yshift=2.5em]msa_d6.north) {\tiny \textbf{MHSA}};
    \node [mcanode,anchor=south] (legend_mca) at ([yshift=2.5em]mca_d6.north) {\tiny \textbf{MHCA}};

    \draw[->, standard] ([yshift=-1.5em, xshift=1em]msa1.south) -- ([yshift=-0.5em, xshift=1em]msa1.south);
    \draw[->, standard] ([yshift=-1.5em]mca_d1.south) -- ([yshift=-0.5em]mca_d1.south);
        
\end{tikzpicture}

%% file: figures/tikz/local_sparsity_swap.tex
\begin{tikzpicture}
    \tikzstyle{standard} = [rounded corners=3pt];
    \tikzstyle{ffwnode} = [minimum height=1.5em,minimum width=2.0em,inner sep=1pt,rounded corners=3pt,draw,fill=cyan!30];
    \tikzstyle{msanode} = [minimum height=1.5em,minimum width=2.0em,inner sep=1pt,rounded corners=3pt,draw,fill=green!30];
    \tikzstyle{mcanode} = [minimum height=1.5em,minimum width=2.0em,inner sep=1pt,rounded corners=3pt,draw,fill=purple!30];
    
    \node [msanode,anchor=west, fill=green!4!white] (msa1) at (0,0) {\tiny \textbf{4.0\%}};
    \node [ffwnode,anchor=west, fill=cyan!41!white] (ffw1) at ([xshift=0.75em]msa1.east) {\tiny \textbf{41\%}};
    \node [anchor=south west] at ([xshift=-1.25em,yshift=0.0em]msa1.south west) {\textbf{1}};
    
    \node [ffwnode,anchor=south, fill=cyan!27!white] (ffw2) at ([yshift=0.4em]ffw1.north) {\tiny \textbf{27\%}};
    \node [msanode,anchor=south, fill=green!28!white] (msa2) at ([yshift=0.4em]msa1.north) {\tiny \textbf{28\%}};
    \node [anchor=south west] at ([xshift=-1.25em,yshift=0.0em]msa2.south west) {\textbf{2}};
    
    \node [ffwnode,anchor=south, fill=cyan!31!white] (ffw3) at ([yshift=0.4em]ffw2.north) {\tiny \textbf{31\%}};
    \node [msanode,anchor=south, fill=green!12!white] (msa3) at ([yshift=0.4em]msa2.north) {\tiny \textbf{12\%}};
    \node [anchor=south west] at ([xshift=-1.25em,yshift=0.0em]msa3.south west) {\textbf{3}};
    
    \node [ffwnode,anchor=south, fill=cyan!33!white] (ffw4) at ([yshift=0.4em]ffw3.north) {\tiny \textbf{33\%}};
    \node [msanode,anchor=south, fill=green!11!white] (msa4) at ([yshift=0.4em]msa3.north) {\tiny \textbf{11\%}};
    \node [anchor=south west] at ([xshift=-1.25em,yshift=0.0em]msa4.south west) {\textbf{4}};
    
    \node [ffwnode,anchor=south, fill=cyan!29!white] (ffw5) at ([yshift=0.4em]ffw4.north) {\tiny \textbf{29\%}};
    \node [msanode,anchor=south, fill=green!14!white] (msa5) at ([yshift=0.4em]msa4.north) {\tiny \textbf{14\%}};
    \node [anchor=south west] at ([xshift=-1.25em,yshift=0.0em]msa5.south west) {\textbf{5}};
    
    \node [ffwnode,anchor=south, fill=cyan!32!white] (ffw6) at ([yshift=0.4em]ffw5.north) {\tiny \textbf{32\%}};
    \node [msanode,anchor=south, fill=green!9!white] (msa6) at ([yshift=0.4em]msa5.north) {\tiny \textbf{9.0\%}};
    \node [anchor=south west] at ([xshift=-1.25em,yshift=0.0em]msa6.south west) {\textbf{6}};
    
    \node [anchor=south] (encoder) at ([xshift=0.5em,yshift=0.5em]msa6.north) {\textbf{Encoder}};
    
    \draw[rounded corners=10pt, thick, draw=black] ([xshift=-1.25em,yshift=-0.5em]msa1.south west) rectangle ([xshift=.5em,yshift=0.5em]ffw6.north east);
        
    \node [msanode,anchor=west, fill=green!0!white] (msa_d1) at ([xshift=6em]msa1.east) {\tiny \textbf{0.0\%}};
    \node [mcanode,anchor=west, fill=purple!23!white] (mca_d1) at ([xshift=0.75em]msa_d1.east) {\tiny \textbf{23\%}};
    \node [ffwnode,anchor=west, fill=cyan!20!white] (ffw_d1) at ([xshift=0.75em]mca_d1.east) {\tiny \textbf{20\%}};
    
    \node [ffwnode,anchor=south, fill=cyan!13!white] (ffw_d2) at ([yshift=0.4em]ffw_d1.north) {\tiny \textbf{13\%}};
    \node [msanode,anchor=south, fill=green!4!white] (msa_d2) at ([yshift=0.4em]msa_d1.north) {\tiny \textbf{4.0\%}};
    \node [mcanode,anchor=south, fill=purple!18!white] (mca_d2) at ([yshift=0.4em]mca_d1.north) {\tiny \textbf{18\%}};
        
    \node [ffwnode,anchor=south, fill=cyan!3!white] (ffw_d3) at ([yshift=0.4em]ffw_d2.north) {\tiny \textbf{3.0\%}};
    \node [msanode,anchor=south, fill=green!2!white] (msa_d3) at ([yshift=0.4em]msa_d2.north) {\tiny \textbf{2.0\%}};
    \node [mcanode,anchor=south, fill=purple!16!white] (mca_d3) at ([yshift=0.4em]mca_d2.north) {\tiny \textbf{16\%}};
    
    \node [ffwnode,anchor=south, fill=cyan!1!white] (ffw_d4) at ([yshift=0.4em]ffw_d3.north) {\tiny \textbf{1.0\%}};
    \node [msanode,anchor=south, fill=green!2!white] (msa_d4) at ([yshift=0.4em]msa_d3.north) {\tiny \textbf{2.0\%}};
    \node [mcanode,anchor=south, fill=purple!29!white] (mca_d4) at ([yshift=0.4em]mca_d3.north) {\tiny \textbf{29\%}};
    
    \node [ffwnode,anchor=south, fill=cyan!1!white] (ffw_d5) at ([yshift=0.4em]ffw_d4.north) {\tiny \textbf{1.0\%}};
    \node [msanode,anchor=south, fill=green!1!white] (msa_d5) at ([yshift=0.4em]msa_d4.north) {\tiny \textbf{1.0\%}};
    \node [mcanode,anchor=south, fill=purple!21!white] (mca_d5) at ([yshift=0.4em]mca_d4.north) {\tiny \textbf{21\%}};
    
    \node [ffwnode,anchor=south, fill=cyan!1!white] (ffw_d6) at ([yshift=0.4em]ffw_d5.north) {\tiny \textbf{1.0\%}};
    \node [msanode,anchor=south, fill=green!1!white] (msa_d6) at ([yshift=0.4em]msa_d5.north) {\tiny \textbf{1.0\%}};
    \node [mcanode,anchor=south, fill=purple!40!white] (mca_d6) at ([yshift=0.4em]mca_d5.north) {\tiny \textbf{40\%}};
    
    \node [anchor=south] (decoder) at ([xshift=1.0em,yshift=0.5em]msa_d6.north) {\textbf{Decoder}};
    
    \draw[rounded corners=10pt, thick, draw=black] ([xshift=-0.5em,yshift=-0.5em]msa_d1.south west) rectangle ([xshift=0.5em,yshift=0.5em]ffw_d6.north east);

    \draw[->,standard] ([xshift=0.5em,yshift=0em]ffw4.east) -- ([xshift=-.5em,yshift=0em]msa_d4.west);

    \node [ffwnode,anchor=south] (legend_ffw_encoder) at ([yshift=2.5em]ffw6.north) {\tiny \textbf{FF}};
    \node [msanode,anchor=south] (legend_msa_encoder) at ([yshift=2.5em]msa6.north) {\tiny \textbf{MHSA}};
    
    \node [ffwnode,anchor=south] (legend_ffw) at ([yshift=2.5em]ffw_d6.north) {\tiny \textbf{FF}};
    \node [msanode,anchor=south] (legend_msa) at ([yshift=2.5em]msa_d6.north) {\tiny \textbf{MHSA}};
    \node [mcanode,anchor=south] (legend_mca) at ([yshift=2.5em]mca_d6.north) {\tiny \textbf{MHCA}};

    \draw[->, standard] ([yshift=-1.5em, xshift=1em]msa1.south) -- ([yshift=-0.5em, xshift=1em]msa1.south);
    \draw[->, standard] ([yshift=-1.5em]mca_d1.south) -- ([yshift=-0.5em]mca_d1.south);
        
\end{tikzpicture}

%% file: figures/tikz/local_sparsity_shift.tex
\begin{tikzpicture}
    \tikzstyle{standard} = [rounded corners=3pt];
    \tikzstyle{ffwnode} = [minimum height=1.5em,minimum width=2.0em,inner sep=1pt,rounded corners=3pt,draw,fill=cyan!30];
    \tikzstyle{msanode} = [minimum height=1.5em,minimum width=2.0em,inner sep=1pt,rounded corners=3pt,draw,fill=green!30];
    \tikzstyle{mcanode} = [minimum height=1.5em,minimum width=2.0em,inner sep=1pt,rounded corners=3pt,draw,fill=purple!30];
    
    \node [msanode,anchor=west, fill=green!10!white] (msa1) at (0,0) {\tiny \textbf{10\%}};
    \node [ffwnode,anchor=west, fill=cyan!42!white] (ffw1) at ([xshift=0.75em]msa1.east) {\tiny \textbf{42\%}};
    \node [anchor=south west] at ([xshift=-1.25em,yshift=0.0em]msa1.south west) {\textbf{1}};
    
    \node [ffwnode,anchor=south, fill=cyan!32!white] (ffw2) at ([yshift=0.4em]ffw1.north) {\tiny \textbf{32\%}};
    \node [msanode,anchor=south, fill=green!20!white] (msa2) at ([yshift=0.4em]msa1.north) {\tiny \textbf{20\%}};
    \node [anchor=south west] at ([xshift=-1.25em,yshift=0.0em]msa2.south west) {\textbf{2}};
    
    \node [ffwnode,anchor=south, fill=cyan!28!white] (ffw3) at ([yshift=0.4em]ffw2.north) {\tiny \textbf{28\%}};
    \node [msanode,anchor=south, fill=green!19!white] (msa3) at ([yshift=0.4em]msa2.north) {\tiny \textbf{19\%}};
    \node [anchor=south west] at ([xshift=-1.25em,yshift=0.0em]msa3.south west) {\textbf{3}};
    
    \node [ffwnode,anchor=south, fill=cyan!29!white] (ffw4) at ([yshift=0.4em]ffw3.north) {\tiny \textbf{29\%}};
    \node [msanode,anchor=south, fill=green!13!white] (msa4) at ([yshift=0.4em]msa3.north) {\tiny \textbf{13\%}};
    \node [anchor=south west] at ([xshift=-1.25em,yshift=0.0em]msa4.south west) {\textbf{4}};
    
    \node [ffwnode,anchor=south, fill=cyan!25!white] (ffw5) at ([yshift=0.4em]ffw4.north) {\tiny \textbf{25\%}};
    \node [msanode,anchor=south, fill=green!13!white] (msa5) at ([yshift=0.4em]msa4.north) {\tiny \textbf{13\%}};
    \node [anchor=south west] at ([xshift=-1.25em,yshift=0.0em]msa5.south west) {\textbf{5}};
    
    \node [ffwnode,anchor=south, fill=cyan!35!white] (ffw6) at ([yshift=0.4em]ffw5.north) {\tiny \textbf{35\%}};
    \node [msanode,anchor=south, fill=green!13!white] (msa6) at ([yshift=0.4em]msa5.north) {\tiny \textbf{17\%}};
    \node [anchor=south west] at ([xshift=-1.25em,yshift=0.0em]msa6.south west) {\textbf{6}};
    
    \node [anchor=south] (encoder) at ([xshift=0.5em,yshift=0.5em]msa6.north) {\textbf{Encoder}};
    
    \draw[rounded corners=10pt, thick, draw=black] ([xshift=-1.25em,yshift=-0.5em]msa1.south west) rectangle ([xshift=.5em,yshift=0.5em]ffw6.north east);
        
    \node [msanode,anchor=west, fill=green!0!white] (msa_d1) at ([xshift=6em]msa1.east) {\tiny \textbf{0.0\%}};
    \node [mcanode,anchor=west, fill=purple!24!white] (mca_d1) at ([xshift=0.75em]msa_d1.east) {\tiny \textbf{24\%}};
    \node [ffwnode,anchor=west, fill=cyan!20!white] (ffw_d1) at ([xshift=0.75em]mca_d1.east) {\tiny \textbf{20\%}};
    
    \node [ffwnode,anchor=south, fill=cyan!19!white] (ffw_d2) at ([yshift=0.4em]ffw_d1.north) {\tiny \textbf{19\%}};
    \node [msanode,anchor=south, fill=green!3!white] (msa_d2) at ([yshift=0.4em]msa_d1.north) {\tiny \textbf{3.0\%}};
    \node [mcanode,anchor=south, fill=purple!21!white] (mca_d2) at ([yshift=0.4em]mca_d1.north) {\tiny \textbf{21\%}};
        
    \node [ffwnode,anchor=south, fill=cyan!2!white] (ffw_d3) at ([yshift=0.4em]ffw_d2.north) {\tiny \textbf{3.0\%}};
    \node [msanode,anchor=south, fill=green!3!white] (msa_d3) at ([yshift=0.4em]msa_d2.north) {\tiny \textbf{3.0\%}};
    \node [mcanode,anchor=south, fill=purple!18!white] (mca_d3) at ([yshift=0.4em]mca_d2.north) {\tiny \textbf{18\%}};
    
    \node [ffwnode,anchor=south, fill=cyan!2!white] (ffw_d4) at ([yshift=0.4em]ffw_d3.north) {\tiny \textbf{2.0\%}};
    \node [msanode,anchor=south, fill=green!3!white] (msa_d4) at ([yshift=0.4em]msa_d3.north) {\tiny \textbf{3.0\%}};
    \node [mcanode,anchor=south, fill=purple!27!white] (mca_d4) at ([yshift=0.4em]mca_d3.north) {\tiny \textbf{27\%}};
    
    \node [ffwnode,anchor=south, fill=cyan!1!white] (ffw_d5) at ([yshift=0.4em]ffw_d4.north) {\tiny \textbf{1.0\%}};
    \node [msanode,anchor=south, fill=green!3!white] (msa_d5) at ([yshift=0.4em]msa_d4.north) {\tiny \textbf{3.0\%}};
    \node [mcanode,anchor=south, fill=purple!22!white] (mca_d5) at ([yshift=0.4em]mca_d4.north) {\tiny \textbf{22\%}};
    
    \node [ffwnode,anchor=south, fill=cyan!1!white] (ffw_d6) at ([yshift=0.4em]ffw_d5.north) {\tiny \textbf{1.0\%}};
    \node [msanode,anchor=south, fill=green!2!white] (msa_d6) at ([yshift=0.4em]msa_d5.north) {\tiny \textbf{2.0\%}};
    \node [mcanode,anchor=south, fill=purple!37!white] (mca_d6) at ([yshift=0.4em]mca_d5.north) {\tiny \textbf{37\%}};
    
    \node [anchor=south] (decoder) at ([xshift=1.0em,yshift=0.5em]msa_d6.north) {\textbf{Decoder}};
    
    \draw[rounded corners=10pt, thick, draw=black] ([xshift=-0.5em,yshift=-0.5em]msa_d1.south west) rectangle ([xshift=0.5em,yshift=0.5em]ffw_d6.north east);

    \draw[->,standard] ([xshift=0.5em,yshift=0em]ffw4.east) -- ([xshift=-.5em,yshift=0em]msa_d4.west);

    \node [ffwnode,anchor=south] (legend_ffw_encoder) at ([yshift=2.5em]ffw6.north) {\tiny \textbf{FF}};
    \node [msanode,anchor=south] (legend_msa_encoder) at ([yshift=2.5em]msa6.north) {\tiny \textbf{MHSA}};
    
    \node [ffwnode,anchor=south] (legend_ffw) at ([yshift=2.5em]ffw_d6.north) {\tiny \textbf{FF}};
    \node [msanode,anchor=south] (legend_msa) at ([yshift=2.5em]msa_d6.north) {\tiny \textbf{MHSA}};
    \node [mcanode,anchor=south] (legend_mca) at ([yshift=2.5em]mca_d6.north) {\tiny \textbf{MHCA}};

    \draw[->, standard] ([yshift=-1.5em, xshift=1em]msa1.south) -- ([yshift=-0.5em, xshift=1em]msa1.south);
    \draw[->, standard] ([yshift=-1.5em]mca_d1.south) -- ([yshift=-0.5em]mca_d1.south);
        
\end{tikzpicture}

%% file: figures/tikz/local_sparsity_append.tex
\begin{tikzpicture}
    \tikzstyle{standard} = [rounded corners=3pt];
    \tikzstyle{ffwnode} = [minimum height=1.5em,minimum width=2.0em,inner sep=1pt,rounded corners=3pt,draw,fill=cyan!30];
    \tikzstyle{msanode} = [minimum height=1.5em,minimum width=2.0em,inner sep=1pt,rounded corners=3pt,draw,fill=green!30];
    \tikzstyle{mcanode} = [minimum height=1.5em,minimum width=2.0em,inner sep=1pt,rounded corners=3pt,draw,fill=purple!30];

    \node [msanode,anchor=west, fill=green!30!white] (msa1) at (0,0) {\tiny \textbf{30\%}};
    \node [ffwnode,anchor=west, fill=cyan!65!white] (ffw1) at ([xshift=0.75em]msa1.east) {\tiny \textbf{65\%}};

    \node [anchor=south west] at ([xshift=-1.25em,yshift=0.0em]msa1.south west) {\textbf{1}};
    
    \node [ffwnode,anchor=south, fill=cyan!51!white] (ffw2) at ([yshift=0.4em]ffw1.north) {\tiny \textbf{51\%}};
    \node [msanode,anchor=south, fill=green!51!white] (msa2) at ([yshift=0.4em]msa1.north) {\tiny \textbf{51\%}};
    \node [anchor=south west] at ([xshift=-1.25em,yshift=0.0em]msa2.south west) {\textbf{2}};
    
    \node [ffwnode,anchor=south, fill=cyan!25!white] (ffw3) at ([yshift=0.4em]ffw2.north) {\tiny \textbf{25\%}};
    \node [msanode,anchor=south, fill=green!40!white] (msa3) at ([yshift=0.4em]msa2.north) {\tiny \textbf{40\%}};
    \node [anchor=south west] at ([xshift=-1.25em,yshift=0.0em]msa3.south west) {\textbf{3}};
    
    \node [ffwnode,anchor=south, fill=cyan!35!white] (ffw4) at ([yshift=0.4em]ffw3.north) {\tiny \textbf{35\%}};
    \node [msanode,anchor=south, fill=green!26!white] (msa4) at ([yshift=0.4em]msa3.north) {\tiny \textbf{26\%}};
    \node [anchor=south west] at ([xshift=-1.25em,yshift=0.0em]msa4.south west) {\textbf{4}};
    
    \node [ffwnode,anchor=south, fill=cyan!36!white] (ffw5) at ([yshift=0.4em]ffw4.north) {\tiny \textbf{36\%}};
    \node [msanode,anchor=south, fill=green!29!white] (msa5) at ([yshift=0.4em]msa4.north) {\tiny \textbf{29\%}};
    \node [anchor=south west] at ([xshift=-1.25em,yshift=0.0em]msa5.south west) {\textbf{5}};
    
    \node [ffwnode,anchor=south, fill=cyan!55!white] (ffw6) at ([yshift=0.4em]ffw5.north) {\tiny \textbf{55\%}};
    \node [msanode,anchor=south, fill=green!17!white] (msa6) at ([yshift=0.4em]msa5.north) {\tiny \textbf{17\%}};
    \node [anchor=south west] at ([xshift=-1.25em,yshift=0.0em]msa6.south west) {\textbf{6}};
    
    \node [anchor=south] (encoder) at ([xshift=0.5em,yshift=0.5em]msa6.north) {\textbf{Encoder}};
    
    \draw[rounded corners=10pt, thick, draw=black] ([xshift=-1.25em,yshift=-0.5em]msa1.south west) rectangle ([xshift=.5em,yshift=0.5em]ffw6.north east);
        
    \node [msanode,anchor=west, fill=green!21!white] (msa_d1) at ([xshift=6em]msa1.east) {\tiny \textbf{21\%}};
    \node [mcanode,anchor=west, fill=purple!60!white] (mca_d1) at ([xshift=0.75em]msa_d1.east) {\tiny \textbf{60\%}};
    \node [ffwnode,anchor=west, fill=cyan!57!white] (ffw_d1) at ([xshift=0.75em]mca_d1.east) {\tiny \textbf{57\%}};

    
    \node [ffwnode,anchor=south, fill=cyan!29!white] (ffw_d2) at ([yshift=0.4em]ffw_d1.north) {\tiny \textbf{29\%}};
    \node [msanode,anchor=south, fill=green!62!white] (msa_d2) at ([yshift=0.4em]msa_d1.north) {\tiny \textbf{62\%}};
    \node [mcanode,anchor=south, fill=purple!53!white] (mca_d2) at ([yshift=0.4em]mca_d1.north) {\tiny \textbf{53\%}};
        
    \node [ffwnode,anchor=south, fill=cyan!13!white] (ffw_d3) at ([yshift=0.4em]ffw_d2.north) {\tiny \textbf{13\%}};
    \node [msanode,anchor=south, fill=green!36!white] (msa_d3) at ([yshift=0.4em]msa_d2.north) {\tiny \textbf{36\%}};
    \node [mcanode,anchor=south, fill=purple!31!white] (mca_d3) at ([yshift=0.4em]mca_d2.north) {\tiny \textbf{31\%}};
    
    \node [ffwnode,anchor=south, fill=cyan!6!white] (ffw_d4) at ([yshift=0.4em]ffw_d3.north) {\tiny \textbf{6.0\%}};
    \node [msanode,anchor=south, fill=green!22!white] (msa_d4) at ([yshift=0.4em]msa_d3.north) {\tiny \textbf{22\%}};
    \node [mcanode,anchor=south, fill=purple!44!white] (mca_d4) at ([yshift=0.4em]mca_d3.north) {\tiny \textbf{44\%}};
    
    \node [ffwnode,anchor=south, fill=cyan!4!white] (ffw_d5) at ([yshift=0.4em]ffw_d4.north) {\tiny \textbf{4.0\%}};
    \node [msanode,anchor=south, fill=green!11!white] (msa_d5) at ([yshift=0.4em]msa_d4.north) {\tiny \textbf{11\%}};
    \node [mcanode,anchor=south, fill=purple!35!white] (mca_d5) at ([yshift=0.4em]mca_d4.north) {\tiny \textbf{35\%}};
    
    \node [ffwnode,anchor=south, fill=cyan!5!white] (ffw_d6) at ([yshift=0.4em]ffw_d5.north) {\tiny \textbf{5.0\%}};
    \node [msanode,anchor=south, fill=green!10!white] (msa_d6) at ([yshift=0.4em]msa_d5.north) {\tiny \textbf{10\%}};
    \node [mcanode,anchor=south, fill=purple!49!white] (mca_d6) at ([yshift=0.4em]mca_d5.north) {\tiny \textbf{49\%}};
    
    \node [anchor=south] (decoder) at ([xshift=1.0em,yshift=0.5em]msa_d6.north) {\textbf{Decoder}};
    
    \draw[rounded corners=10pt, thick, draw=black] ([xshift=-0.5em,yshift=-0.5em]msa_d1.south west) rectangle ([xshift=0.5em,yshift=0.5em]ffw_d6.north east);

    \draw[->,standard] ([xshift=0.5em,yshift=0em]ffw4.east) -- ([xshift=-.5em,yshift=0em]msa_d4.west);

    \node [ffwnode,anchor=south] (legend_ffw_encoder) at ([yshift=2.5em]ffw6.north) {\tiny \textbf{FF}};
    \node [msanode,anchor=south] (legend_msa_encoder) at ([yshift=2.5em]msa6.north) {\tiny \textbf{MHSA}};
    
    \node [ffwnode,anchor=south] (legend_ffw) at ([yshift=2.5em]ffw_d6.north) {\tiny \textbf{FF}};
    \node [msanode,anchor=south] (legend_msa) at ([yshift=2.5em]msa_d6.north) {\tiny \textbf{MHSA}};
    \node [mcanode,anchor=south] (legend_mca) at ([yshift=2.5em]mca_d6.north) {\tiny \textbf{MHCA}};

    \draw[->, standard] ([yshift=-1.5em, xshift=1em]msa1.south) -- ([yshift=-0.5em, xshift=1em]msa1.south);
    \draw[->, standard] ([yshift=-1.5em]mca_d1.south) -- ([yshift=-0.5em]mca_d1.south);
        
\end{tikzpicture}

%% file: figures/tikz/local_sparsity_prepend.tex
\begin{tikzpicture}
    \tikzstyle{standard} = [rounded corners=3pt];
    \tikzstyle{ffwnode} = [minimum height=1.5em,minimum width=2.0em,inner sep=1pt,rounded corners=3pt,draw,fill=cyan!30];
    \tikzstyle{msanode} = [minimum height=1.5em,minimum width=2.0em,inner sep=1pt,rounded corners=3pt,draw,fill=green!30];
    \tikzstyle{mcanode} = [minimum height=1.5em,minimum width=2.0em,inner sep=1pt,rounded corners=3pt,draw,fill=purple!30];
    
    \node [msanode,anchor=west, fill=green!76!white] (msa1) at (0,0) {\tiny \textbf{76\%}};
    \node [ffwnode,anchor=west, fill=cyan!47!white] (ffw1) at ([xshift=0.75em]msa1.east) {\tiny \textbf{47\%}};
    \node [anchor=south west] at ([xshift=-1.25em,yshift=0.0em]msa1.south west) {\textbf{1}};
    
    \node [ffwnode,anchor=south, fill=cyan!63!white] (ffw2) at ([yshift=0.4em]ffw1.north) {\tiny \textbf{63\%}};
    \node [msanode,anchor=south, fill=green!66!white] (msa2) at ([yshift=0.4em]msa1.north) {\tiny \textbf{66\%}};
    \node [anchor=south west] at ([xshift=-1.25em,yshift=0.0em]msa2.south west) {\textbf{2}};
    
    \node [ffwnode,anchor=south, fill=cyan!43!white] (ffw3) at ([yshift=0.4em]ffw2.north) {\tiny \textbf{43\%}};
    \node [msanode,anchor=south, fill=green!43!white] (msa3) at ([yshift=0.4em]msa2.north) {\tiny \textbf{43\%}};
    \node [anchor=south west] at ([xshift=-1.25em,yshift=0.0em]msa3.south west) {\textbf{3}};
    
    \node [ffwnode,anchor=south, fill=cyan!34!white] (ffw4) at ([yshift=0.4em]ffw3.north) {\tiny \textbf{34\%}};
    \node [msanode,anchor=south, fill=green!45!white] (msa4) at ([yshift=0.4em]msa3.north) {\tiny \textbf{45\%}};
    \node [anchor=south west] at ([xshift=-1.25em,yshift=0.0em]msa4.south west) {\textbf{4}};
    
    \node [ffwnode,anchor=south, fill=cyan!47!white] (ffw5) at ([yshift=0.4em]ffw4.north) {\tiny \textbf{47\%}};
    \node [msanode,anchor=south, fill=green!32!white] (msa5) at ([yshift=0.4em]msa4.north) {\tiny \textbf{32\%}};
    \node [anchor=south west] at ([xshift=-1.25em,yshift=0.0em]msa5.south west) {\textbf{5}};
    
    \node [ffwnode,anchor=south, fill=cyan!52!white] (ffw6) at ([yshift=0.4em]ffw5.north) {\tiny \textbf{52\%}};
    \node [msanode,anchor=south, fill=green!29!white] (msa6) at ([yshift=0.4em]msa5.north) {\tiny \textbf{29\%}};
    \node [anchor=south west] at ([xshift=-1.25em,yshift=0.0em]msa6.south west) {\textbf{6}};
    
    \node [anchor=south] (encoder) at ([xshift=0.5em,yshift=0.5em]msa6.north) {\textbf{Encoder}};
    
    \draw[rounded corners=10pt, thick, draw=black] ([xshift=-1.25em,yshift=-0.5em]msa1.south west) rectangle ([xshift=.5em,yshift=0.5em]ffw6.north east);
        
    \node [msanode,anchor=west, fill=green!27!white] (msa_d1) at ([xshift=6em]msa1.east) {\tiny \textbf{27\%}};
    \node [mcanode,anchor=west, fill=purple!65!white] (mca_d1) at ([xshift=0.75em]msa_d1.east) {\tiny \textbf{65\%}};
    \node [ffwnode,anchor=west, fill=cyan!59!white] (ffw_d1) at ([xshift=0.75em]mca_d1.east) {\tiny \textbf{59\%}};
    
    \node [ffwnode,anchor=south, fill=cyan!30!white] (ffw_d2) at ([yshift=0.4em]ffw_d1.north) {\tiny \textbf{30\%}};
    \node [msanode,anchor=south, fill=green!68!white] (msa_d2) at ([yshift=0.4em]msa_d1.north) {\tiny \textbf{68\%}};
    \node [mcanode,anchor=south, fill=purple!56!white] (mca_d2) at ([yshift=0.4em]mca_d1.north) {\tiny \textbf{56\%}};
        
    \node [ffwnode,anchor=south, fill=cyan!16!white] (ffw_d3) at ([yshift=0.4em]ffw_d2.north) {\tiny \textbf{16\%}};
    \node [msanode,anchor=south, fill=green!40!white] (msa_d3) at ([yshift=0.4em]msa_d2.north) {\tiny \textbf{40\%}};
    \node [mcanode,anchor=south, fill=purple!30!white] (mca_d3) at ([yshift=0.4em]mca_d2.north) {\tiny \textbf{30\%}};
    
    \node [ffwnode,anchor=south, fill=cyan!1!white] (ffw_d4) at ([yshift=0.4em]ffw_d3.north) {\tiny \textbf{1.0\%}};
    \node [msanode,anchor=south, fill=green!28!white] (msa_d4) at ([yshift=0.4em]msa_d3.north) {\tiny \textbf{28\%}};
    \node [mcanode,anchor=south, fill=purple!43!white] (mca_d4) at ([yshift=0.4em]mca_d3.north) {\tiny \textbf{43\%}};
    
    \node [ffwnode,anchor=south, fill=cyan!1!white] (ffw_d5) at ([yshift=0.4em]ffw_d4.north) {\tiny \textbf{1.0\%}};
    \node [msanode,anchor=south, fill=green!11!white] (msa_d5) at ([yshift=0.4em]msa_d4.north) {\tiny \textbf{11\%}};
    \node [mcanode,anchor=south, fill=purple!38!white] (mca_d5) at ([yshift=0.4em]mca_d4.north) {\tiny \textbf{38\%}};
    
    \node [ffwnode,anchor=south, fill=cyan!1!white] (ffw_d6) at ([yshift=0.4em]ffw_d5.north) {\tiny \textbf{1.0\%}};
    \node [msanode,anchor=south, fill=green!1!white] (msa_d6) at ([yshift=0.4em]msa_d5.north) {\tiny \textbf{1.0\%}};
    \node [mcanode,anchor=south, fill=purple!48!white] (mca_d6) at ([yshift=0.4em]mca_d5.north) {\tiny \textbf{48\%}};
    
    \node [anchor=south] (decoder) at ([xshift=1.0em,yshift=0.5em]msa_d6.north) {\textbf{Decoder}};
    
    \draw[rounded corners=10pt, thick, draw=black] ([xshift=-0.5em,yshift=-0.5em]msa_d1.south west) rectangle ([xshift=0.5em,yshift=0.5em]ffw_d6.north east);

    \draw[->,standard] ([xshift=0.5em,yshift=0em]ffw4.east) -- ([xshift=-.5em,yshift=0em]msa_d4.west);

    \node [ffwnode,anchor=south] (legend_ffw_encoder) at ([yshift=2.5em]ffw6.north) {\tiny \textbf{FF}};
    \node [msanode,anchor=south] (legend_msa_encoder) at ([yshift=2.5em]msa6.north) {\tiny \textbf{MHSA}};
    
    \node [ffwnode,anchor=south] (legend_ffw) at ([yshift=2.5em]ffw_d6.north) {\tiny \textbf{FF}};
    \node [msanode,anchor=south] (legend_msa) at ([yshift=2.5em]msa_d6.north) {\tiny \textbf{MHSA}};
    \node [mcanode,anchor=south] (legend_mca) at ([yshift=2.5em]mca_d6.north) {\tiny \textbf{MHCA}};

    \draw[->, standard] ([yshift=-1.5em, xshift=1em]msa1.south) -- ([yshift=-0.5em, xshift=1em]msa1.south);
    \draw[->, standard] ([yshift=-1.5em]mca_d1.south) -- ([yshift=-0.5em]mca_d1.south);
        
\end{tikzpicture}

%% file: figures/tikz/local_sparsity_remove_first.tex
\begin{tikzpicture}
    \tikzstyle{standard} = [rounded corners=3pt];
    \tikzstyle{ffwnode} = [minimum height=1.5em,minimum width=2.0em,inner sep=1pt,rounded corners=3pt,draw,fill=cyan!30];
    \tikzstyle{msanode} = [minimum height=1.5em,minimum width=2.0em,inner sep=1pt,rounded corners=3pt,draw,fill=green!30];
    \tikzstyle{mcanode} = [minimum height=1.5em,minimum width=2.0em,inner sep=1pt,rounded corners=3pt,draw,fill=purple!30];
    
    \node [msanode,anchor=west, fill=green!43!white] (msa1) at (0,0) {\tiny \textbf{43\%}};
    \node [ffwnode,anchor=west, fill=cyan!78!white] (ffw1) at ([xshift=0.75em]msa1.east) {\tiny \textbf{78\%}};
    \node [anchor=south west] at ([xshift=-1.25em,yshift=0.0em]msa1.south west) {\textbf{1}};
    
    \node [ffwnode,anchor=south, fill=cyan!63!white] (ffw2) at ([yshift=0.4em]ffw1.north) {\tiny \textbf{63\%}};
    \node [msanode,anchor=south, fill=green!54!white] (msa2) at ([yshift=0.4em]msa1.north) {\tiny \textbf{54\%}};
    \node [anchor=south west] at ([xshift=-1.25em,yshift=0.0em]msa2.south west) {\textbf{2}};
    
    \node [ffwnode,anchor=south, fill=cyan!48!white] (ffw3) at ([yshift=0.4em]ffw2.north) {\tiny \textbf{48\%}};
    \node [msanode,anchor=south, fill=green!41!white] (msa3) at ([yshift=0.4em]msa2.north) {\tiny \textbf{41\%}};
    \node [anchor=south west] at ([xshift=-1.25em,yshift=0.0em]msa3.south west) {\textbf{3}};
    
    \node [ffwnode,anchor=south, fill=cyan!46!white] (ffw4) at ([yshift=0.4em]ffw3.north) {\tiny \textbf{46\%}};
    \node [msanode,anchor=south, fill=green!32!white] (msa4) at ([yshift=0.4em]msa3.north) {\tiny \textbf{32\%}};
    \node [anchor=south west] at ([xshift=-1.25em,yshift=0.0em]msa4.south west) {\textbf{4}};
    
    \node [ffwnode,anchor=south, fill=cyan!44!white] (ffw5) at ([yshift=0.4em]ffw4.north) {\tiny \textbf{44\%}};
    \node [msanode,anchor=south, fill=green!25!white] (msa5) at ([yshift=0.4em]msa4.north) {\tiny \textbf{25\%}};
    \node [anchor=south west] at ([xshift=-1.25em,yshift=0.0em]msa5.south west) {\textbf{5}};
    
    \node [ffwnode,anchor=south, fill=cyan!41!white] (ffw6) at ([yshift=0.4em]ffw5.north) {\tiny \textbf{41\%}};
    \node [msanode,anchor=south, fill=green!24!white] (msa6) at ([yshift=0.4em]msa5.north) {\tiny \textbf{24\%}};
    \node [anchor=south west] at ([xshift=-1.25em,yshift=0.0em]msa6.south west) {\textbf{6}};
    
    \node [anchor=south] (encoder) at ([xshift=0.5em,yshift=0.5em]msa6.north) {\textbf{Encoder}};
    
    \draw[rounded corners=10pt, thick, draw=black] ([xshift=-1.25em,yshift=-0.5em]msa1.south west) rectangle ([xshift=.5em,yshift=0.5em]ffw6.north east);
        
    \node [msanode,anchor=west, fill=green!5!white] (msa_d1) at ([xshift=6em]msa1.east) {\tiny \textbf{5.0\%}};
    \node [mcanode,anchor=west, fill=purple!29!white] (mca_d1) at ([xshift=0.75em]msa_d1.east) {\tiny \textbf{29\%}};
    \node [ffwnode,anchor=west, fill=cyan!25!white] (ffw_d1) at ([xshift=0.75em]mca_d1.east) {\tiny \textbf{25\%}};
    
    \node [ffwnode,anchor=south, fill=cyan!18!white] (ffw_d2) at ([yshift=0.4em]ffw_d1.north) {\tiny \textbf{18\%}};
    \node [msanode,anchor=south, fill=green!18!white] (msa_d2) at ([yshift=0.4em]msa_d1.north) {\tiny \textbf{18\%}};
    \node [mcanode,anchor=south, fill=purple!34!white] (mca_d2) at ([yshift=0.4em]mca_d1.north) {\tiny \textbf{34\%}};
        
    \node [ffwnode,anchor=south, fill=cyan!6!white] (ffw_d3) at ([yshift=0.4em]ffw_d2.north) {\tiny \textbf{6.0\%}};
    \node [msanode,anchor=south, fill=green!8!white] (msa_d3) at ([yshift=0.4em]msa_d2.north) {\tiny \textbf{8.0\%}};
    \node [mcanode,anchor=south, fill=purple!27!white] (mca_d3) at ([yshift=0.4em]mca_d2.north) {\tiny \textbf{27\%}};
    
    \node [ffwnode,anchor=south, fill=cyan!1!white] (ffw_d4) at ([yshift=0.4em]ffw_d3.north) {\tiny \textbf{3.0\%}};
    \node [msanode,anchor=south, fill=green!1!white] (msa_d4) at ([yshift=0.4em]msa_d3.north) {\tiny \textbf{6.0\%}};
    \node [mcanode,anchor=south, fill=purple!21!white] (mca_d4) at ([yshift=0.4em]mca_d3.north) {\tiny \textbf{32\%}};
    
    \node [ffwnode,anchor=south, fill=cyan!3!white] (ffw_d5) at ([yshift=0.4em]ffw_d4.north) {\tiny \textbf{3.0\%}};
    \node [msanode,anchor=south, fill=green!5!white] (msa_d5) at ([yshift=0.4em]msa_d4.north) {\tiny \textbf{5.0\%}};
    \node [mcanode,anchor=south, fill=purple!25!white] (mca_d5) at ([yshift=0.4em]mca_d4.north) {\tiny \textbf{25\%}};
    
    \node [ffwnode,anchor=south, fill=cyan!4!white] (ffw_d6) at ([yshift=0.4em]ffw_d5.north) {\tiny \textbf{4.0\%}};
    \node [msanode,anchor=south, fill=green!7!white] (msa_d6) at ([yshift=0.4em]msa_d5.north) {\tiny \textbf{7.0\%}};
    \node [mcanode,anchor=south, fill=purple!48!white] (mca_d6) at ([yshift=0.4em]mca_d5.north) {\tiny \textbf{38\%}};
    
    \node [anchor=south] (decoder) at ([xshift=1.0em,yshift=0.5em]msa_d6.north) {\textbf{Decoder}};
    
    \draw[rounded corners=10pt, thick, draw=black] ([xshift=-0.5em,yshift=-0.5em]msa_d1.south west) rectangle ([xshift=0.5em,yshift=0.5em]ffw_d6.north east);

    \draw[->,standard] ([xshift=0.5em,yshift=0em]ffw4.east) -- ([xshift=-.5em,yshift=0em]msa_d4.west);

    \node [ffwnode,anchor=south] (legend_ffw_encoder) at ([yshift=2.5em]ffw6.north) {\tiny \textbf{FF}};
    \node [msanode,anchor=south] (legend_msa_encoder) at ([yshift=2.5em]msa6.north) {\tiny \textbf{MHSA}};
    
    \node [ffwnode,anchor=south] (legend_ffw) at ([yshift=2.5em]ffw_d6.north) {\tiny \textbf{FF}};
    \node [msanode,anchor=south] (legend_msa) at ([yshift=2.5em]msa_d6.north) {\tiny \textbf{MHSA}};
    \node [mcanode,anchor=south] (legend_mca) at ([yshift=2.5em]mca_d6.north) {\tiny \textbf{MHCA}};

    \draw[->, standard] ([yshift=-1.5em, xshift=1em]msa1.south) -- ([yshift=-0.5em, xshift=1em]msa1.south);
    \draw[->, standard] ([yshift=-1.5em]mca_d1.south) -- ([yshift=-0.5em]mca_d1.south);
        
\end{tikzpicture}

%% file: figures/tikz/local_sparsity_remove_second.tex
\begin{tikzpicture}
    \tikzstyle{standard} = [rounded corners=3pt];
    \tikzstyle{ffwnode} = [minimum height=1.5em,minimum width=2.0em,inner sep=1pt,rounded corners=3pt,draw,fill=cyan!30];
    \tikzstyle{msanode} = [minimum height=1.5em,minimum width=2.0em,inner sep=1pt,rounded corners=3pt,draw,fill=green!30];
    \tikzstyle{mcanode} = [minimum height=1.5em,minimum width=2.0em,inner sep=1pt,rounded corners=3pt,draw,fill=purple!30];
    
    \node [msanode,anchor=west, fill=green!13!white] (msa1) at (0,0) {\tiny \textbf{13\%}};
    \node [ffwnode,anchor=west, fill=cyan!32!white] (ffw1) at ([xshift=0.75em]msa1.east) {\tiny \textbf{32\%}};
    \node [anchor=south west] at ([xshift=-1.25em,yshift=0.0em]msa1.south west) {\textbf{1}};
    
    \node [ffwnode,anchor=south, fill=cyan!28!white] (ffw2) at ([yshift=0.4em]ffw1.north) {\tiny \textbf{28\%}};
    \node [msanode,anchor=south, fill=green!27!white] (msa2) at ([yshift=0.4em]msa1.north) {\tiny \textbf{27\%}};
    \node [anchor=south west] at ([xshift=-1.25em,yshift=0.0em]msa2.south west) {\textbf{2}};
    
    \node [ffwnode,anchor=south, fill=cyan!29!white] (ffw3) at ([yshift=0.4em]ffw2.north) {\tiny \textbf{29\%}};
    \node [msanode,anchor=south, fill=green!21!white] (msa3) at ([yshift=0.4em]msa2.north) {\tiny \textbf{21\%}};
    \node [anchor=south west] at ([xshift=-1.25em,yshift=0.0em]msa3.south west) {\textbf{3}};
    
    \node [ffwnode,anchor=south, fill=cyan!44!white] (ffw4) at ([yshift=0.4em]ffw3.north) {\tiny \textbf{44\%}};
    \node [msanode,anchor=south, fill=green!19!white] (msa4) at ([yshift=0.4em]msa3.north) {\tiny \textbf{19\%}};
    \node [anchor=south west] at ([xshift=-1.25em,yshift=0.0em]msa4.south west) {\textbf{4}};
    
    \node [ffwnode,anchor=south, fill=cyan!33!white] (ffw5) at ([yshift=0.4em]ffw4.north) {\tiny \textbf{33\%}};
    \node [msanode,anchor=south, fill=green!18!white] (msa5) at ([yshift=0.4em]msa4.north) {\tiny \textbf{18\%}};
    \node [anchor=south west] at ([xshift=-1.25em,yshift=0.0em]msa5.south west) {\textbf{5}};
    
    \node [ffwnode,anchor=south, fill=cyan!33!white] (ffw6) at ([yshift=0.4em]ffw5.north) {\tiny \textbf{33\%}};
    \node [msanode,anchor=south, fill=green!13!white] (msa6) at ([yshift=0.4em]msa5.north) {\tiny \textbf{13\%}};
    \node [anchor=south west] at ([xshift=-1.25em,yshift=0.0em]msa6.south west) {\textbf{6}};
    
    \node [anchor=south] (encoder) at ([xshift=0.5em,yshift=0.5em]msa6.north) {\textbf{Encoder}};
    
    \draw[rounded corners=10pt, thick, draw=black] ([xshift=-1.25em,yshift=-0.5em]msa1.south west) rectangle ([xshift=.5em,yshift=0.5em]ffw6.north east);
        
    \node [msanode,anchor=west, fill=green!0!white] (msa_d1) at ([xshift=6em]msa1.east) {\tiny \textbf{0.0\%}};
    \node [mcanode,anchor=west, fill=purple!16!white] (mca_d1) at ([xshift=0.75em]msa_d1.east) {\tiny \textbf{16\%}};
    \node [ffwnode,anchor=west, fill=cyan!15!white] (ffw_d1) at ([xshift=0.75em]mca_d1.east) {\tiny \textbf{15\%}};
    
    \node [ffwnode,anchor=south, fill=cyan!10!white] (ffw_d2) at ([yshift=0.4em]ffw_d1.north) {\tiny \textbf{10\%}};
    \node [msanode,anchor=south, fill=green!1!white] (msa_d2) at ([yshift=0.4em]msa_d1.north) {\tiny \textbf{1.0\%}};
    \node [mcanode,anchor=south, fill=purple!23!white] (mca_d2) at ([yshift=0.4em]mca_d1.north) {\tiny \textbf{23\%}};
        
    \node [ffwnode,anchor=south, fill=cyan!2!white] (ffw_d3) at ([yshift=0.4em]ffw_d2.north) {\tiny \textbf{2.0\%}};
    \node [msanode,anchor=south, fill=green!0!white] (msa_d3) at ([yshift=0.4em]msa_d2.north) {\tiny \textbf{0.0\%}};
    \node [mcanode,anchor=south, fill=purple!18!white] (mca_d3) at ([yshift=0.4em]mca_d2.north) {\tiny \textbf{18\%}};
    
    \node [ffwnode,anchor=south, fill=cyan!1!white] (ffw_d4) at ([yshift=0.4em]ffw_d3.north) {\tiny \textbf{1.0\%}};
    \node [msanode,anchor=south, fill=green!1!white] (msa_d4) at ([yshift=0.4em]msa_d3.north) {\tiny \textbf{1.0\%}};
    \node [mcanode,anchor=south, fill=purple!21!white] (mca_d4) at ([yshift=0.4em]mca_d3.north) {\tiny \textbf{21\%}};
    
    \node [ffwnode,anchor=south, fill=cyan!1!white] (ffw_d5) at ([yshift=0.4em]ffw_d4.north) {\tiny \textbf{1.0\%}};
    \node [msanode,anchor=south, fill=green!1!white] (msa_d5) at ([yshift=0.4em]msa_d4.north) {\tiny \textbf{1.0\%}};
    \node [mcanode,anchor=south, fill=purple!21!white] (mca_d5) at ([yshift=0.4em]mca_d4.north) {\tiny \textbf{21\%}};
    
    \node [ffwnode,anchor=south, fill=cyan!2!white] (ffw_d6) at ([yshift=0.4em]ffw_d5.north) {\tiny \textbf{2.0\%}};
    \node [msanode,anchor=south, fill=green!2!white] (msa_d6) at ([yshift=0.4em]msa_d5.north) {\tiny \textbf{2.0\%}};
    \node [mcanode,anchor=south, fill=purple!41!white] (mca_d6) at ([yshift=0.4em]mca_d5.north) {\tiny \textbf{41\%}};
    
    \node [anchor=south] (decoder) at ([xshift=1.0em,yshift=0.5em]msa_d6.north) {\textbf{Decoder}};
    
    \draw[rounded corners=10pt, thick, draw=black] ([xshift=-0.5em,yshift=-0.5em]msa_d1.south west) rectangle ([xshift=0.5em,yshift=0.5em]ffw_d6.north east);

    \draw[->,standard] ([xshift=0.5em,yshift=0em]ffw4.east) -- ([xshift=-.5em,yshift=0em]msa_d4.west);

    \node [ffwnode,anchor=south] (legend_ffw_encoder) at ([yshift=2.5em]ffw6.north) {\tiny \textbf{FF}};
    \node [msanode,anchor=south] (legend_msa_encoder) at ([yshift=2.5em]msa6.north) {\tiny \textbf{MHSA}};
    
    \node [ffwnode,anchor=south] (legend_ffw) at ([yshift=2.5em]ffw_d6.north) {\tiny \textbf{FF}};
    \node [msanode,anchor=south] (legend_msa) at ([yshift=2.5em]msa_d6.north) {\tiny \textbf{MHSA}};
    \node [mcanode,anchor=south] (legend_mca) at ([yshift=2.5em]mca_d6.north) {\tiny \textbf{MHCA}};

    \draw[->, standard] ([yshift=-1.5em, xshift=1em]msa1.south) -- ([yshift=-0.5em, xshift=1em]msa1.south);
    \draw[->, standard] ([yshift=-1.5em]mca_d1.south) -- ([yshift=-0.5em]mca_d1.south);
        
\end{tikzpicture}